\documentclass[journal,9pt]{IEEEtran}

\usepackage{color,xcolor}

\usepackage{bbm}
\usepackage{balance}
\usepackage{graphicx}

\usepackage{url}

\usepackage{amsmath,amssymb,amsfonts}

\usepackage{algorithm}
\usepackage{algorithmic,eqparbox,array}

\usepackage{multirow}

\usepackage{float}
\usepackage{cite}

\usepackage{lipsum}

\def\POV{t}

\begin{document}

\title{Approximation of Images via Generalized Higher Order Singular Value Decomposition over Finite-dimensional Commutative Semisimple Algebra}

\author{
	Liang Liao$^1$, 
	Sen Lin$^1$, 
	Lun Li$^2$, 
	Xiuwei Zhang$^3$,
	Song Zhao$^4$,  
	Yan Wang$^1$, 
	Xinqiang Wang$^1$,
	Qi Gao$^1$, 
	Jingyu Wang$^1$	\\
	\vspace{0.5em}
	$^1\,$School of Electronics and Information, Zhongyuan University of Technology, China 
	\vspace{0.2em}\\
	$^2\,$School of Information Engineering, Zhengzhou University, China \vspace{0.2em}\\
	$^3\,$School of Computer Science, Northwestern Polytechnical University, China \vspace{0.2em}\\
	$^4\,$School of Intelligent Engineering, Zhengzhou University of Aeronautics, China \vspace{1em} \\
	liaoliang@ieee.org or liaoliangis@126.com \\
}

\maketitle

\begin{abstract}
Low-rank approximation of images via singular value decomposition is well-received in the era of big data. However, singular value decomposition (SVD) is only for 
order-two data, i.e., matrices. 
It is necessary to flatten a higher order input into a matrix or break it 
into a series of order-two slices
to tackle higher order data such  as multispectral images and videos with the SVD. 
Higher order singular value decomposition (HOSVD) extends 
the
SVD and can approximate higher order data 
using sums of a few rank-one
components.
We consider the problem of generalizing HOSVD over a finite-dimensional commutative algebra. 
This algebra, referred to as t-algebra, generalizes the field of complex numbers. 
The elements of the algebra, called t-scalars, are fixed-sized arrays of complex numbers. 
To obtain higher-performance versions, one can generalize matrices and tensors over t-scalars and then extend many canonical matrix and tensor algorithms, including HOSVD. 
The generalization of HOSVD is called THOSVD. 
An alternating algorithm can further improve its performance in approximating multiway data.
THOSVD also unifies a wide range of principal component analysis algorithms. 
To exploit the potential of generalized algorithms using t-scalars for approximating images, we use a pixel neighborhood strategy to convert each pixel to a ``deeper-order'' t-scalar. Experiments on publicly available images show that the generalized algorithm over t-scalars, namely THOSVD, compares favorably with its canonical counterparts.
\end{abstract}


\section{Introduction}
\label{section:001}

\subsection{Backgrounds}
In the era of information deluge, data are often multiway arrays. For example, numerical multispectral or hyperspectral images and videos are all multiway formats.

Concerning the spatial properties of multiway data, many authors studied tensorial algorithms rather than mere matrix algorithms in data analysis.  
Hong et al. apply singular value decomposition (SVD) to image recognition and use the results of SVD to represent the features of images \cite{1991Algebraic}. 
Then, many scholars have accepted the SVD as an essential tool for analyzing images and applied it to image compression \cite{2005Mei,2019Dabass}, watermarking \cite{2020RST,2021Hybrid}, and low-rank approximation \cite{2017Computing,2015Li}, to name a few.

However, SVD does not directly involve multiway data in a non-matrix format. Tensor algorithms, backed by multilinear algebras are more effective than the SVD in capturing  multiway structural information.

Tensors and their applications have developed since the last century. One pioneer of tenor decompositions and their applications is F. L. Hitchcock \cite{1928Multiple}. After Hitchcock, Tucker et al. studied tensor decomposition further and proposed the well-received Tucker decomposition algorithms \cite{Tucker1963, Tucker1964, Tucker1966}. Since Tucker’s work, many authors have developed various tensor algorithms with applications, including but not limited to signal processing \cite{2017Tensor}, computer vision \cite{2021Panagakis} and image analysis \cite{2018Pre,2017Ren}.

Among the Tucker-decomposition-based algorithms extending the SVD, higher order singular value decomposition (HOSVD) advertised by Lathauwer et al. is well received \cite{1994Blind,2000A}. HOSVD can be used for applications such as image fusion \cite{2011Higher,2012Image}, image denoising \cite{2015Denoising} and target detection  \cite{2013A}.

However, HOSVD is a higher-order generalization of SVD rather than a ``deeper order'' generalization. 
Here, ``deeper order'' means replacing the complex numbers by the elements of a finite dimensional algebra.
Matrices and tensors with entries in a finite dimensional algebra have been investigated. For example, many authors have constructed matrix algorithms 
over Hamilton’s quaternions. Quaternions form a finite-dimensional non-commutative algebra generalizing the field of complex numbers. 
Quaternion matrices and tensors are used in histopathological image analysis \cite{shi2019quaternion}, image \& video recovery, and  representations\cite{miao2020low,miao2021color,hosny2019new}, 
color image denoising \cite{chen2019low,yu2019quaternion}, 
color face recognition \cite{zou2016quaternion,zou2019grayscale}, color image 
inpainting\cite{miao2020quaternion,jia2019robust},  
to name a few.

However, the multiplication of quaternions is not commutative. The non-commutativity prevents quaternionic matrices from being straightforward analogues of matrices with complex entries. 
Further, quaternions are only suitable for characterizing objects with four or fewer components.
There is no consistent and straightforward extension to objects with five or more dimensions.  
One might query the existence of a finite-dimensional algebra generalizing complex numbers and allowing a consistent and straightforward extension to higher dimensions?  
More concretely, can one find matrices or tensors with fixed-sized multiway arrays as entries, rather than just complex numbers?
If yes, one can represent the generalized matrices and tensors by multiway arrays where the multiway is two-fold, i.e., a combination of higher-order and deeper-order. Higher-order characterizes their superstructure (rows, columns, and beyond), and 
deeper order characterizes their substructure (i.e., entries).

Concerning this issue, 
Kilmer et al. propose the ``t-product'' model, establishing a finite-dimensional algebra of fixed-sized order-one arrays of real numbers. The t-product model establishes matrices over the algebra. The generalized matrices are order-three arrays of real numbers, two orders (higher-order) for indexing rows and columns, the third (deeper-order) used to characterize generalized 
entries (i.e., t-scalars ) \cite{kilmer2011factorization,kilmer2013third}.

Liao and Maybank extend 
\cite{2017Ren,liao2020generalized,liao2020general}
Kilmer et al.'s generalized scalars from the one-way (i.e., order-one) to multiway and from real to complex, calling the finite-dimensional commutative algebra a
``t-algebra'' and the generalized matrices ``t-matrices''.
Using t-matrices instead of their canonical counterparts, one can generalize many matrix algorithms straightforwardly.

This paper validates the t-matrix model and 
interprets the model
via the standard matrix representation theory. It shows that one can generalize many canonical tensor algorithms, such as 
Higher Order Singular Value Decomposition (HOSVD), by replacing scalars with t-scalars. The generalized tensor algorithm compares favorably with its canonical counterparts for approximating multiway images.

\subsection{Organization of this paper}

The rest of this paper is organized as follows. Section \ref{section:canonical-tenosr-HOSVD} briefly introduces the basic notions and rationale of canonical tensors and HOSVD, laying the foundations for the generalization over t-algebra.

Section \ref{section:tscalars-and-talgebra} introduces the finite-dimensional commutative algebra called t-algebra by Liao and Maybank \cite{liao2020generalized}. The elements of the t-algebra are fixed-sized arrays generalizing complex numbers and enabling one to build generalized vectors, matrices, and even tensors with entries in the t-algebra. These generalized scalars, vectors, and matrices are called t-scalars, t-vectors, and t-matrices. 
The standard matrix representation of t-scalars, t-vectors, and t-matrices is also discussed in this section. 
The standard t-matrix representation is the basis for many linearity issues for modules over the t-algebra and connects them to their t-scalar-valued versions.

A neighborhood strategy to convert a lower-order image is proposed in Section \ref{section:neighborhood-strategy}. 
The neighborhood strategy exploits the potential of the generalized complex numbers, called t-scalars, 
and characterizes the spatial constraints of an image.

Section \ref{section:THOSVD} introduces the generalized tensors over the t-algebra, called g-tensors, and the generalized higher-order singular value decomposition, called THOSVD. THOSVD generalizes the canonical higher-order singular value decomposition (HOSVD) and 
unifies a wide range of algorithms for principal components analysis. Section \ref{section:THOSVD} also discusses the generalization of the higher-order orthogonal iteration algorithm HOOI 
over a t-algebra in order to optimize the performance of THOSVD.

Section \ref{section:005} reports the experimental results on approximating public multiway images and compares the performances of THOSVD and HOSVD and a wide range of PCA-based algorithms from a unified THOSVD perspective. Experiments show that the generalized algorithms over t-scalars compare favorably with their canonical counterparts over canonical scalars. Finally, we conclude this paper in Section \ref{section:Conclusions}.

\section{Canonical tensors and higher order singular value decomposition}
\label{section:canonical-tenosr-HOSVD}

\subsection{Fundamental notions of canonical tensors}

By canonical matrices or canonical tensors, or sometimes just matrices or tensors, we mean the classical matrices or tensors with real or complex numbers as entries. Tensors generalize matrices, vectors, and scalars.

\begin{figure}[bth]
	\centering
	\small
	\begin{tabular}{ccc}
		\includegraphics[width = 0.136\textwidth]{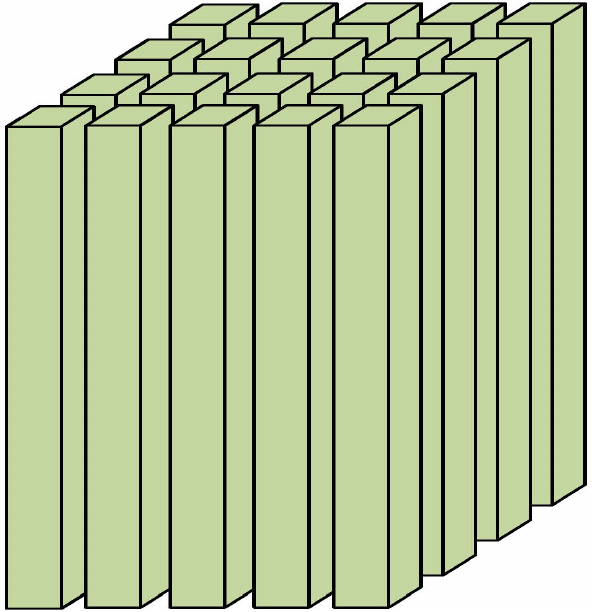}&
		\includegraphics[width = 0.136\textwidth]{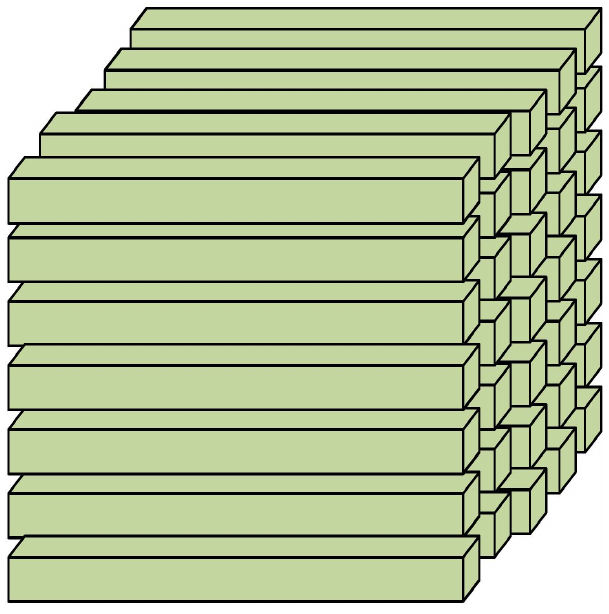}&
		\includegraphics[width = 0.136\textwidth]{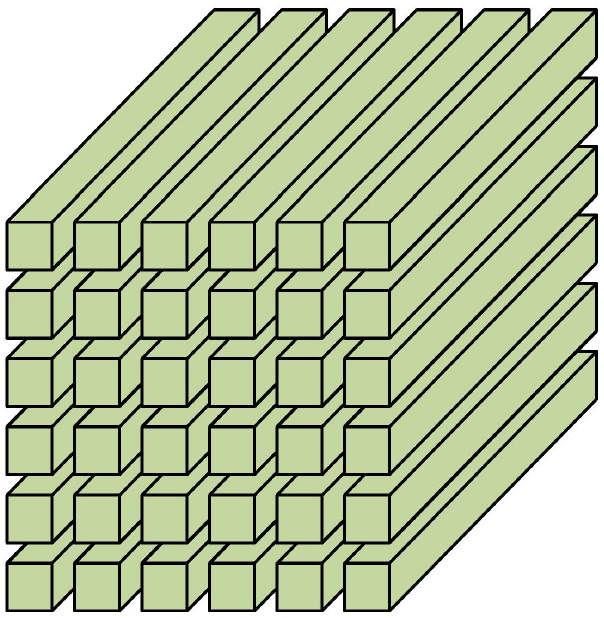} \\	
		(a) &  (b) & (c) \\		
	\end{tabular}
	\caption{Three modes of fibers 
		of a canonical order-three tensor\;\;
		(a) column fibers\;\;
		(b) row fibers\;\;
		(c) tube fibers\;\;		
	}
	\label{fig:tensor-fiber-slice}
\end{figure}

The basic notions can be explained with order-three tensors. 
Let $\mathcal{X}$ be an order-three tensor in 
$\mathbbm{C}^{D_1\times D_2 \times D_3}$. 
Then, $\mathcal{X}$ has three modes of fibers, i.e., column fibers, row fibers, and tube fibers, respectively obtained by fixing all but one of its indices.
If all but two indices are fixed, one has three modes of slices, 
i.e., horizontal slices, lateral slices, and frontal slices. 
Figure \ref{fig:tensor-fiber-slice} shows the column fibers, row fibers, and tube fibers of an order-three tensor.

One can also unfold a higher-order tensor to a matrix. More specifically, the mode-$k$ unfolding of 
a tensor $\mathcal{X} \in \mathbbm{C}^{D_1\times \cdots \times D_M}$ is a matrix ${X}_{(k)} \in \mathbbm{C}^{D_k\times D}$ where $D \doteq \frac{\prod_{m=1}^{M}D_m }{D_{k}} 
$ for all $k \in [M] \doteq \{1,2,\cdots,M\}$ such that 
the $(d_1,\cdots,d_M)$-th scalar entry of $\mathcal{X}$ maps the $(d_k, j)$-th scalar entry of ${X}_{(k)}$, namely
\begin{equation}
	(\mathcal{X})_{d_1,\cdots,d_M} = (X_{(k)})_{d_k, j} \in \mathbbm{C}
\end{equation}
and the following condition holds for all $j \in [D]$,
\begin{equation}
	j = 1 + \sum\nolimits_{m \in [M]\setminus\{k\}}
	(d_m - 1)\cdot J_m  	
\label{equation:index-correspondence}
\end{equation}
where $J_m \doteq \prod\nolimits_{n \in [m-1]\setminus\{k\}} D_n $. 

Namely, the mode-$k$ unfolding $X_{(k)}$  of a tensor has the mode-$k$ fibers as its columns. For example, let $\mathcal{X}$ be a tensor in 
$\mathbbm{C}^{3\times4\times5}$. 
The mode-$1$ unfolding is a $3\times20$ matrix. 
The mode-$2$ unfolding is a $4\times15$ matrix. 
The mode-$3$ unfolding is a $5\times12$ matrix.

Given a tensor $\mathcal{X}\in \mathbbm{C}^{D_1\times \cdots \times D_M}$ and a matrix $Y\in \mathbbm{C}^{J\times D_k}$, the mode-$k$ multiplication 
$\mathcal{A} \doteq \mathcal{X} \,\times_{k}\, Y$ is 
a tensor in $\mathbbm{C}^{D_1\times 
	\cdots \times D_{k-1}\times J \times D_{k+1}\times  \cdots \times D_M}$ such that the following condition holds for all $k \in [M]$, 
\begin{equation}
	A_{(k)} = Y \cdot X_{(k)} \;.
\end{equation}

\subsection{HOSVD: higher-order singular value decomposition}
The higher order singular value
decomposition (HOSVD) is a variant of the Tucker decomposition of a tensor and generalizes the canonical singular value decomposition (SVD). 

Let $\mathcal{X}$ be a tensor in $\mathbbm{C}^{D_1\times D_2 \times \cdots \times D_M}$. HOSVD decomposes $\mathcal{X}$ as the following multimode multiplication
\begin{equation} 
	\mathcal{X} =\mathcal{S} \,\times_{1}\, U_1 \,\times_{2}\, U_2  \cdots \,\times_{M}\, U_M
	\label{equation:HOSVD}
\end{equation}
where the $U_k \in\mathbbm{C}^{D_k\times D_k} $ are unitary matrices
for all $k \in [M]$, whose $i$-th column is the $i$-th leading left singular vector of the mode-$k$ unfolding of $\mathcal{X}$ for all $i \in [D_k]$,  
and $\mathcal{S} \in \mathbbm{C}^{D_1\times \cdots \times D_M}$, called core tensor of $\mathcal{X}$, is computed as follows
\begin{equation} 
	\mathcal{S}=\mathcal{X} \,\times_{1}\, U_1^{*} \,\times_{2}\, U_2^{*}
	\cdots \,\times_{M}\, 
	U_M^{*}
	\label{equation:core tensor}
\end{equation}
where the  $U_{k}^{*}$ denotes the Hermitian transpose of 
$U_{k}$ for all $k\in [M]$.

Let $(r_1, r_2, \cdots, r_M)$ be a rank tuple subject to 
$r_k\leqslant D_k$ for all $k \in [M]$. When $\mathcal{S}$ and $U_1,\cdots,U_M$ are computed, the approximation of $\mathcal{X}$ is given as follows  
\begin{equation}
	\hat{\mathcal{X}} = \hat{\mathcal{S}}
	\,\times_1\,
	\hat{U}_1  
	\,\times_2\,
	\hat{U}_2
	\cdots 
	\,\times_M\,
	\hat{U}_M
\end{equation}
where $\hat{\mathcal{S}} \in \mathbbm{C}^{r_1\times \cdots \times r_M}$ denotes the sub-core tensor containing the leading $r_1\times \cdots \times r_M$ entries of $\mathcal{S}$ and $\hat{U}_{k} \in \mathbbm{C}^{D_k\times  r_k}$ denotes the sub-matrix containing the leading $r_k$ columns of ${U}_k$ for all $k \in [M]$.

The approximation of the tensor $\mathcal{X}$ can also be written as the following multimode multiplication 
\begin{equation}
	\hat{\mathcal{X}} = \mathcal{X} 
	\,\times_1\, 
	P_1 
	\,\times_2\, 
	P_2 
	\cdots 
	\,\times_M\, 
	P_M 
\end{equation}
where $P_{k} \doteq \hat{U}_{k} \cdot \hat{U}_{k}^{*} \in \mathbbm{C}^{D_k\times D_k}$ denotes the $k$-th idempotent matrix where   $\operatorname{rank} (P_{k}) = r_k$ holds for all $k \in [M]$.

\section{T-scalars and t-algebra}
\label{section:tscalars-and-talgebra}

\subsection{Background and notions}
Many matrix algorithms or tensor algorithms, including HOSVD, can be 
straightforwardly generalized to deeper-order versions with elements in a finite dimensional algebra called t-algebra. We briefly introduce t-algebra, which is finite-dimensional, commutative, and generalizes complex numbers.

The genesis of the t-algebra is Kilmer et al. s’ t-product model \cite{kilmer2013third,kilmer2011factorization}, in which the generalized scalars are fixed-sized order-one real arrays, arranged as the tube fibers of order-three arrays. 
Kilmer et al. don’t explicitly call these tubal fibers ``generalized scalars'' but instead refer to the order-three arrays with these particularly-purposed tubal fibers as just tensors.
However, these ``tensors'' are essentially matrices of generalized scalars. One can add or multiply these generalized scalars (i.e., fixed-sized order-one real arrays) or multiply any generalized scalar with a real or complex number.

Kilmer et al. \cite{kilmer2013third,kilmer2011factorization} chose a circular matrix formulation to 
describe the properties of generalized scalars, matrices, and tensors. In the t-matrix model \cite{liao2020generalized,liao2020general}, Liao and Maybank extended Kilmer et al.’s generalized scalars from real one-way arrays to complex multiway arrays. These generalized multiway scalars, called t-scalars, form a finite-dimensional commutative algebra, generalizing complex numbers and reserving many well-known properties of complex numbers. For example, one can compute the conjugate, the ``real'' and the ``imaginary'' parts of a t-scalar. One can also compute the modulus of a t-scalar, which is nonnegative
and self-conjugate. 
A partial ordering is defined for comparing self-conjugate t-scalars. 
Liao and Maybank 
also define generalized matrices, called t-matrices, over t-scalars. The t-matrix operations are analogous to the corresponding canonical matrix operations, but the underlying operations follow the rules defined for t-scalars.

In this paper, we follow the notations, protocols, and symbols in \cite{liao2020generalized,liao2020general} as much as possible. For example, all indices begin from $1$ rather than $0$, and 
different subscripts other than fonts are used to denote the data types over t-scalars. Some notations and their descriptions are given in Table \ref{tab:table001}. 
For those notations not appearing in Table \ref{tab:table001}, we give their descriptions in context, as necessary.

\newcommand\liaoliangtab[2]{
	\begin{tabular}{ll}
		\hspace{-0.7em}\text{#1} \\
		\hspace{-0.7em}\text{#2} \\
	\end{tabular}
}

\newcommand\mytab[3]{
	\begin{tabular}{ll}
		\hspace{-0.7em}\text{#1} \\
		\hspace{-0.7em}\text{#2} \\
		\hspace{-0.7em}\text{#3} \\	
	\end{tabular}
}

\begin{table}[tbh]
	\centering
	\caption{Some notations defined over t-scalars}
	\vspace{-0.5em}
	\resizebox{0.49\textwidth}{!}{
		\begin{tabular}{|l|l|}
			\hline 
			
			\hline
			
			\hline
			Notations & Descriptions\\
			\hline
			\hline

			$C \equiv \mathbbm{C}^{I_1\times \cdots \times I_N}$     & t-algebra of t-scalars  \\
			\hline
			$X_\mathit{T} \in C$    & 
			a t-scalar in $C$ \\
			\hline
			$Z_{T}, E_{T} \in C$ & zero t-scalar, identity t-scalar\\
			\hline
			$Z_\mathit{TM}, I_\mathit{TM} $ & zero t-matrix, identity t-matrix\\
			\hline
			$X_{T} + Y_{T} \in C $ & t-scalar addition\\
			\hline
			$X_\mathit{T} \circ Y_\mathit{T} \in C$ & t-scalar multiplication\\
			\hline
			$X_\mathit{TM} \in C^{D_1\times D_2} $ &  a t-matrix in $C^{D_1\times D_2}$ \\
			\hline
			$X_\mathit{TM}^{\raisebox{0em}{$*$}}$ & conjugate tranpose of a t-matrix $X_\mathit{TM}$ \\
			\hline        
			$X_\mathit{TM} + Y_\mathit{TM}$ & t-matrix addition\\
			\hline
			$X_\mathit{TM} \circ Y_\mathit{TM}$ & t-matrix multiplication\\
			\hline
			$\raisebox{-0.3em}{$\tilde{X}_\mathit{T} \doteq F(X_\mathit{T})$}  $ & 
			\liaoliangtab{multiway spectral transform of}{a t-scalar $X_\mathit{T}$
			} \\
			\hline
			$\tilde{Y}_\mathit{TM} \doteq F(Y_\mathit{TM})  $ & 
			\liaoliangtab{multiway spectral transform of}
			{a t-matrix $Y_\mathit{TM}$}
			\\
			\hline			
			$(X_\mathit{T})_{i_1,\cdots,i_N} \in \mathbbm{C}$  & \liaoliangtab{$(i_1,\cdots,i_N)$-th complex entry of}{a t-scalar $X_{T} \in C$}  \\
			\hline
			$[X_\mathit{TM}]_{d_1,\,d_2} \in C$     & 
			\liaoliangtab{$(d_1, d_2)$-th t-scalar entry of}{a t-matrix  $X_\mathit{TM}$} 		\\
			\hline
			$S^{nonneg}$ &  a set of nonnegative t-scalars \\
			\hline  
			$\| {X_\mathit{TM}} \|_{\POV, F} \in S^{nonneg}$ & 
			\mytab{t-scalar-valued Frobenius norm of a}{t-matrix $X_{TM}$, i.e., a nonnegative}{t-scalar } 			  \\
			\hline  
			\liaoliangtab{$X_T \leq Y_T$ where}{$X_T, Y_T  \in S^{nonneg}$
			}		 & 
			\liaoliangtab{nonnegative t-scalars satisfying}
			{partial order relations ``$\leq$''
			}		 \\
			\hline
			$\operatorname{rank}_\POV X_\mathit{TM} \geq Z_{T} $  & 
			\mytab{t-scalar-valued rank of a}{t-matrix, i.e., a nonnegative}{t-scalar}   
			\vspace{0.1em}\\
			\hline
			$X_\mathit{GT} \in C^{D_1\times \cdots \times D_M}$ &a g-tensor in 
			$C^{D_1\times \cdots \times D_M}$ \\		
			\hline      
			
			\hline
			
			\hline
		\end{tabular}
	}
	\label{tab:table001}
\end{table}

\subsection{T-scalars in a nutshell} 
T-scalars are fixed-sized multiway arrays of complex numbers. One can add and multiply any pair of these arrays and multiply any of these arrays with a complex number. The t-algebra is the set of these fixed-sized arrays.

Let $C \equiv \mathbbm{C}^{I_1\times \cdots \times I_N}$ be the t-algebra whose elements are $I_1\times \cdots \times I_N$ complex arrays, called t-scalars. 
A t-scalar is defined by a linear spectral transform and the Hadamard product.
Let $F
$ be a bijective linear spectral transform. 
For example, $F$ can be the discrete Fourier transform, discrete cosine transform, or other bijective linear transforms. 
Then, given any t-scalar $X_{T} \in C \equiv \mathbbm{C}^{I_1\times \cdots\times I_N}$ , its transform 
$\tilde{X}_{T} \in \mathbbm{C}^{I_1\times \cdots\times I_N}$
is given by the following multimode multiplication 
\begin{equation}
	\tilde{X}_{T} \doteq F(X_{T}) = X_{T}
	\,\times_1\, W_{1}
	\cdots 
	\,\times_N\, W_{N} 
	\label{equation:transform-tscalar}
\end{equation}
where $W_{k} $ is a full rank matrix in $\mathbbm{C}^{I_k\times I_k}$, 
i.e., the one-way spectral transform 
that defines the spectral transform $F$.

The inverse transform is given by 
\begin{equation}
	{X}_{T} \doteq F^{-1}(\tilde{X}_{T}) = \tilde{X}_{T}
	\,\times_1\, W^{-1}_{1}
	\cdots 
	\,\times_N\, W^{-1}_{N}
	\label{equation:inver-transform-tscalar}
\end{equation}
where, for example, $W_{k}$ can be
the $I_k\times I_k$ Fourier-transform matrix or the cosine-transform matrix such that the $(m_1, m_2)$-th entry of $W_{k}$ is defined 
as follow 
\begin{eqnarray}
\text{Fourier:}\,\,  (W_{k})_{m_1, m_2} = \exp\left[
\frac{-2\pi i}{I_k}(m_1-1)(m_2-1)
\right]\,,  \label{equation:sepctral-transform0001}\\
\text{Cosine:}\,\,\, (W_{k})_{m_1, m_2} = \,\cos\left[
\frac{\pi}{I_k}(m_1-1)(m_2-1/2)
\,\,\right]\,.  \label{equation:sepctral-transform0002}
\end{eqnarray}

Let $X_{T}\circ Y_{T} \in C$ be the multiplication of any two t-scalars 
$X_{T}$ and $Y_{T}$. By the t-scalar multiplication defined in \cite{liao2020generalized}, the following condition holds for all $X_{T}, Y_{T} \in C$,  
\begin{equation} 
	F(X_{T}\circ Y_{T}) = F(X_{T})  \odot F(X_{T})
\label{equation:hadamarm}
\end{equation}
where $\odot$ is the Hadamard product (i.e., entry-wise product) of two complex arrays of the same size.

It is not difficult to show that the t-algebra $C$ is a commutative semisimple algebra. 
A nontrivial t-algebra $C$ is semisimple if it is isomorphic to the direct sum of $K$ copies of the field of complex numbers, where 
\begin{equation}
	K \doteq I_1\cdots I_N
\end{equation} denotes the dimension of $C$.    
When $I_{1} =\cdots = I_N = 1$, $C$ reduces to the field 
$\mathbbm{C}$ \cite{liao2020generalized,liao2020general}.

It is noted by equation (12) that a specific transform defines a corresponding t-scalar multiplication. One can use different transforms to get different t-scalar multiplications.

When the Fourier transform is chosen, the corresponding t-scalar multiplication is called circular convolution \cite{osgood2019lectures}, which is becoming popular in convolutional networks {\cite{krizhevsky2012imagenet,zeiler2014visualizing,
Simonyan15,szegedy2015going,he2016deep,goodfellow2016deep}} in recent years.

\subsection{Matrix representation of t-scalars}

Each t-scalar represents a linear operator from $C$ to $C$ 
since the following condition holds for all 
$X_{T}, A_{T}, B_{T} \in C$ and 
$\alpha, \beta \in \mathbbm{C}$, 
\begin{equation}
X_{T} \circ \big(
\alpha \cdot A_{T} + 
\beta \cdot B_{T}
\big) = 
\alpha \cdot (X_{T} \circ A_{T})
+ 
\beta \cdot (X_{T} \circ B_{T}) \in C.
\end{equation}

Thus, each t-scalar is representable by a square matrix in $C^{K\times K}$ (where $K \doteq \operatorname{dim} C = I_1 \cdots I_N$), and has  $K$ eigenvalues in $\mathbbm{C}$, counting multiplicity.  

Let $E_{T} $ be the identity t-scalar in $C$ such that $E_T \circ X_{T} = X_{T}$ holds for all $X_{T} \in C$. 
By equation (\ref{equation:hadamarm}), the following identity holds for all $X_{T} \in C$,  
\begin{equation}
F(E_T) \odot F(X_{T}) = F(X_{T}) 
\end{equation}
or equivalently, $F(E_T)$ is an array of ones.

A scalar $\lambda \in \mathbbm{C}$ is an eigenvalue of a t-scalar $X_{T} \in C$ if and only if the t-scalar $(X_{T} - \lambda\cdot E_{T})$ is not
multiplicatively invertible, or equivalently, the transform 
\begin{equation}
F(X_{T} - \lambda\cdot E_{T}) \equiv F(X_{T}) - \lambda \cdot F(E_{T}) \in \mathbbm{C}^{I_1\times \cdots \times I_N}
\label{equation:eigen}
\end{equation} 
contains at least one entry equal to zero.

Because $\lambda \cdot F(E_{T})$ is 
an array of $\lambda$'s, it  
is equivalent  to say that $\lambda$ is an eigenvalue of $X_{T}$ if and only if 
$\lambda$ is equal to an entry of $F(X_{T})$. 
Specifically, let the entries of $F(X_T)$ be $F_1(X_{T}),\cdots,F_K(X_{T}) \in \mathbbm{C}$
for any t-scalar $X_{T} \in C$.
Then, these entries of $F(X_{T})$ are the $K$ eigenvalues of $X_{T} \in C$, counting multiplicity.

Because of the set of eigenvalues defines the spectrum of a (bounded) linear operator, and $F$ gives all eigenvalues of any t-scalar $X_{T} \in C$, we call the mapping $F$ a spectral transform.

Furthermore, any t-scalar $X_{T} \in C$ is representable by the diagonal matrix formed by its eigenvalues, namely 
\begin{equation}
	X_{T} \sim \operatorname{diag}\big[F_1(X_T),\cdots,F_K(X_T)
	\big] \in \mathbbm{C}^{K\times K}\; .
\label{equation:spectral-transform}
\end{equation}

If someone uses the Fourier-transform defining $F$, but another uses the cosine transform defining $F$, they usually have different diagonal matrices representing the same raw t-scalar. Since the spectral transform $F$ is always determined prior to the genesis of t-scalars, the eigenvalues and the diagonal matrix representation of any t-scalar $X_{T} \in C$ are uniquely determined. 

It is not difficult to follow that all $K\times K$ diagonal matrices form a commutative semisimple subalgebra of the matrix algebra of all $K\times K$ complex matrices. This subalgebra of all $K\times K$ diagonal complex matrices is isomorphic to the t-algebra $C$. There is a one-to-one mapping between any t-scalar $X_{T}$ and its diagonal matrix representation.

For example, the zero t-scalar $Z_{T}$ is a $I_1\times \cdots \times I_N$ array of zeros, its diagonal matrix representation is the $K\times K$ zero matrix. The matrix representation of the identity t-scalar $E_{T}$ is the $K\times K$ identity matrix. 
Let 
\begin{equation}
	M(X_{T}) \doteq \operatorname{diag}[F_1(X_T),\cdots,F_K(X_T)]
	\label{equation:diagonal-matrix-representation}
\end{equation}
be the diagonal matrix representation of any t-scalar $X_{T}$.
Then, by the algebra isomorphism mentioned previously,  
the following relationships hold for all $X_T, Y_T \in C$ and 
$\alpha, \beta \in \mathbbm{C}$, under the t-scalar operations and the usual matrix operations, 
\begin{equation}
	\begin{aligned}	
		X_{T} \circ Y_{T} &\sim\, M(X_{T}) \cdot  M(Y_{T}) \;,
		 \\
		\alpha \cdot X_{T} +  \beta \cdot Y_{T} &\sim\,  
		\alpha \cdot M(X_{T}) + \beta \cdot M(Y_{T}) \;.
		\\
	\end{aligned}
	\label{tscalar-matrix-equivalence}
\end{equation}

Since each t-scalar is representable by a diagonal matrix, one can use the matrix theory to explain many notions of t-scalars.

For example, the conjugate $X^{*}_{T}$ of a t-scalar $X_{T} \in C$, generalizing the conjugate of a complex number, is representable by the Hermitian transpose of the corresponding 
diagonal matrix. 
More specifically, the following representation holds
\begin{equation}  
X^{*}_{T} \sim 
M(X^{*}_{T}) = 
\operatorname{diag}[\,\overline{F_1(X_T)},\cdots,\overline{F_K(X_T)}\,]
\end{equation}
where $\overline{F_k(X_T)} \in \mathbbm{C}$ denotes the complex conjugate of $F_k(X_T) \in \mathbbm{C}$ for all $k \in [K]$.

A t-scalar $X_{T}$ is called self-conjugate if and only if $X_{T} = X^{*}_{T}$ holds. 
It is equivalent to say that the matrix $M(X_{T}) $ is conjugate symmetric, more specifically, 
\begin{equation} 
M(X_{T}) = M(X^{*}_{T}) = \operatorname{diag}[\overline{F_1(X_T)},\cdots,\overline{F_K(X_T)}] 
\;.
\end{equation}
In other words, all eigenvalues $F_1(X_T),\cdots, F_K(X_T)$ are real numbers.

The standard matrix representation gives a unique perspective to explain many t-scalar-valued
notions in \cite{liao2020generalized,liao2020general}.
We follow this perspective to explain our algorithms and formulations in the following sections.

\subsection{T-matrix and block-diagonal matrix representation}

A t-matrix is a rectangular array of t-scalars. Since t-scalars 
are arrays in $\mathbbm{C}^{I_1\times \cdots \times I_N}$, we choose the underlying data structure of a t-matrix in $C^{D_1\times D_2}$ as an array in $\mathbbm{C}^{I_1\times \cdots \times I_N \times D_1\times D_2}$.  
Because of the underlying multiway array format, many authors call t-matrices tensors \cite{kilmer2011factorization,kilmer2013third}. 
However, they are different from (canonical) tensors with complex entries. 
For example, any t-matrix $Y_\mathit{TM}$  in $C^{D_1\times D_2}$ is representable by a block matrix $M(Y_\mathit{TM})$ in $C^{K D_1\times K D_2}$, each block is a $K\times K$ diagonal matrix,
representing the corresponding t-scalar.

\begin{figure}[htb]
	\centering
	\includegraphics[width = 0.45\textwidth]{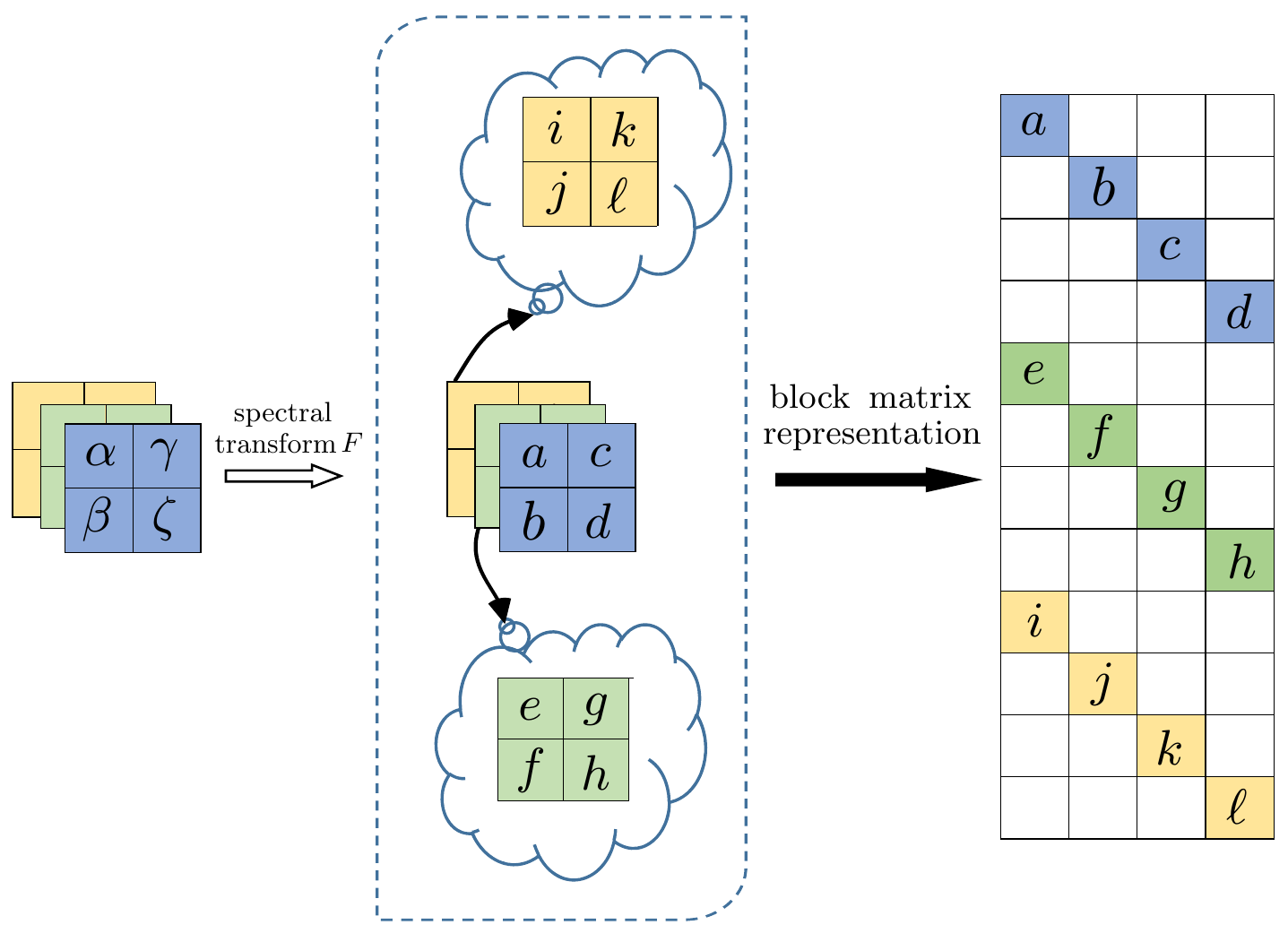} 
	\caption{A diagram of the block-diagonal matrix representation of a t-matrix 
		$Y_\mathit{TM} \in C^{3\times 1} \equiv \mathbbm{C}^{2\times 2\times 3\times 1}$.\;\;   Left: a t-matrix $Y_\mathit{TM}$ (i.e., a t-vector) with $I_1\times  I_2 = 2\times 2$ and $D_1 \times D_2 =3\times 1$. \;\; Middle: transformed t-matrix $F(X_\mathit{TM})$. \;\;Right: a matrix $M(Y_\mathit{TM})$  containing $D_1\times D_2 = 3\times 1$ blocks, each block a $4\times 4$ diagonal matrix.}
	\label{fig:block-matrix-representation}
\end{figure}

Figure \ref{fig:block-matrix-representation} shows the  
block-diagonal matrix representation $M(Y_\mathit{TM})$  of a t-matrix $Y_\mathit{TM}$ in $C^{3\times 1} \equiv 
\mathbbm{C}^{2\times 2\times 3\times 1}$. 
The block-diagonal matrix contains $3\times 1$ blocks, each block a $4\times 4$ diagonal matrix.

To get the matrix representation of a t-matrix $Y_\mathit{TM}$, one needs to compute the eigenvalues of each t-scalar entry, or equivalently, transform $Y_\mathit{TM} \in \mathbbm{C}^{I_1\times \cdots \times I_N\times D_1\times D_2}$ by the $N$-way Fourier transform. The transform of a t-matrix 
$Y_\mathit{TM}$ is obtained by transforming each t-scalar entry of $Y_\mathit{TM}$. The transformed t-matrix is denoted by
$F(Y_\mathit{TM})$.

We choose the first 
$N$ modes of the underlying complex array to accommodate the t-scalar entries of $Y_\mathit{TM}$. 
The formulations of $F(Y_\mathit{TM})$ and $F^{-1}(Y_\mathit{TM})$ are consistent with equations (\ref{equation:transform-tscalar}) and (\ref{equation:inver-transform-tscalar}), namely 
\begin{equation}
	\begin{aligned}
		\tilde{Y}_\mathit{TM} \doteq F(Y_\mathit{TM}) = {Y}_\mathit{TM} \times_1 W_{1} \cdots \times_N W_N \;, \\
		{Y}_\mathit{TM} \doteq  
		F^{-1}(\tilde{Y}_\mathit{TM}) = 
		\tilde{Y}_\mathit{TM} \times_1 W_{1}^{-1} \cdots \times_N W_N^{-1} \;.
	\end{aligned}
	\label{equation:transform-of-a-tmatrix}
\end{equation}

The transformm $F(Y_\mathit{TM})$ is an array containing all eigenvalues of each t-scalar entry of the t-matrix $Y_\mathit{T}$. 
It is easy to convert $F(X_{T}) $ to the block-diagonal matrix representation $M(X_{T}) $ as shown in Figure  \ref{fig:block-matrix-representation}.

Analogous to equation (\ref{tscalar-matrix-equivalence}), the following representations hold for all $\alpha, \beta \in \mathbbm{C}$, and t-matrices $X_\mathit{TM}, Y_\mathit{TM}$ of compatible sizes,  
\begin{equation}
	\begin{aligned}	
		X_\mathit{TM} \circ Y_\mathit{TM} &\sim\, M(X_\mathit{TM}) \cdot  M(Y_\mathit{TM}) \;,\\
		\alpha \cdot X_\mathit{TM} +  \beta \cdot Y_\mathit{TM} &\sim\,  
		\alpha \cdot M(X_\mathit{TM}) + \beta \cdot M(Y_\mathit{TM}) \;.\\				
	\end{aligned}
	\label{tscalar-matrix-equivalence2}
\end{equation}

The block-diagonal matrix representation can help define and explain many notions of the t-matrix model. For example, the determinant of a t-scalar (or a square t-matrix) is equal to the determinant of its matrix representation. The inverse or pseudo-inverse of a t-scalar (or t-matrix) is representable by the inverse or pseudo-inverse of its matrix representation.

\subsection{T-scalar-valued notions}

The matrix representation describes the (canonical) linear properties of t-matrices. Further properties of t-matrices are described in this section. For example, as defined in \cite{liao2020generalized}, one can multiply a t-matrix with a t-scalar. Such an operation, generalizing the multiplication of a complex matrix with a complex number, is an operation for characterizing 
the generalized linear aspect, called the $C$-linearity, of t-matrices.

More specifically, for all $A_{T}, B_{T} \in C$ and $X_\mathit{TM}, Y_\mathit{TM} \in C^{D_1\times D_2}$, the $C$-linear sum  
\begin{equation}
W_\mathit{TM} \doteq A_{T} \circ X_\mathit{TM} + 
B_{T} \circ Y_\mathit{TM}
\end{equation} 
is a t-matrix in $C^{D_1\times D_2}$
such that the following condition holds 
\begin{equation}
[W_\mathit{TM}]_{d_1, d_2} 
= A_{T} \circ [X_\mathit{TM}]_{d_1, d_2} 
+ B_{T} \circ [Y_\mathit{TM}]_{d_1, d_2}
\in C
\label{equation:tscalar-multiplication-tmatrix}
\end{equation}
where $[\cdot]_{d_1, d_2}: X_\mathit{TM} \mapsto 
[X_\mathit{TM}]_{d_1,d_2}
\in C$ denotes the $(d_1, d_2)$-th t-scalar entry of a t-matrix for all $(d_1, d_2) \in [D_1]\times [D_2]$. 

Note that any linear sum of two t-matrices is just a special case a $C$-linear sum of two t-matrices. Specifically, the following identity holds for all 
$\alpha, \beta \in C$ and $X_\mathit{TM}, Y_\mathit{TM} \in C^{D_1\times D_2}$,  
\begin{equation}
\alpha \cdot X_\mathit{TM} + 
\beta \cdot Y_\mathit{TM} \equiv (\alpha \cdot E_{T}) \circ X_\mathit{TM} 
+
(\beta \cdot E_{T}) \circ Y_\mathit{TM} \;.
\label{equation:bi-module-equation}
\end{equation}

The linearity is just a constrained property of the $C$-linearity. Therefore, the introduced matrix representation can not fully characterize $C$-modules.  
For example, a nontrivial t-scalar multiplication $A_{T} \circ X_\mathit{TM}$ can not be represented by the multplication of the matrices $M(A_{T}) \in \mathbbm{C}^{K\times K}$ and $M(X_\mathit{TM}) \in \mathbbm{C}^{KD_1 \times KD_2}$. Such a multiplication of $M(A_{T})$ and $M(X_{T})$ is not defined in classical linear algebra.

The linearity and $C$-linearity require both scalar-valued notions (for the linearity) and t-scalar-valued notions (for the $C$-linearity). Further, a t-scalar-valued notion should be reducible to its scalar-valued counterpart.

Before any detailed examples, let $S^\mathit{nonneg}$ be the commutative semiring of nonnegative t-scalars. 
A t-scalar $X_{T}$ is an element in $S^\mathit{nonneg}$ if and only its matrix representation 
\begin{displaymath}
M(X_{T}) \doteq \operatorname{diag}[F_1(X_T),\cdots,F_K(X_T)]
\end{displaymath}
is positive semidefinite. 
It is equivalent to stating the diagonal entries 
$F_1(X_T),\cdots,F_K(X_T)$ 
are all nonnegative real numbers. 

The above definition is equivalent to the original definition given in \cite{liao2020generalized}. Namely, a t-scalar 
$X_{T}$ is nonnegative if and only if a t-scalar $Y_{T}$ exists such that $X_{T} = Y_{T}\circ Y_{T}^*$ holds.

It shows that all nonnegative t-scalars are also self-conjugate. 
The nonnegativity defines a partial order ``$\leq$'', i.e., a reflexive, antisysmmetric and transitive relationship among self-conjugate t-scalars. More specifically, given any self-conjugate t-scalars $X_{T}$ and $Y_{T}$, the partial order $X_{T} \leq Y_{T}$ holds if and only if  the t-scalar $(Y_{T} - X_{T})$ is nonnegative. 
It also shows that the zero t-scalar $Z_{T}$ is the least element in $S^\mathit{nonneg}$, namely $X_{T} \geq Z_{T}$ holds for all 
nonnegative t-scalar $X_{T}$.

Then, many notions are defined as a nonnegative t-scalar. For example, the generalized modulus $|X_{T}|_\POV$ of a t-scalar $X_{T} \in C$ is a nonnegative t-scalar \cite{liao2020generalized} such that its matrix representation is given by 
\begin{equation}
	|X_{T}|_\POV \sim \operatorname{diag}\big(
	\, 
	|F_1(X_{T}) |, \cdots
	|F_K(X_{T}) | \,
	\big)  \;.
\label{equation:POV-modulus}
\end{equation}
where the subscript ``$\POV$'' 
denotes that the corresponding term is t-scalar-valued, more specifically, a nonnegative t-scalar valued term.

For another example, the trace of a t-scalar $X_{T}$ is the sum of the eigenvalues of $X_{T}$, namely 
\begin{equation}
	\operatorname{trace} X_{T} \doteq F_1(X_{T}) + \cdots +  F_K(X_{T})  
	= \operatorname{trace} M(X_{T})\,.
\end{equation}

Then, one can use the t-scalar-valued modulus $|X_{T}|_t \in S^\mathit{noneng}$ to define the (canonical) Frobenius norm of a t-scalar $X_\mathit{T} \in C$ by 
\begin{equation}
	\begin{aligned}
		\|X_{T}\|_F &\doteq  
		\sqrt{\operatorname{trace} |X_{T}|_\POV^{2}} 
		= \| M(X_T) \|_{F} \\
		& = \sqrt{|F_1(X_T)|^{2} + \cdots + 
			|F_K(X_T)|^{2}}  
		\;\geqslant\; 0\;.
	\end{aligned}
	\label{equation:p-norm}
\end{equation}

If the spectral transform $F$ is not isometric, the norm given by equation (\ref{equation:p-norm}) is not equal to the usual Frobenius norm defined in the spatial domain. Specifically, the following inequality usually holds 
if the spectral transform $F$ is not isometric. 
\begin{equation}  
	\begin{matrix}
		\|X_{T} \|_F \neq 
		\sqrt{
			\sum\nolimits_{i_1,\cdots,i_N} 
			\big| (X_{T})_{i_1,\cdots,i_N} \big|^{2}
		} \;\;.
	\end{matrix}
	\label{equation:non-canonical-frobeius-norm}
\end{equation}

Inequality (\ref{equation:non-canonical-frobeius-norm}) shows that the norm $\|X_{T} \|_{F}$ is defined in the transformed domain 
$F(C)$ rather than the spatial domain $C$.

Following equation (\ref{equation:POV-modulus}), the t-scalar-valued
Frobenius norm of a t-matrix $X_\mathit{TM}$ is a nonnegative t-scalar given by
\begin{equation}   
		\|X_\mathit{TM} \|_{\POV, F} \doteq \sqrt{\sum\nolimits_{d_1, d_2} 
			\big| [X_\mathit{TM}]_{d_1, d_2} \big|_{\POV}^{2} 
		} \in S^\mathit{nonneg} \;.
\end{equation}

Then, one can define the scalar-valued Frobenius norm of the t-matrix $X_\mathit{TM}$ via the t-scalar-valued
Frobenius norm  as follows
\begin{equation}
	\|X_\mathit{TM}\|_{F} \doteq \|M(X_\mathit{TM} ) \|_{F} = 
	\sqrt{ 
		\operatorname{trace} \|X_\mathit{TM} \|_{\POV,F}^{2}
	} \,\geqslant 0 \;.
	\label{equation:canonical-Fnorm-tmatrix}
\end{equation}

Furthermore, if the transform $F$ is not congruent,  
the norm given by equation (\ref{equation:canonical-Fnorm-tmatrix}) is not equal to 
the usual Frobenius norm of 
the underlying (canonical) tensor in $\mathbbm{C}^{I_1\times \cdots \times I_N \times D_1\times D_2 }$.  

It is noted that the cosine transform given by 
equation (\ref{equation:sepctral-transform0002}) is isometric, but the Fourier transform is not.

\section{Pixel neighborhood strategy}
\label{section:neighborhood-strategy}

\subsection{Neighborhood strategy}

The above theory provides a paradigm for analyzing multiway data. 
This paradigm compares favorably with
its canonical counterpart over (canonical) scalars  \cite{2017Ren, liao2020generalized, zhang2017exact,hou2021robust,jiang2019robust,dian2019hyperspectral,cheng2018tensor,yin2018multiview}.

To tackle usual images and exploit the potentials of the t-algebra and the t-matrix model, we use a small neighborhood
of each pixel to obtain a deeper-order representation of a grayscale image.

A matrix can characterize a grayscale image. Let the size of the matrix be $D_1\times D_2$. Each pixel has a pixel neighborhood with size $I_1\times I_2$. If a pixel is at the image's border, at the border of the image, one can pad with ``0'' where necessary to get an $I_{1}\times I_{2}$ neighborhood of the pixel.

Replacing each pixel by its neighborhood of pixels, one has a multiway array representation for a grayscale image, i.e., a t-matrix in $C^{D_1\times D_2} \equiv \mathbbm{C}^{I_1\times I_2\times D_1\times D_2}$. The use of a multiway array improves
the performance of generalized algorithms over t-scalars.

\begin{figure*}[tbh]
	\centering
	\includegraphics[width = 0.88\textwidth]{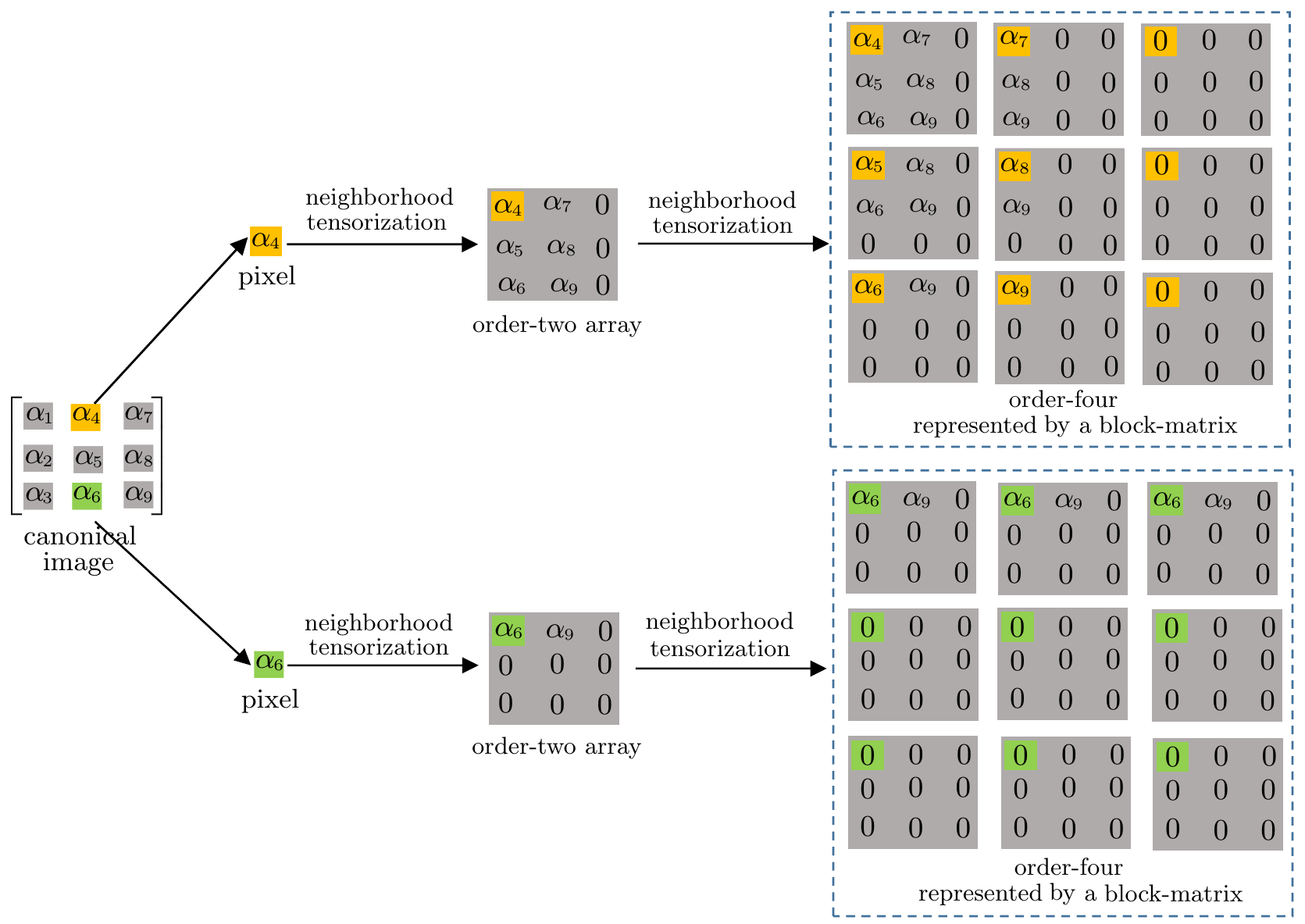}
	\caption{The $3\times 3$ neighborhood of each pixel is used to increase the orders of a grayscale image. Two example pixels $\alpha_4$ and $\alpha_6$ are released by order-two arrays and then by order-four arrays.}
	\label{fig:33neighborhood}
\end{figure*}

Figure \ref{fig:33neighborhood} shows the $3\times 3$ neighborhood strategy, which increases a pixel to an order-two
array and then to an order-four array. In this figure, order-four arrays are illustrated in the form of block matrices. 
Each deeper-order representation of a pixel is specified by a t-scalar.

Given a fixed neighborhood size, there are different neighborhood variants, each with the input pixel at a different position in the neighborhood. 
For example, an 
input pixel can be at
the center or at the corner of its neighborhood.

\begin{figure}[tbh]
	\centering
	\includegraphics[width = 0.48\textwidth]{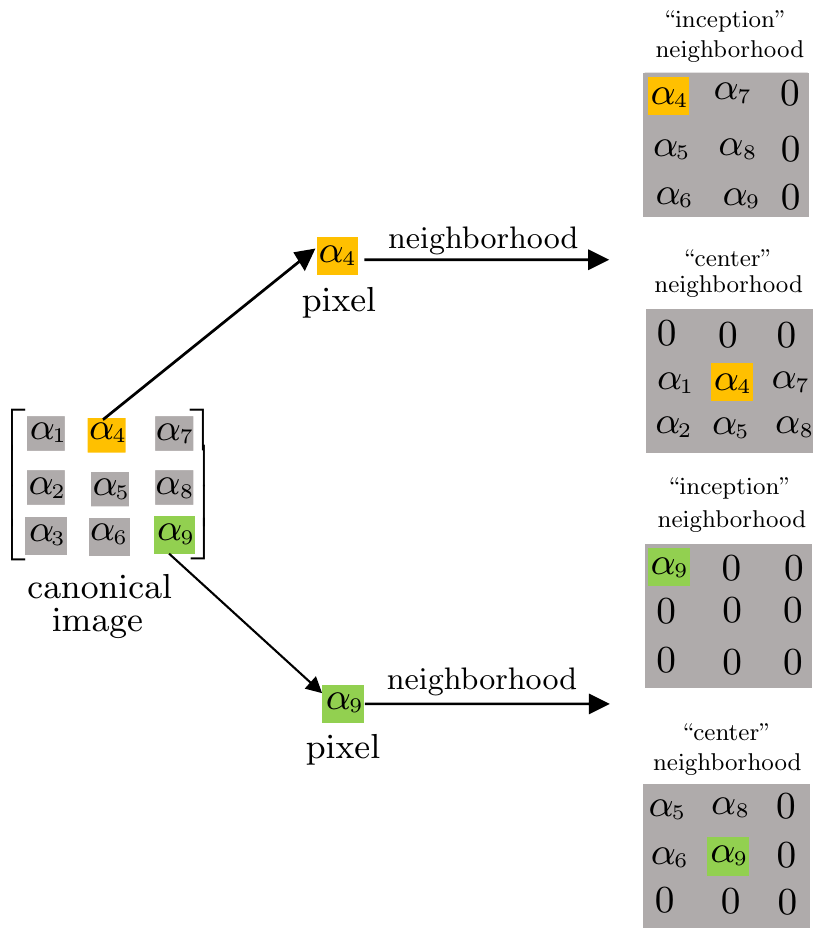}
	\caption{Two types of neighborhoods of a raw pixel, one called ``center'' neighborhood with the pixel at the center, the other called ``inceptive'' neighborhood with the pixel at the inceptive corner}
	\label{fig:inception-neighborhood}
\end{figure}

Figure \ref{fig:inception-neighborhood} shows two types of $3\times 3$ neighborhoods of a raw pixel: the ``center'' neighborhood and the ``inception'' neighborhood. In a ``center'' neighborhood, a raw pixel is at the center of the neighborhood. In an ``inception'' neighborhood, a raw pixel is in the inceptive corner of the neighborhood.  

Our experiments use the inception neighborhood strategy (with the original pixel in the top-left corner) to increase a usual image to a deeper-order version for the following reasons.

If one ues the multiway Fourier transform defined in equation (\ref{equation:sepctral-transform0001}),
it is not difficult to prove that 
the identity t-scalar $E_T$ is a multiway array whose only non-zero entry is equal to 1 and located at the inceptive position (i.e., $i_1 = \cdots =i_N = 1$). Any t-scalar in the form of $\alpha \cdot E_{T}$ (where $\alpha \in \mathbbm{C}$) is called  an inceptive t-scalar in this paper.    
Then, any inception neighborhood, namely, its non-inceptive positions  (i.e., $\exists n \in [N], i_n \neq 1$) valued as $0$, can be represented as an inception t-scalar in the form $\alpha \cdot E_{T}$ where 
$\alpha \in \mathbbm{C}$.

An inception t-scalar $\alpha \cdot E_T$ 
has properties similar to these od the complex number $\alpha$, in that the following conditions hold for all 
$(\alpha \cdot E_T)$ and $(\beta \cdot E_T)$, where $\alpha, \beta \in \mathbbm{C}$,
\begin{equation} 
	\begin{aligned}
		(\alpha \cdot E_{T})^{*} = \bar{\alpha} \cdot E_T &\sim \bar{\alpha}  \;,\\
		(\alpha \cdot E_{T}) \circ (\beta \cdot E_{T}) = (\alpha \cdot \beta) \cdot E_T &\sim (\alpha \cdot \beta) \;,\\
		(\alpha \cdot E_{T}) + (\beta \cdot E_{T}) = (\alpha + \beta) \cdot E_T &\sim (\alpha + \beta)  \;.\\
	\end{aligned} 
\label{equation:equivalent-to-complex-number}
\end{equation}

Equation (\ref{equation:equivalent-to-complex-number}) shows all inception t-scalars form a subalgebra of $C$, isomorphic to the field of complex numbers. If all t-scalars are constrained to be inception t-scalars, an algorithm using inception t-scalars is equivalent to its (canonical) counterpart using the corresponding complex numbers. 

Fortunately, the t-scalars obtained by the strategy of neighborhoods (central neighborhoods or inception neighborhoods) are usually not constrained to inception t-scalars. The experiments in this paper show that generalized algorithms using non-constrained t-scalars outperform their counterparts using complex numbers (or equivalently using merely inception t-scalars). These experiments are discussed in Section \ref{section:005}.

\section{Generalized tensors over t-algebra}
\label{section:THOSVD}

\subsection{Generalized tensors}

Besides t-matrices, one can establish generalized tensors over t-scalars. 
We call them g-tensors (generalized tensors) and denote a g-tensor with $M$-modes by 
$X_\mathit{GT} \in \mathit{C}^{D_1\times \cdots \times D_M} $, where the subscript ``$\mathit{GT}$'' implies a generalized tensor.

G-tensors are the higher-order generalization of t-matrices and the deeper-order generalization of canonical tensors. 
Analogous to a canonical tensor, a g-tensor can be unfolded along its $k$-mode. Specifically, 
the mode-$k$ unfolding of a g-tensor 
$X_\mathit{GT} \in {C}^{D_1\times \cdots \times D_M} $ is a t-matrix 
$X_{\mathit{TM}(k)}$ in ${C}^{D_k\times  D_{k}^{-1} \prod_{m=1}^{M} D_{m} }$
such that the following condition holds for all $j \in \left[D_{k}^{-1} \prod_{m=1}^{M} D_{m} \right]$,  
\begin{equation} 
	[X_\mathit{GT}]_{d_1,\cdots,d_M} = [X_{\mathit{TM}(k)}]_{d_k, j} \in C 
\end{equation}
where $[X_\mathit{GT}]_{d_1,\cdots,d_M}$ denotes the 
the $(d_1,\cdots,d_M)$-th t-scalar entry of  $X_\mathit{GT}$,
$[X_{\mathit{TM}(k)}]_{d_k, j}$ denotes 
the $(d_k, j)$-th t-scalar t-scalar entry of $X_{\mathit{TM}(k)}$
and the index $j$ is determined by equation (\ref{equation:index-correspondence}).

One can also define the generalized mode-$k$ multiplication of a g-tensor $X_\mathit{GT} \in C^{D_1\times \cdots \times D_M}$ 
with a t-matrix $Y_\mathit{TM} \in C^{J\times D_k} $. 
Their mode-$k$ multiplication  
$A_{GT} \doteq X_\mathit{GT} \circ_{k} Y_\mathit{TM} $, 
analogous to the mode-$k$ multiplication of a (canonical) tensor with a (canonical) matrix, is a g-tensor in $C^{D_1\times \cdots  D_{k-1} \times J\times D_{k+1} \times D_M} $ such that the following condition holds  
\begin{equation}
	A_\mathit{GT} = X_\mathit{GT} \circ_{k} Y_\mathit{TM} 	
	\,\Leftrightarrow\, 
	A_{\mathit{TM}(k)} = Y_\mathit{TM} \circ X_{\mathit{TM}(k)}  \,.
\end{equation}

\subsection{Serialization problem: little-endianess versus big-endianess}
\label{section:little-endian}

A convenient approach is to organize a t-matrix as a multimode array of complex numbers, some modes for the t-scalar entries and the others for the t-matrix. The principle of this approach also applies to a g-tensor. 

For example, Kilmer et al. \cite{kilmer2013third,kilmer2011factorization} organize a t-matrix as a three-mode array, the first two modes for its rows and columns, the last for t-scalar entries. 
The modes chosen for t-scalars depend on the underlying structure of the data. Our implementation chooses the leading modes for t-scalars and differs from Kilmer et al.'s implementation. Our implementation uses the little-endian protocol (reverse lexicographic ordering), and Kilmer et al. use the big-endian protocol (co-lexicographic ordering) \cite{dolgov2014alternating,liu2021tensors}. 

Although the Endian wars 
have never been quenched \cite{cohen1981holy}, the little-endian protocol 
offers the consistency of the transform of a t-matrix (or g-tensor) given by equation (\ref{equation:transform-of-a-tmatrix}) and the transformation of a t-scalar given by equation (\ref{equation:transform-tscalar}).
Using the little-endian protocol, we call the leading modes for t-scalars the ``deeper orders'' and the rest the ``higher orders.''

To this end, we denote 
\begin{equation}
	\mathbbm{C}^{(I_1\times \cdots I_N) \times [D_1\times \cdots \times D_M]} \equiv C^{D_1\times \cdots \times D_M} 
\end{equation} 
to emphasize the first $N$ modes of the underlying array are for t-scalars. 

Then, given a g-tensor $X_\mathit{GT} \in C^{D_1\times \cdots \times D_M}$, 
the underlying array of the 
generalized mode-$k$ unfolding $X_{\mathit{TM}(k)}$ is 
an element in $\mathbbm{C}^{(I_1\times \cdots \times I_N)\times [D_k \times D_k^{-1} \prod_{m=1}^{M}D_m] }$.

\subsection{THOSVD: t-algebra based HOSVD}
THOSVD (T-algebra based HOSVD) is a generalization of HOSVD (Higher Order Singular Value Decomposition) over t-scalars. 
Specifically, given a g-tensor $X_\mathit{GT} \in C^{D_1\times \cdots \times D_M}$, the g-tensor can be written by the following generalized multimode multiplication 
\begin{equation}
	X_\mathit{GT} =  S_\mathit{GT} 
	\circ_{1}~ U_{\mathit{TM}, 1}
	\circ_{2}~ U_{\mathit{TM}, 2} \cdots
	\circ_{M}~ U_{\mathit{TM}, M}
	\label{equation:THOSVD}
\end{equation} 
where $U_{\mathit{TM}, m}\in {C}^{D_M\times D_M}$ denotes the orthogonal t-matrix whose columns are the left singular t-vectors of the generalized mode-$k$ unfolding 
of $X_\mathit{GT}$ for all $m \in [M]$ and $S_\mathit{GT}\in {C}^{D_1\times \cdots \times D_M}$  is the core g-tensor given by
\begin{equation} 
	S_\mathit{GT}= {X}_{GT} \,
	\circ_{1}\, U^{*}_{\mathit{TM}, 1}
	\,\circ_{2}\, U^{*}_{\mathit{TM}, 2} \cdots
	\,\circ_{M}\, U^{*}_{\mathit{TM}, M}
	\label{equation:core g-tensor}
\end{equation}
where $U_{\mathit{TM},m}^{*}$ denotes the conjugate transpose of the t-matrix $U_{\mathit{TM},m}$  such that the matrix representation $
M(U_{\mathit{TM},m}^{*})$ is the Hermitian transpose of 
$M(U_{\mathit{TM},m})$ for all $m \in [M]$.

Let $\hat{U}_{\mathit{TM},k} \in C^{D_k\times r_k } $ be the t-matrix containing the first $r_k$ columns of the t-matrix $U_{\mathit{TM},k} \in C^{D_k\times D_k}$. 
Then, one can define the following idempotent t-matrix 
\begin{equation} 
	P_{\mathit{TM},k} = \hat{U}_{\mathit{TM}, k} \,\circ\, 
	\hat{U}^{*}_{\mathit{TM}, k} \,\in\, C^{D_k\times D_k} \;\;.
	\label{equation:idempotent-tmatrix}
\end{equation}
such that $M(P_{\mathit{TM},k})$ is an idempotent matrix in $\mathbbm{C}^{KD_k \times KD_k}$ for all $k \in [M]$. 

The t-scalar-valued rank of the t-matrix $P_{\mathit{TM},k}$ is given by
\begin{equation}
	\operatorname{rank}_\POV P_{\mathit{TM},k} = r_k \cdot E_T  
	\geq Z_{T} \;.
\end{equation}

The t-scalar-valued
rank of a t-matrix is defined in the appendix of \cite{liao2020generalized}. Interested readers are referred to this definition for more details.\footnote{The definition in \cite{liao2020generalized} uses the notation $\operatorname{rank}(\cdot)$ rather than $\operatorname{rank}_\POV(\cdot)$ to denote the 
t-scalar-valued rank.} 

One can also define the (canonical) rank of a t-matrix $X_\mathit{TM} \in C^{D_1\times D_2} $ as follows. 
\begin{equation} 
	\operatorname{rank} X_\mathit{TM} \doteq \operatorname{rank}  M(X_\mathit{TM})
	\in [0, K\cdot \min(D_1, D_2) ] \;.
	\label{equation:rank}
\end{equation}
and the following condition holds for all t-matrices $X_\mathit{TM}$,
\begin{equation} 
	\operatorname{rank} X_\mathit{TM} = \operatorname{trace}( 
	\operatorname{rank}_\POV X_\mathit{TM} 
	)
	\;.
\end{equation}

Thus, the canonical rank of the idempotent matrix $P_{\mathit{TM}, k}$ is given as follows 
\begin{equation}
	\operatorname{rank} P_{\mathit{TM}, k} \doteq \operatorname{rank} M(P_{\mathit{TM},k}) = \operatorname{trace} (r_k \cdot E_{T}) = K\cdot r_{k} 
	\;.
\end{equation}

When the condition $r_k = D_k$ holds, the t-matrix $P_{\mathit{TM},k}$ is the identity t-matrix in $C^{D_k\times D_k}$ and $M(P_{\mathit{TM},k})$ is the identity matrix in $\mathbbm{C}^{KD_k\times KD_k}$.

Then, given the positive integers $r_1,\cdots, r_M$, one has the idempotent t-matrices $P_{\mathit{TM},1},\cdots,P_{\mathit{TM},M}$ and can project the g-tensor $X_\mathit{GT}$ on a low-dimensional sub-module of $C^{D_1\times \cdots \times D_M}$ as follows
\begin{equation}
	\hat{X}_\mathit{GT} = {X}_\mathit{GT} 
	\,\circ_1\, P_{\mathit{TM},1}
	\cdots 
	\,\circ_M\, P_{\mathit{TM},M} 
	\;.
	\label{equation:r1-rN-approximation}
\end{equation}

Let the low-dimensional sub-module be $\mathcal{M}$. The 
t-scalar-valued dimension of $\mathcal{M}$ is given by   
\begin{equation}
	\begin{aligned}
		\operatorname{dim}_\POV \mathcal{M} &= (r_1 r_2 \cdots r_M) \cdot E_{T} \\
		&\leq (D_1  D_2\cdots D_M) \cdot E_{T} 
		\;. 
		\label{equation:postive-operator-valued-rank}
	\end{aligned}
\end{equation}
or equivalently, the (canonical) dimension of $\mathcal{M}$ is given by   
\begin{equation}
	\begin{aligned}
		\operatorname{dim} \mathcal{M} &= \operatorname{trace}
		\big(
		\operatorname{dim}_\POV \mathcal{M} 
		\big) = K\cdot (r_1 r_2\cdots r_M)  \\
		& \leqslant  K\cdot (D_1D_2\cdots D_M) 
		\;. 
	\end{aligned}
\end{equation}

The t-scalar-valued rank given by equation (\ref{equation:postive-operator-valued-rank}) is an inception t-scalar. The 
t-scalar-valued rank of a t-matrix has not to be just an inception t-scalar. However, we only discuss inception-tscalar-valued rank in this paper.

\subsection{Local alternating optimization}

The optimal $(r_1,\cdots,r_M)$-approximation of a g-tensor $X_\mathit{GT}$ is  a solution $\hat{X}_\mathit{GT}$ 
minimizing the following function under the partial order ``$\leq$'' on nonnegative t-scalars. 
\begin{equation}
	\{
	P^{*}_{\mathit{TM},k} 
	\}_{k=1}^{M}  = \mathop{\operatorname{argmin}}\nolimits_{
		\{P_{\mathit{TM},k} \}_{k=1}^{M}
	}
	\|X_\mathit{GT} - \hat{X}_\mathit{GT} \|_{\POV, F}
	\label{equation:argmini}
\end{equation}
where $\hat{Y}_\mathit{GT}$, the $(r_1,\cdots,r_M)$-approximation, is given by  
equation (\ref{equation:r1-rN-approximation}) and  
$\|\cdot \|_{\POV, F}: Y_\mathit{GT} \mapsto \| Y_\mathit{GT} \|_{\POV, F} $ denotes the 
t-scalar-valued
Frobenius norm of a g-tensor $Y_\mathit{GT}$ such that the following condition holds for all g-tensors $Y_\mathit{GT}$ 
\begin{equation}
	\| Y_\mathit{GT} \|_{\POV, F} = \| Y_{\mathit{TM}(1)} \|_{\POV, F} 
	\in S^\mathit{nonneg} \;. 
\end{equation}

To characterize the linearity rather than $C$-linearity, one can define the following (canonical) norm
\begin{equation}
	\|Y_\mathit{GT}\|_{F} \doteq \|M(Y_{{TM}(1)} ) \|_{F}
	= \sqrt{
		\operatorname{trace} \|Y_\mathit{GT}\|_{\POV, F}^{2} 
	}  \,\geqslant\, 0\;.
\end{equation}

Then, equation (\ref{equation:argmini}) is equivalent to the following (canonical) minimization problem under the total order ``$\leqslant$'' on nonnegative real numbers
\begin{equation}
	\{P^{*}_{\mathit{TM},k} \}_{k=1}^{M}  = \mathop{\operatorname{argmin}}\nolimits_{
		\{P_{\mathit{TM},k} \}_{k=1}^{M}
	}
	\|X_\mathit{GT} - \hat{X}_\mathit{GT} \|_{F} \;.
	\label{equation:canonical-argmini}
\end{equation}

Analogous to the well-received conclusions \cite{de2000best,1980Principal}, there is no known deterministic global optimizer of equation (1). However, the ALS (Alternating Least Squares) algorithm called HOOI (Higher Order Orthogonal Iteration) can be generalized over t-scalars for locally minimizing equation (\ref{equation:canonical-argmini}) or equation (\ref{equation:argmini}).
The generalized algorithm, called THOOI (t-algebra based HOOI), is a straightforward generalization of HOOI as shown in Algorithm \ref{algorithm:THOSVD}. 
When $I_1 =\cdots = I_N = 1$, THOOI reduces to HOOI.

\renewcommand{\algorithmicrequire}{\textbf{Input:}}
\renewcommand\algorithmicensure {\textbf{Output:} }
\begin{algorithm}[tbh]
	\caption{THOOI: t-algebra based HOOI }
	\begin{algorithmic}[1]
		\REQUIRE a g-tensor  
		$\mathit{X}_{GT}\in {C}^{D_1\times \cdots \times D_M} 
		$ and $M$ initial idempotent t-matrices  computed as in equation   (\ref{equation:idempotent-tmatrix}) by THOSVD such that 
		$\operatorname{rank}_\POV P_{\mathit{TM}, k} = r_k \cdot E_{T}$  holds for 
		all $k \in [M]$. 		
		\ENSURE the locally optimized g-tensor approximation $\hat{X}_\mathit{GT}$.  
		\REPEAT 
		\FORALL {$k \in [M]$}
		\STATE $P_{\mathit{TM},k} \leftarrow I_\mathit{TM} \in C^{D_m\times D_m}$  \;\; {\footnotesize\it $\triangleleft$ set the $k$-th idempotent t-matrix to the identity t-matrix}
		\STATE $\hat{X}_\mathit{GT} \leftarrow X_\mathit{GT} 
		\,\circ_1\, P_{\mathit{TM}, 1}
		\cdots
		\,\circ_M\, P_{\mathit{TM}, M}
		$
		\STATE  $P_{\mathit{TM},k} \leftarrow \hat{U}_{\mathit{TM},k} \circ \hat{U}^{*}_{\mathit{TM},k}$ where $\hat{U}_{\mathit{TM},k}$ contains the leading $r_k$ left singular t-vectors of the generalized mode-$k$ unfolding of $\hat{X}_\mathit{GT}$. 
		\ENDFOR	
		\UNTIL{convergence is obtained or the maximum number of iterations is reached}
		\RETURN the updated $P_{\mathit{TM},1},\cdots,P_{\mathit{TM},M}$ and the 
		optimized approximation $\hat{X}_\mathit{GT}$
	\end{algorithmic}
	\label{algorithm:THOSVD}
\end{algorithm}

\subsection{Unification of principal component analysis algorithms}

A wide range of PCA-based algorithms can be unified by HOSVD \cite{sheehan2007higher} or, more generally, THOSVD. Specifically, 
given $Q$ multiway arrays 
$\mathcal{X}_1, \cdots,\mathcal{X}_{Q}$
in $\mathbbm{C}^{(I_1\times \cdots \times I_N) \times [D_1\times \cdots \times D_M]}$,    
one can concatenate them as a multiway 
array 
\begin{equation}
	\mathcal{X} \overset{\mathit{cat}}{=} (\mathcal{X}_1,\cdots,\mathcal{X}_Q) 
\end{equation} 
in $\mathbbm{C}^{(I_1\times \cdots \times I_N) \times [D_1\times \cdots \times D_M \times Q]}$. 

If $\mathcal{X}_1 +\cdots+ \mathcal{X}_Q $ is not zero, one can use $\mathcal{X}_{k} \leftarrow (\mathcal{X}_k - \bar{\mathcal{X}})$
to update each sample array, where  
$
\bar{\mathcal{X}} \doteq {Q}^{-1}\cdot (\mathcal{X}_1+\cdots+ \mathcal{X}_Q)
$ denotes the mean of the $Q$ arrays.

To have a consistent notation, 
we re-denote the multiway array by $\mathcal{X} \in \mathbbm{C}^{D_1\times \cdots \times D_{J}} $ where $J = N + M +1$. 
Then, a wide range of PCA-based algorithms on $\mathcal{X}$ can be unified by HOSVD in the following multilinear multiplication,  
\begin{equation}
	\hat{\mathcal{X}} = 
	\mathcal{X} 
	\,\times_1 P_1
	\cdots
	\,\times_{J} P_{J}
	\label{equation:reconstruction}
\end{equation}
where $P_i \doteq \hat{U}_{i} \hat{U}_{i}^{*} \in \mathbbm{C}^{D_i\times D_i} $ denotes the $i$-th idempotent matrix 
and $\hat{U}_{i} \in \mathbbm{C}^{D_i\times r_i} $ denotes the matrix containing the $r_i$
leading left singular vectors of the mode-$i$ unfolding of $\mathcal{X}$ for all $i \in [J]$.  If $r_i = D_i$, the 
idempotent matrix $P_i$ is full rank, or equivalently, the identity matrix $I_{D_i}$ in $\mathbbm{C}^{D_i\times D_i}$. 

If $P_i$ is rank deficient for all $i < J$ and the only full rank idemponent matrix 
is $P_J$, then the approximation $\hat{\mathcal{X}} \overset{\mathit{cat}}{=} (\hat{\mathcal{X}}_1,\cdots,\hat{\mathcal{X}}_{Q})$ is the PCA reconstruction of 
${\mathcal{X}} \overset{cat}{=} (\mathcal{X}_1,\cdots, \mathcal{X}_{Q})$.

When $P_2$ is rank deficient, and the other idempotent matrices are of full rank, the arrays $\hat{\mathcal{X}}_1,\cdots,\hat{\mathcal{X}}_Q $ are the so-called 2DPCA reconstructions of 
$\mathcal{X}_1,\cdots,\mathcal{X}_Q $. 
When only $P_1$ and $P_2$ are  
rank-deficient, $\hat{\mathcal{X}}_1,\cdots,\hat{\mathcal{X}}_Q $ are the reconstructions of the so-called (2D)\textsuperscript{2}PCA approximations proposed by Zhang and Zhou et al \cite{zhang20052d}. 

Further, when $P_1,\cdots,P_{J-1}$ are 
rank-deficient and $P_J$ is of full rank, equation (\ref{equation:reconstruction}) gives the reconstruction of raw MPCA (multilinear PCA), a multiway generalization of PCA, 2DPCA and (2D)\textsuperscript{2}PCA proposed by Lu et al\cite{lu2008mpca}.

However, since PCA and 2DPCA are already generalized over t-scalars, see TPCA and T-2DPCA in \cite{liao2020generalized,2017Ren,liao2021TPCA}, we are interested in unifying them via THOSVD and generalizing Lu's MPCA over t-scalars. 
This unification provides a panoramic view of seemingly unrelated algorithms and might help in the design of new feature extractors.

The unification is straightforward. Given $Q$ multiway arrays 
$\mathcal{X}_{1},\cdots, \mathcal{X}_{Q} \in \mathbbm{C}^{(I_1\times \cdots\times I_N)\times [D_1\times \cdots \times D_M]}$ such that their sum is zero, these arrays 
can be characterized by $Q$ g-tensors in $C^{D_1\times \cdots \times D_M}$
over t-scalars in $\mathbbm{C}^{I_1\times \cdots \times I_N}$.

The $Q$ g-tensors can be concatenated and characterized by a higher-order g-tensor 
$\mathit{X}_\mathit{GT}$ in ${C}^{D_1\times \cdots\times D_M\times D_{M+1}} $ (where $D_{M+1} \doteq Q$).  Then, similar to equation (\ref{equation:r1-rN-approximation}), the g-tensor $X_\mathit{GT}$ is approximated  
by the following generalized multimode multiplication 
\begin{equation} 
	\hat{X}_\mathit{GT} = {X}_\mathit{GT}  
	\,\circ_1 P_{\mathit{TM},\,1}
	\cdots
	\,\circ_{(M+1)} P_{\mathit{TM},\,(M+1)}  \,.
	\label{equation:THOSVD-approximation}
\end{equation}

If $P_{\mathit{TM}, (M+1)}$ is rank-deficient, and the other t-matrices are of full rank, the approximation $\hat{X}_\mathit{GT} \overset{cat}{=}  (\hat{X}_{1},\cdots,\hat{X}_{Q}) $ gives the TPCA approximations of the $Q$ g-tensors ${X}_{1},\cdots,{X}_{Q}$ in $C^{D_1\times \cdots \times D_M}$.

When the only rank-deficient idempotent t-matrix is $P_{\mathit{TM},\,2}$, equation (\ref{equation:THOSVD-approximation}) yields the approximations by T-2DPCA. T-2DPCA generalizes Yang's 2DPCA and can be found with more details in \cite{liao2020generalized}.  
When the only rank-deficient idempotent t-matrices are $P_{\mathit{TM},\,1}$ and $P_{\mathit{TM},\,2}$, equation (\ref{equation:THOSVD-approximation}) generalizes the result given by (2D)\textsuperscript{2}PCA. 
Furthermore, when the only full-rank idempotent t-matrix is $P_{(M+1)}$, equation (\ref{equation:THOSVD-approximation}) generalizes the results by MPCA. We call the generalized algorithm TMPCA (t-algebra based MPC).

If there are two or more rank-deficient idempotent t-matrices, equation (\ref{equation:THOSVD-approximation}) 
cannot solve the minimization described in
equation (\ref{equation:canonical-argmini}). However, one can use Algorithm \ref{algorithm:THOSVD} to optimize the results given by equation (\ref{equation:THOSVD-approximation}).

\section{Experimental Verifications}
\label{section:005}

In this section, two types of experiments are described. One type is the ``vertical'' experiments in which we use HOSVD to approximate multispectral images. We also use THOSVD with the same parameters to approximate the deeper-order versions of these multispectral images.

The other type is the ``horizontal'' experiment in which THOSVD approximates raw multispectral images without the pixel neighborhood strategy, as HOSVD does on the same raw images.  

All the deeper-order versions of the raw images are generated using the ``inception'' neighborhood strategy. The transform used for defining the t-scalar multiplication is based on the Fourier transform. The results of experiments using the ``center'' neighbor strategy and the cosine-based transform are omitted since they are similar to those mentioned above.

\subsection{``Vertical'' experiments}
Two public multispectral images are used in the ``vertical'' experiments of image approximation using HOSVD and THOSVD.
One is the ``food'' image, the other the ``superballs'' image
\footnote{\url{https://www1.cs.columbia.edu/CAVE/databases/multispectral/}}. 

Each image has $31$ channels and is a $512\times 512 \times 31$ array of real numbers 
that defines an order-three canonical tensor. The tensor is approximated using HOSVD.

To exploit the potential of THOSVD in the ``vertical'' experiments, 
it is necessary to extend the raw images to their deeper-order versions.
With the neighborhood strategy, each grayscale of a multispectral image is increased to a $3\times 3$ array of grayscales. 
Thus, each $512\times 512\times 31$ multispectral image is increased to 
a $(3\times 3)\times [512\times 512\times 31]$ array of real numbers, i.e., 
$I_1 \times I_2 = 3\times 3$ and  
$D_1 \times D_2 \times D_3 = 512 \times 512\times 31$.

\begin{figure}[tbh]
	\centering
	\includegraphics[width = 0.48\textwidth]{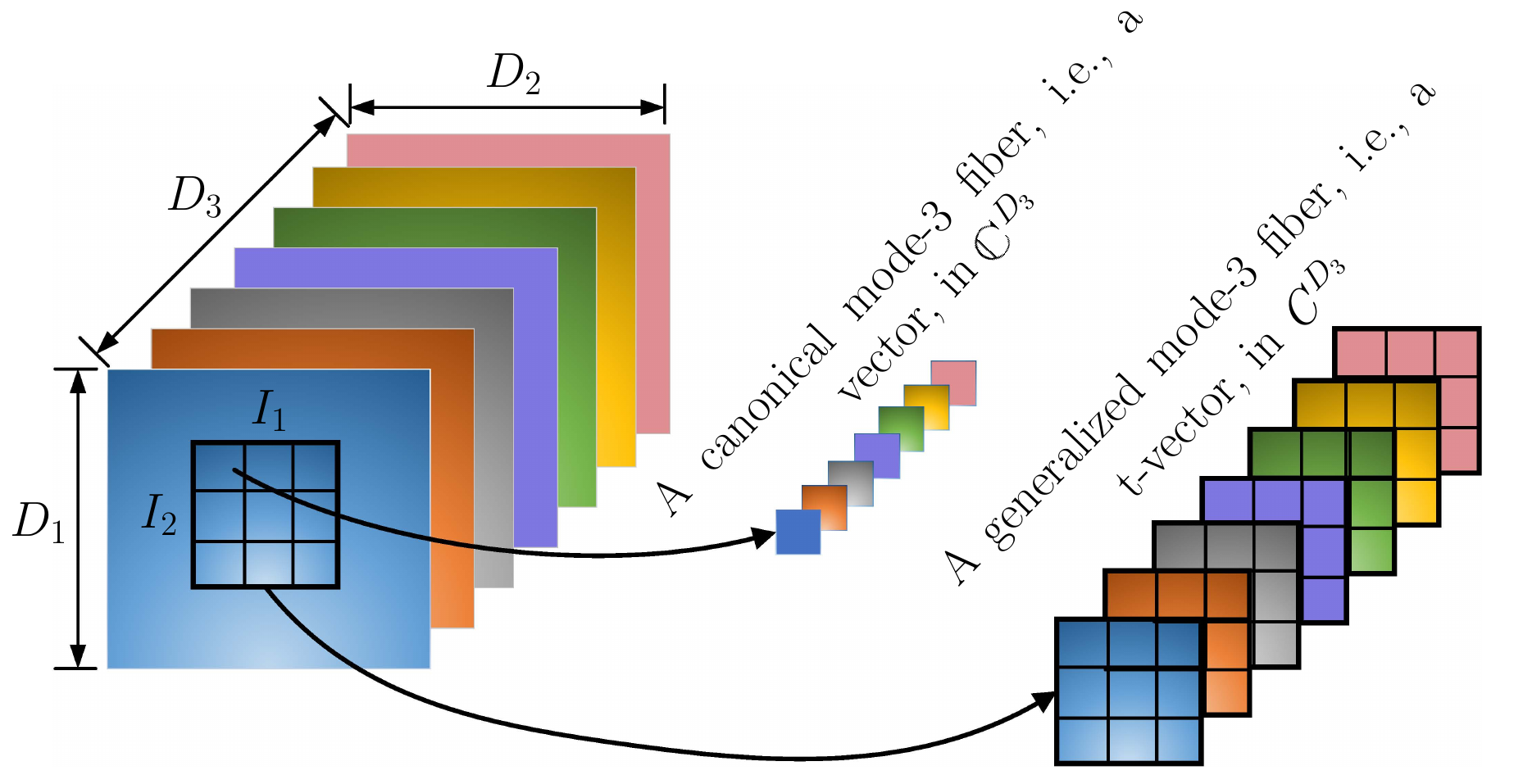}
	\caption{A multispectral image (with $I_1\times I_2$ inception pixel neighborhoods) to the order-five version, where each mode-$3$ fiber, i.e., a vector in $\mathbbm{C}^{D_3}$, is increased to a t-vector in $C^{D_3}\equiv \mathbbm{C}^{(I_1\times I_2)\times [D_3]}$. }
	\label{fig:g-tensor}
\end{figure}

Figure \ref{fig:g-tensor} 
shows the increase of a
$D_1\times D_2\times D_3$ tensor (multispectral image) to its order-five version. Each mode-$3$ fiber (i.e., a vector) in $\mathbbm{C}^{D_3}$, is increased to 
a generalized mode-$3$ fiber (i.e., a t-vector) in $C^{D_3} \equiv \mathbbm{C}^{(I_1\times I_2) \times [D_3]}$.

Using the neighborhood strategy for a second time, one can increase a multispectral image to the size $(3\times 3\times 3\times 3) \times [512\times 512\times 31]$.

Different variants with different t-scalar settings are used in this ``vertical'' experiment. Their settings are listed in Table \ref{tab:table002}. 
Among these methods, HOSVD is the special case of THOSVD 
with t-scalars reduced to order-zero (i.e., canonical scalars). THOSVD uses order-two t-scalars, and THOSVD-A 
uses order-four t-scalars. If THOOI is used, 
the optimized versions are denoted by
by THOSVD-OP and THOSVD-A-OP.

\newcommand\liaoliang[2]{
	$
	\begin{matrix}
		\text{#1} \\
		\text{#2} \\
	\end{matrix}
	$
}
\begin{table}[htb]
\centering
\caption{HOSVD/THOSVD variants}
	\vspace{-0.2em}
	\resizebox{0.5\textwidth}{!}{
		\centering
		\small
		\begin{tabular}{|c|c|c|c|}
			\hline 
			
			\hline
			
			\hline
			{method}
			& \liaoliang{\raisebox{-0.1em}{t-scalar size}}{\raisebox{0.1em}{
					({\scriptsize $I_1\times \cdots\times I_N$})}}  & \liaoliang{\raisebox{-0.1em}{t-scalar}}{\raisebox{0.1em}{order}} & \liaoliangtab{~if locally}{optimized?}\\
			
			\hline    
			\hline
			\raisebox{-0.1em}{HOSVD} & \raisebox{-0.1em}{--} & \raisebox{-0.1em}{$0$} 
			&No	\\	
			\hline
			\raisebox{-0.1em}{HOSVD-OP} & \raisebox{-0.1em}{--} & \raisebox{-0.1em}{$0$} 
			&Yes 	\\
			\hline    
			\hline			
			\raisebox{-0.1em}{THOSVD}  & \raisebox{-0.1em}{$3\times3$}  & \raisebox{-0.1em}{$2$}
			&No  
			\\
			\hline
			\raisebox{-0.1em}{THOSVD-A} & \raisebox{-0.1em}{$3\times3 \times 3\times 3$} & \raisebox{-0.1em}{$4$} 
			& No\\
			\hline
			\hline		
			\raisebox{-0.1em}{THOSVD-OP}  & \raisebox{-0.1em}{$3\times3$}  & \raisebox{-0.1em}{$2$}
			& Yes\\
			\hline 
			\raisebox{-0.1em}{THOSVD-A-OP} & \raisebox{-0.1em}{$3\times3 \times 3\times 3$} & \raisebox{-0.1em}{$4$} 
			& Yes\\
			\hline
			
			\hline
			
			\hline
		\end{tabular}
	}		
	\label{tab:table002}
\end{table}

The peak signal-to-noise ratio (PSNR) is used to measure the quality of the approximations. 
The larger the PSNR, the better the approximation. 
For a multiway array $\mathcal{X}$ and its approximation 
$\hat{\mathcal{X}} $, the PSNR is given by
\begin{equation} 
	\mathit{PSNR} = 20 \log_{10} \frac{\mathit{MAX} \cdot \sqrt{N^{entry}} }    
	{\|\mathcal{X} - \hat{\mathcal{X}}\|_F}
	\label{equation:PSNR}
\end{equation}
where $N^\mathit{entry}$ denotes the number of scalar entries of the given array, 
$\|\cdot\|_{F}: \mathcal{X} \mapsto 
\sqrt{
	\sum\nolimits_{d_1,\cdots,d_M} 
	|(\mathcal{X} )_{d_1,\cdots,d_M}|^{2} 
}$ denotes the Frobenius norm of a canonical tensor in $\mathbb{R}^{D_1\times \cdots \times D_M}$, and $\mathit{MAX}$ denotes the maximum possible value of the scalar entries.

\begin{figure*}[tbh]
	\includegraphics[width = \textwidth]{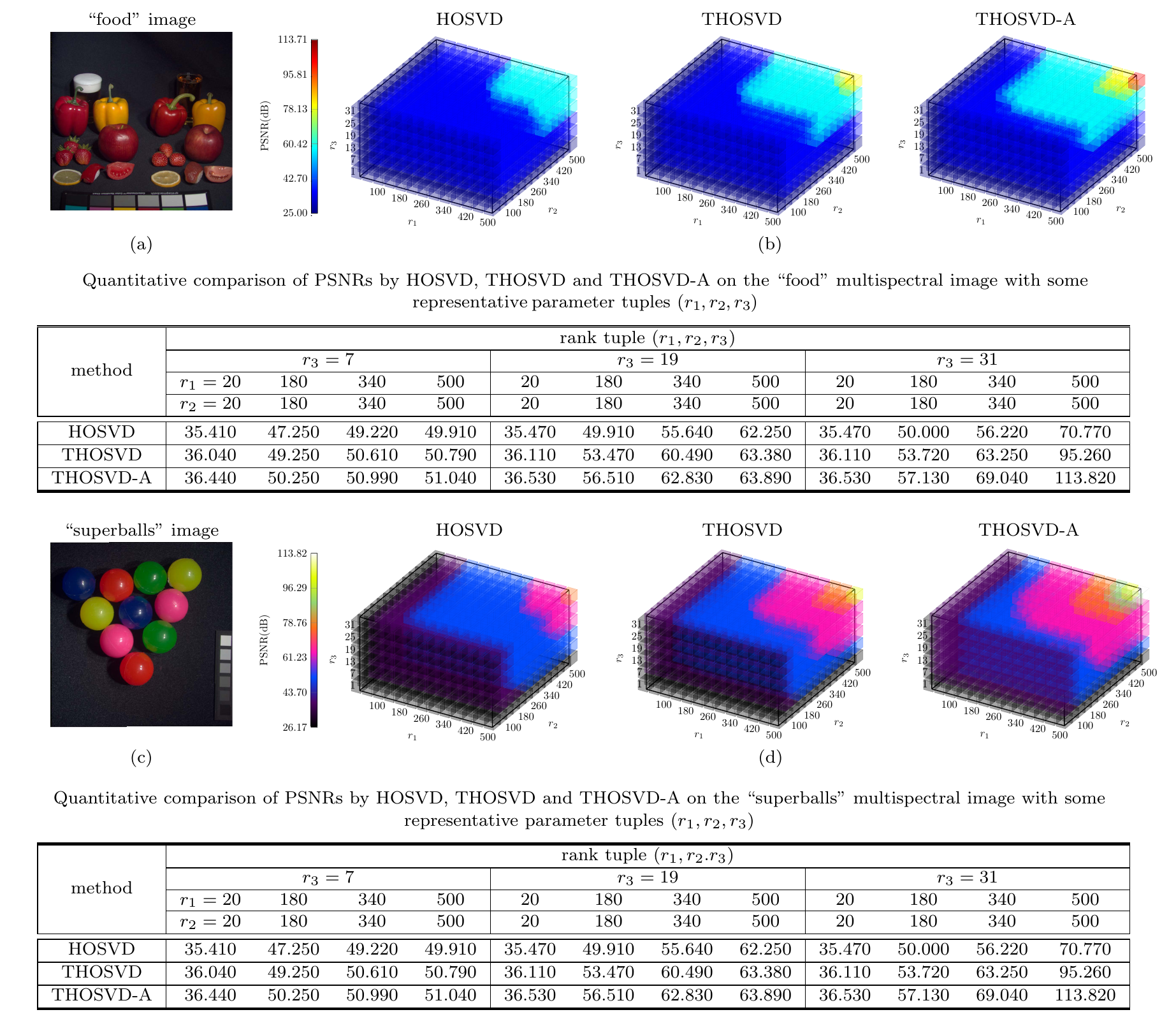}
	\caption{A ``vertical'' comparison of the PSNRs by HOSVD, THOSVD and THOSVD-A with different parameter tuples $(r_1, r_2, r_3)$ on the ``food'' and ``superballs'' multispectral image\;\; 
		(a) RGB version of the ``food'' iamge\;\; 
		(b) Results of different algorithms on the ``food'' image  \;\;
		(c) RGB version of the ``superballs'' image\;\; 
		(d) Results of different algorithms on the ``superballs'' image  \;\;
	}
	\label{fig:foodANDsuperballsApproximation}
\end{figure*}

Figure \ref{fig:foodANDsuperballsApproximation} shows the heatmaps of PSNRs using HOSVD, THOSVD, THOSVD-A with different values of rank tuple $(r_1, r_2, r_3)$ 
for approximating the ``food'' multispectral image and the ``superballs'' multispectral image. 

When $r_1$, $r_2$, and $r_3$ are increased, the PSNR 
of the approximation
also increases. 
The generalized methods THOSVD and THOSVD-A consistently outperform the canonical counterpart HOSVD with the same parameters.  
The tabulated results in Figure \ref{fig:foodANDsuperballsApproximation} provide a quantitative comparison.  
For example, when $(r_1, r_2, r_3) = (500,500, 31)$,  
THOSVD outperforms HOSVD by $22.9$ dB (i.e., $98.55$ dB - $75.65$ dB) and THOSVD-A outperforms HOSVD by $38.06$ dB (i.e., $113.71$ dB - $75.65$ dB) on the ``food'' image.
The approximation results on the ``superballs'' image are consistent with the conclusion drawn from the results on the ``superballs'' image.

Figure \ref{fig:foodANDsuperballsApproximation} 
also shows that the generalized algorithm
using deeper-order t-scalars outperforms 
the version of the algorithm
using shallower-order t-scalars. 
In both the ``food'' image and the ``superballs'' image, THOSVD-A (using order-four t-scalars) 
outperforms THOSVD (using order-two t-scalars). 
Also note, HOSVD is a special case of THOSVD with the 
``shallowest order'' t-scalars (i.e., order-zero t-scalars). HOSVD yields the lowest PSNRs in the ``vertical'' experiments.

\subsection{``Vertical'' experiments using local optimization}

\begin{figure*}[tb]
	\centering
	\begin{tabular}{c}
		\includegraphics[width=0.9\textwidth]{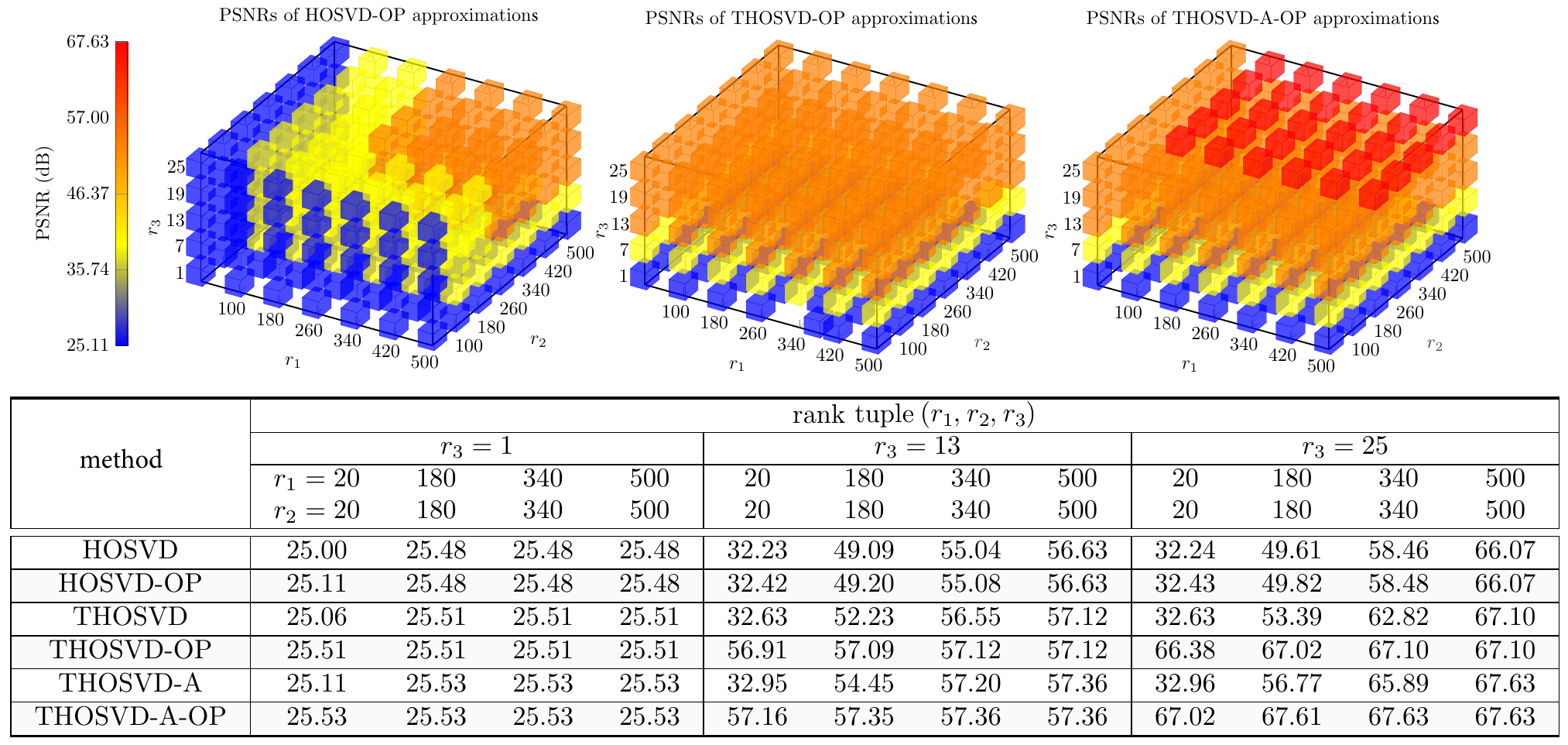} \\
		{\small (a)} results on the ``food'' image  \vspace{0.9em}\\
		\includegraphics[width=0.9\textwidth]{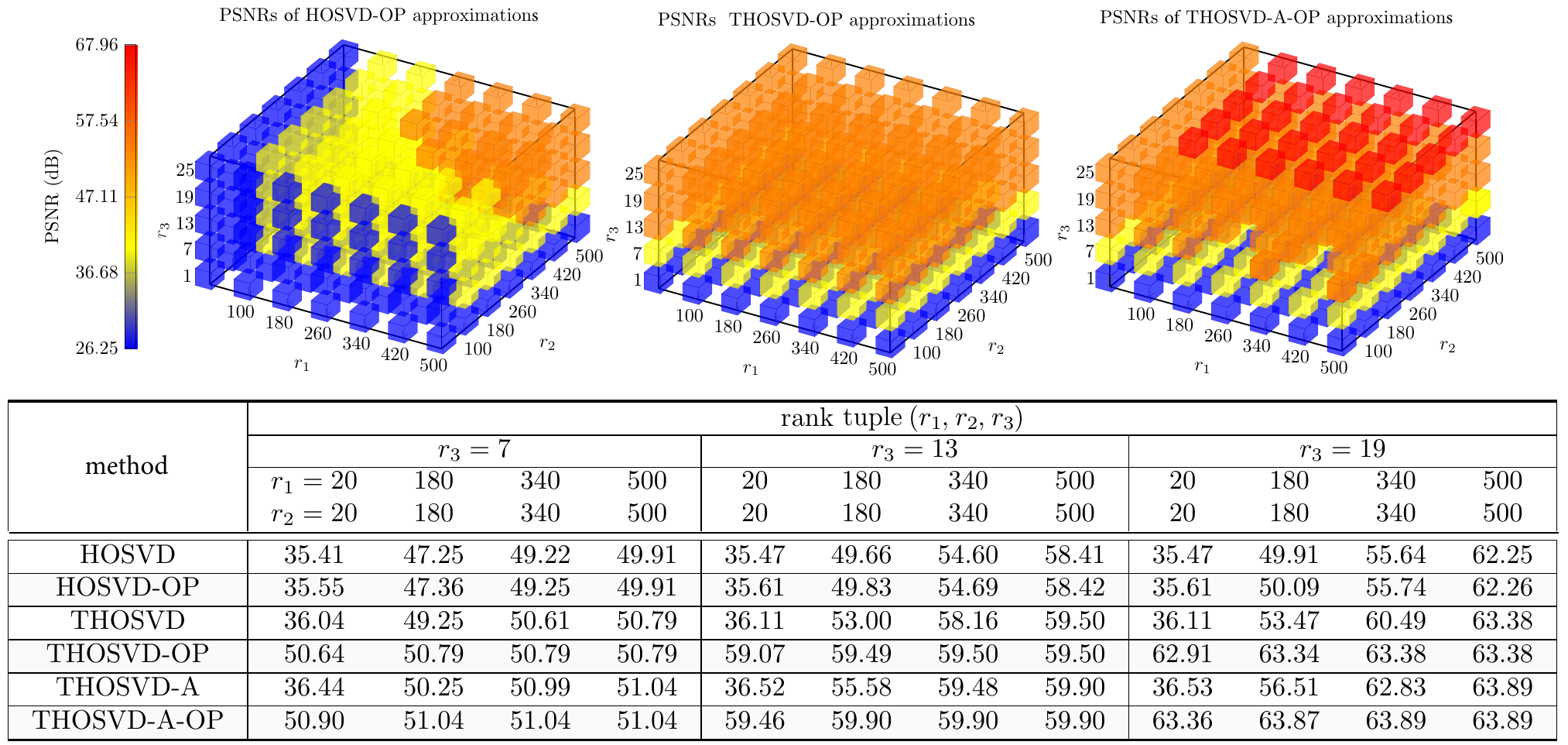} \\
		{\small (b) results on the ``superballs'' image} 
	\end{tabular}
	\caption{A comparison of the results by optimized algorithms HOSVD-OP, THOSVD-OP, and THOSVD-A-OP on the ``food'' image and the ``superballs'' image~~~(a) results on the ``food'' image
		~~~(b) results on the ``superballs'' image}
	\label{fig:comaprison-with-alternating-optimization}
\end{figure*}

A comparison of the approximations by optimized algorithms 
HOSVD-OP, THOSVD-OP, and THOSVD-A-OP is given in Figure \ref{fig:comaprison-with-alternating-optimization}. 
Figure \ref{fig:comaprison-with-alternating-optimization}(a) shows the results on the ``food'' image and \ref{fig:comaprison-with-alternating-optimization}(b) shows the results on the ``superballs'' image.

The results are consistent with the observation that generalized algorithms  outperform their canonical counterpart, and generalized algorithms using high-order t-scalars outperform those using low-order t-scalars.  
Some representative quantitative results are also tabulated in Figure \ref{fig:comaprison-with-alternating-optimization}.  For example, on the ``food'' image, when $(r_1,r_2, r_3)  = (180, 180, 25)$, THOSVD-OP outperforms HOSVD-OP by $17.20$ dB (i.e., $67.02$ dB $–$ $49.82$ dB), and THOSVD-A-OP outperforms THOSVD-OP by $0.59$ dB (i.e., $67.61$ dB $–$ $67.02$ dB).

The tabulated results in Figure \ref{fig:comaprison-with-alternating-optimization} also include the PSNRs yielded by HOSVD, THOSVD, and THOSVD-A (i.e., the algorithms without the alternating optimization). 
It shows that the optimized algorithms HOSVD-OP, THOSVD-OP, and THOSVD-A-OP, respectively outperform non-optimized algorithms HOSVD, THOSVD, and THOSVD-A.

\subsection{``Horizontal'' experiments on USC-SIPI images}

It is of interest to compare HOSVD and THOSVD by approximating the same raw images. For example, a color image is a three-way array of real numbers. It can be regarded as either an order-three tensor of real numbers or a t-matrix of order-one t-scalars.  

When regarded as an order-three tensor, a color image can be approximated by HOSVD along each of the three modes. When regarded as a t-matrix, a color image can be approximated by THOSVD along each of the two generalized modes. 
For example, given a $256\times 256\times 3$ color image, one can model it as a $256\times 256$ t-matrix 
(using the big-endian protocal), with each t-scalar entry being an order-one array containing three real numbers.
To approximate such a t-matrix, THOSVD requires two parameters $r_1 \in [512]$ and $r_2 \in [512]$.

Although the quality of a t-matrix approximation is determined by $\min(r_1, r_2)$, we give the results of THOSVD using the two parameters $r_1$ and $r_2$ to compare with the results of HOSVD using the same parameters.

Also worthy of notice is, with two parameters $r_1$ and $r_2$, THOSVD is equivalent to TSVD introduced in \cite{liao2020generalized} or Kilmer et al.'s t-SVD introduced in \cite{kilmer2011factorization}.   
In the following, we use TSVD to interpret the approximating process of THOSVD. 
More specifically, let $X_\mathit{TM}$ be a t-matrix 
in $\in C^{D_1\times D_2}$ (where $D_1 = D_2 =512$). The compact TSVD of $X_\mathit{TM}$ is given by 
\begin{equation}
	X_\mathit{TM} = U_\mathit{TM} \circ S_\mathit{TM} \circ V_\mathit{TM}^{*}
\label{equation:55}
\end{equation} 
where the t-matrix $U_\mathit{TM} \in \mathit{C}^{D_1\times \min(D_1, D_2) }$ contains the left singular t-vectors of the t-matrix $X_\mathit{TM}$, and the t-matrix 
$V\in C^{D_2\times \min(D_1, D_2)}$ contains the right singular t-vectors of $X_\mathit{TM}$ such that the following condition holds 
\begin{displaymath}
	X_\mathit{TM}^{*} \circ X_\mathit{TM} = 
	V_\mathit{TM}^{*} \circ V_\mathit{TM} = I_\mathit{TM} 
\end{displaymath} where $I_\mathit{TM}$ denotes the 
$\min(D_1, D_2)\times \min(D_1, D_2)$ identity t-matrix, or equivalently, the matrix $M(I_\mathit{TM})$ is the 
(canonical) identity matrix in $\mathbbm{C}^{K \min(D_1, D_2) \times K \min(D_1, D_2)}$.

The t-matrix $S_\mathit{TM} $ denotes the 
$\min(D_1, D_2)\times \min(D_1, D_2)$ 
diagonal t-matrix whose $(i,j)$-th t-scalar entry is given by 
\begin{displaymath}
	[S_\mathit{TM}]_{i,j} = \delta_{i,j} \cdot \beta_{T, i}	
\end{displaymath}
where $\delta_{i,j}$ denotes the Kronecker delta and 
$\beta_{T, i} $ is the $i$-th leading singular t-scalar value of $X_{TM}$ such that the following partial order of nonnegative t-scalars holds for all $i \in [\min(D_1, D_2) - 1]$,  
\begin{displaymath}
	\beta_{T, i}  \geq  
	\beta_{T, (i +1)}  \geq Z_{T} \;.
\end{displaymath}

Then, given two parameters 
$r_1$ and $r_2$, one can approximate the t-matrix $X_\mathit{TM}$ as follows  
\begin{equation} 	
	\hat{X}_\mathit{TM} = 
	U_\mathit{TM} 
	\circ \hat{S}_\mathit{TM} 
	\circ 
	{V}_\mathit{TM}^{*} \;.
	\label{equation:TSVD-approximation}
\end{equation}
where $\hat{S}_\mathit{TM}$ denotes 
the low-rank approximation of $S_\mathit{TM}$ with the 
parameters $r_1, r_2 \in [\min(D_1, D_2)]$ 
such that the following condition holds
\begin{equation} 
	[\hat{S}_\mathit{TM}]_{i, j} = 
	\left\{
	\begin{aligned}
		&[{S}_\mathit{TM}]_{i, j} & \text{if}\, i \in [r_1] \,\text{and}\, j \in [r_2] \\
		&Z_{T} & \text{otherwise}
	\end{aligned}
	\right. \;\;.
\end{equation}
One can find more details of TSVD in \cite{liao2020generalized} and  \cite{kilmer2011factorization}.

To have a fair comparison with the same parameters, $r_1$ and $r_2$, HOSVD approximates a given multispectral image $\mathcal{X} \in \mathbbm{C}^{D_1\times D_2\times D_3}$ as follows 
\begin{displaymath}
	\hat{\mathcal{X}} = \mathcal{X} \,\times_1\, 
	(\hat{U}_1\, \hat{U}_1^{*} ) \,\times_2\,
	(\hat{U}_2\, \hat{U}_2^{*} )  
	\,\times_3\, I_{D_3}
\end{displaymath}
where $I_{D_3}$  
denotes the $D_3\times D_3$ identity matrix, 
$\hat{U}_1 \in \mathbbm{C}^{D_1\times r_1}$
and 
$\hat{U}_2 \in \mathbbm{C}^{D_2\times r_2}$
are semi-orthogonal, the columns of $\hat{U}_1$ are the $r_1$ leading left singular vectors of the mode-$1$ unfolding of $\mathcal{X}$, and the columns of $\hat{U}_2$ are the $r_2$ leading left singular vectors of the mode-$2$ unfolding of $\mathcal{X}$.

Two public USC-SIPI RGB images are used in the ``horizontal'' experiment. One is the ``house'' image, the other is the ``airplane'' image \footnote{\url{https://sipi.usc.edu/database/}}. The ``house'' image size is $256\times 256\times 3$, and the ``airplane'' image size is $512\times 512\times 3$.

\begin{figure*}[tb]
	\centering
	\begin{tabular}{c}
		\includegraphics[width=0.9\textwidth]{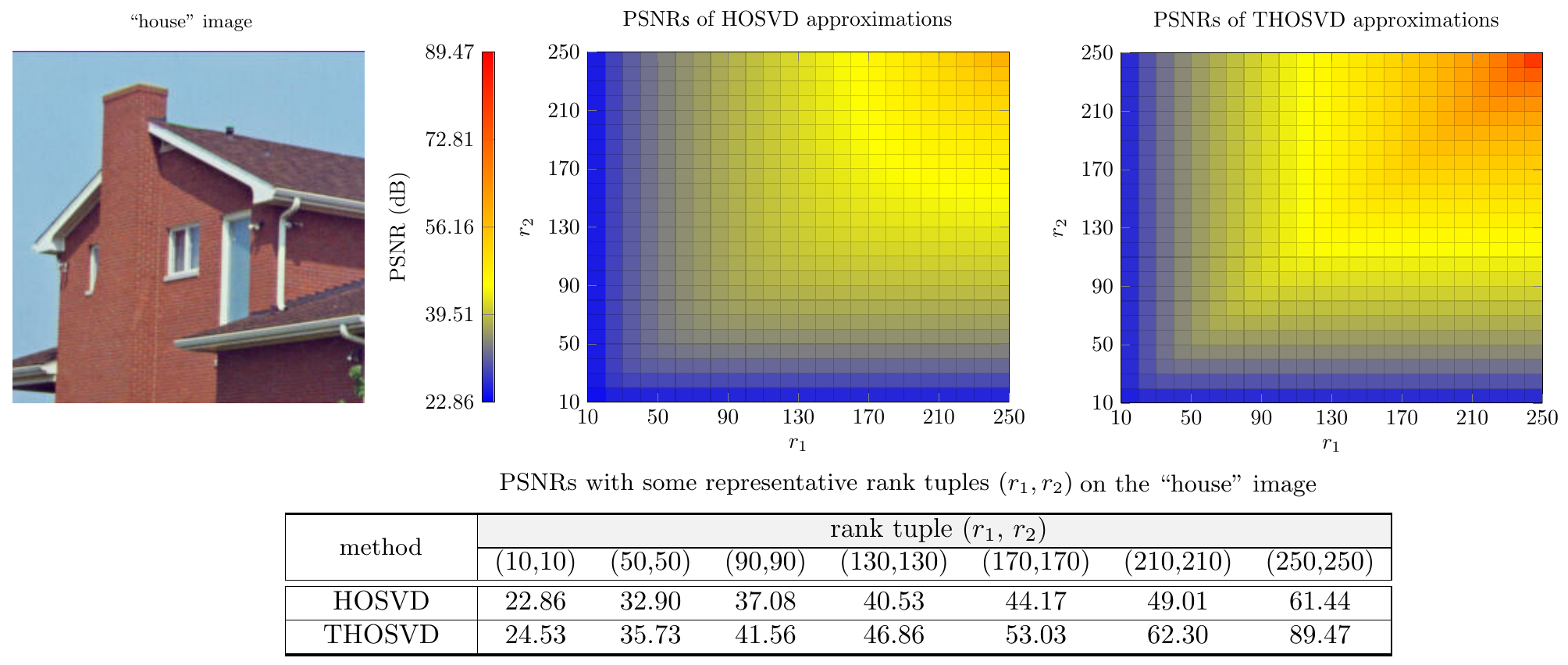} \\
		(a) results on the ``house'' image  \vspace{0.5em}\\
		\includegraphics[width=0.9\textwidth]{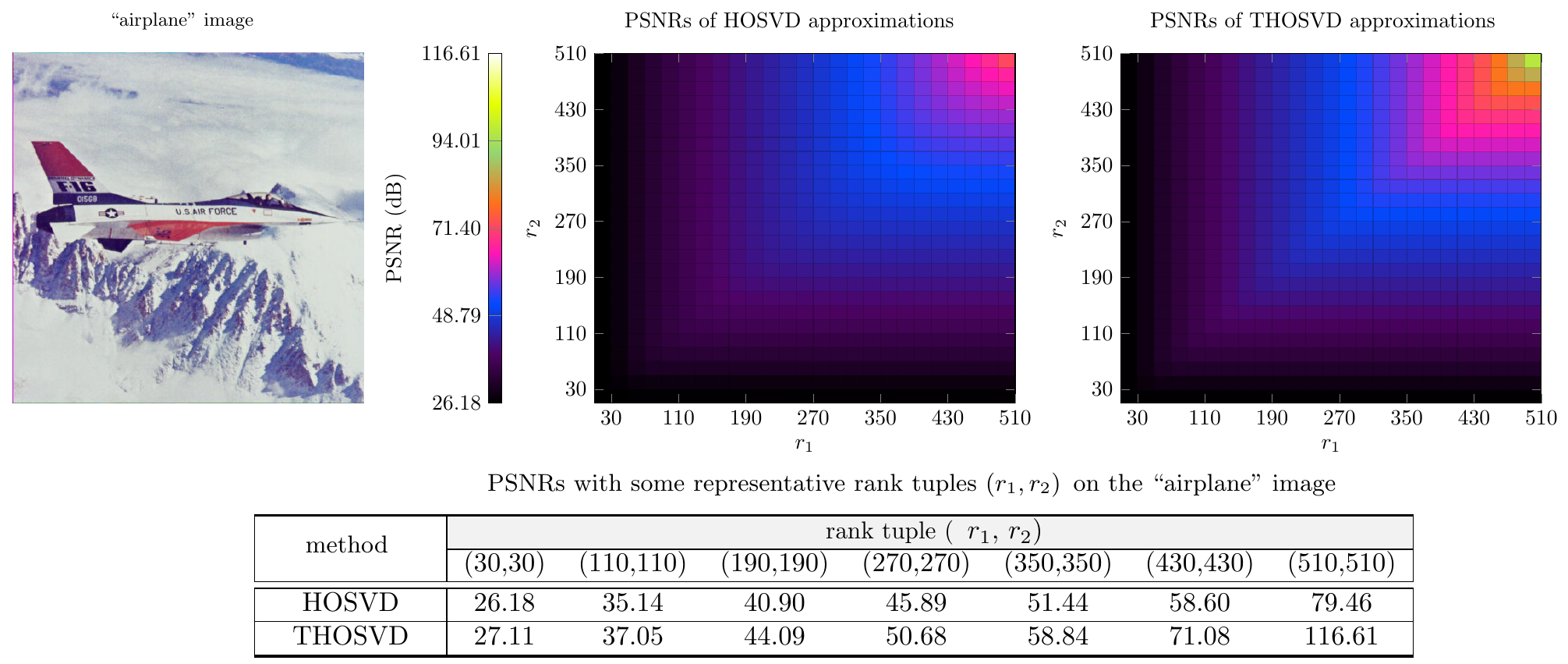} \\
		(b) results on the ``airplane'' image\\
	\end{tabular}
	\caption{A comparison of the PSNRs (dB) by HOSVD and THOSVD on the ``house'' image and the ``airplane'' image}
	\label{fig:comaprison-with-house-airplane}
\end{figure*}

Figure \ref{fig:comaprison-with-house-airplane} shows the PSNRs by HOSVD and THOSVD for approximating the ``house'' image and the ``airplane'' image. The PSNRs for some representative rank tuples $(r_1, r_2)$ are also tabulated in the figure. It shows that, on the same raw data, THOSVD consistently outperforms HOSVD. For example, when $(r_1, r_2) = (210, 210)$, on the ``house'' image, THOSVD outperforms HOSVD by $13.29$ dB ($62.30$ dB $-$ $49.01$ dB).

\subsection{Unification of principal component analysis algorithms}

This section unifies a wide range of principal component analysis algorithms and compares their performances for approximating images in vertical and horizontal experiments.

The CIFA-10 image set is used in the experiments. 
The image set contains $60000$ RGB images. Each image is a $32\times 32\times 3$ array  
\footnote{\url{https://www.cs.toronto.edu/~kriz/cifar.html}}. 
We use the first $1000$ images of this image set for the experiments.   
The $1000$ CIFA images can be organized as a $32 \times 32\times 3\times 1000$ array of real numbers, namely $3000$ grayscale images of size $32\times 32$.

Using the $3\times 3$ neighborhood strategy as shown in Figure \ref{fig:33neighborhood}, one can increase each grayscale image from order-two to order-four, yielding a g-tensor of four modes in 
$C^{32\times 32 \times 3\times 1000}$, i.e., a 
$(3\times 3) \times [32\times 32 \times 3\times 1000]$ 
underlying array of real numbers for the selected CIFA-10 images. 

By reusing the neighborhoods as shown in Figure 5, one can further have a g-tensor in  $C^{32\times 32\times 3\times 1000}$ of t-scalars in $\mathbbm{C}^{3\times 3\times 3\times 3}$, i.e., an underlying order-eight array of real numbers.

No matter what size the underlying array is, the g-tensor has four modes, two modes for the rows and columns of underlying images, one mode for the RGB changes, and the fourth mode for image samples. Let each image sample be subtracted from their mean. Then, by using THOSVD on the row and column modes, one generalizes (2D)\textsuperscript{2}PCA \cite{zhang20052d}. We call the generalized algorithm T-(2D)\textsuperscript{2}PCA.

Note that MPCA is a higher-order generalization of (2D)\textsuperscript{2}PCA, while T-(2D)\textsuperscript{2}PCA is a deeper-order generalization of (2D)\textsuperscript{2}PCA, totally different from MPCA. Further, if one of the two modes mentioned above is left unhandled, T-(2D)\textsuperscript{2}PCA reduces to T-2DPCA, generalizing Yang’s 2DPCA \cite{yang2004two}. Interested readers are referred to \cite{liao2020generalized} for a performance comparison between 2DPCA and T-2DPCA.

\begin{figure*}[htbp]
	\centering
	\begin{tabular}{c}
		\hspace{6em}
		Some CIFA-10 RGB images \\
		\hspace{6em}
		\includegraphics[width=0.8\textwidth]{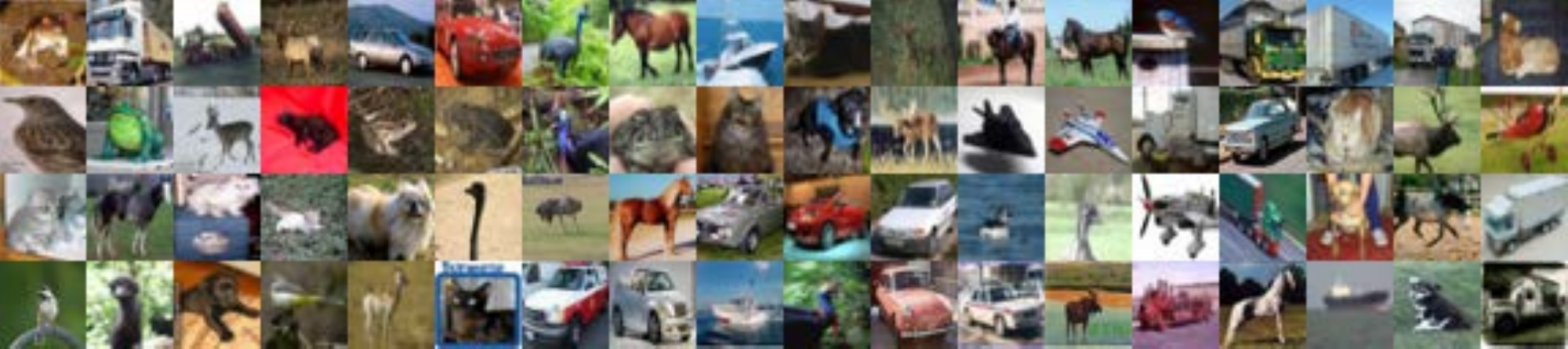} \vspace{0.9em}\\
		\includegraphics[width=0.9\textwidth]{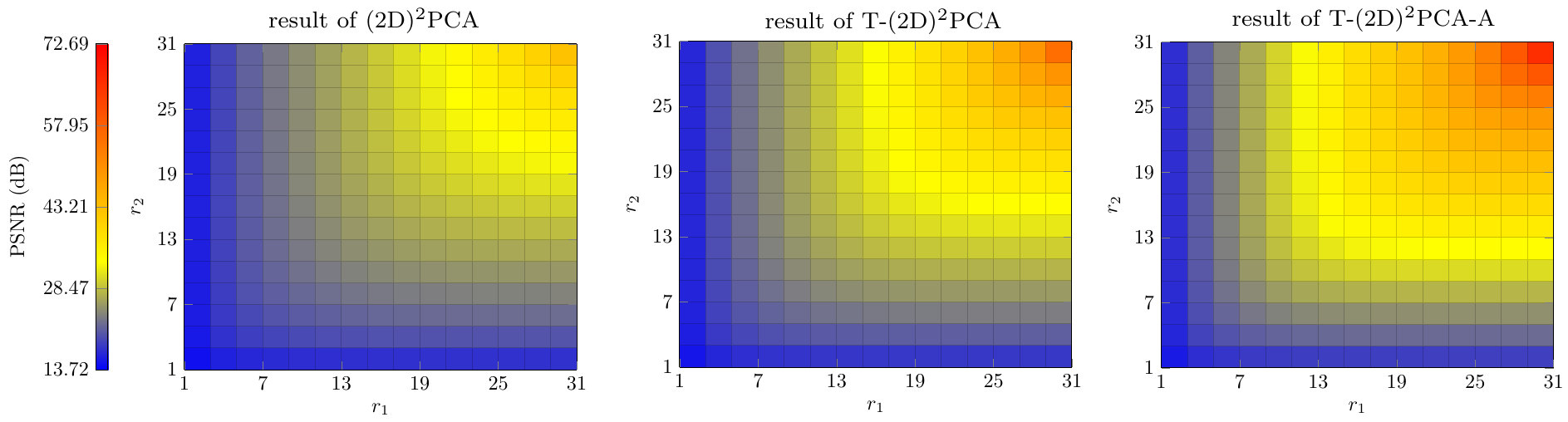} \\
		\hspace{0.1em} PSNRs by (2D)\textsuperscript{2}PCA, 
		T-(2D)\textsuperscript{2}PCA
		and T-(2D)\textsuperscript{2}PCA-A with some representative parameter 
		tuple $(r_1, r_2)$
		\\
		\hspace{0.8em}\includegraphics[width=0.9\textwidth]{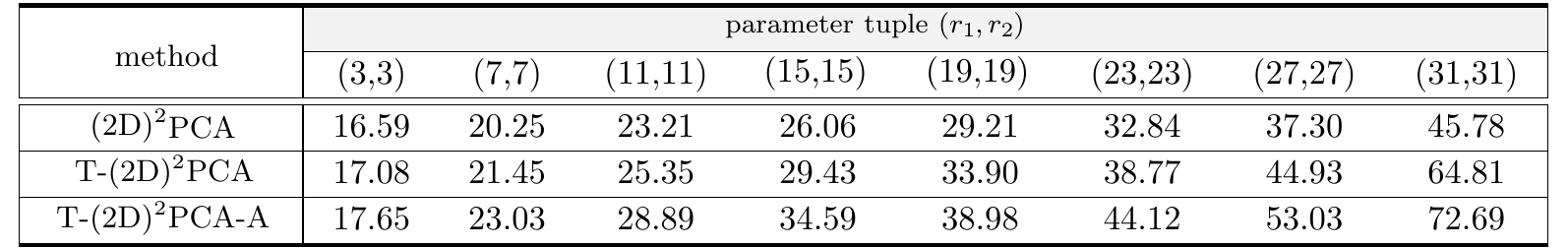} \\
	\end{tabular}
	\caption{A comparison of the PSNRs by (2D)\textsuperscript{2}PCA and 
		T-(2D)\textsuperscript{2}PCA on $1000$ CIFA images where  T-(2D)\textsuperscript{2}PCA is with order-two t-scalars, and
		T-(2D)\textsuperscript{2}PCA-A is withh order-four t-scalars}
	\label{figure:2D2PCA-and-beyond}
\end{figure*}

Figure \ref{figure:2D2PCA-and-beyond} shows the PSNRs by 
(2D)\textsuperscript{2}PCA and 
T-(2D)\textsuperscript{2}PCA using either order-two t-scalars or order-four t-scalars.  
The observation from this figure is consistent with those found in other experiments --- the generalized algorithm outperforms its canonical counterpart, and an algorithm using ``deeper-order'' t-scalars outperforms its counterpart using ``shallower-order'' t-scalars.

One can further optimize the results of (2D)\textsuperscript{2}PCA and T-(2D)\textsuperscript{2}PCA with the alternating least-squares algorithms (HOOI and THOOI). 

Table \ref{tab:with-local-optimization}
lists the quantitative PSNRs of related algorithms, using or not using alternating optimization with some representative parameters. It shows that alternating optimization can increase performance.   
Furthermore, the tabulated results are consistent with the conclusion found in previous experiments --- an algorithm using ``deeper-order'' t-scalars outperforms its counterpart using ``shallower-order'' t-scalars.

\begin{table*}[htbp]
	\centering
	\caption{A quantitative comparison of the PSNRs with some representative parameters using or not using alternating optimization}
	\begin{tabular}{|c|cccccccc|c|}
		\hline 
		
		\hline 
		
		\hline 
		\multirow{2}[0]{*}{method} & \multicolumn{8}{c|}{parameter tuple $(r_1, r_2)$}  & \multirow{2}[0]{*}{optimized?} \\
		\cline{2-9}
		& \small$(3,3)$ & \small$(7,7)$ & \small$(11,11)$ & \small$(15,15)$ & \small$(19,19)$ & \small$(23,23)$ & \small$(27,27)$ & \small$(31,31)$ &  \\
		\hline
		\hline
		(2D)\textsuperscript{2}PCA     & $16.59$ & $20.25$ & $23.21$ & $26.06$ & $29.21$ & $32.84$ & $37.30$  & $45.78$ & no \\
		\hline
		2D\textsuperscript{2}PCA-OP     & 16.61 & 20.26 & 23.21 & 26.06 & 29.22 & 32.84 & 37.31 & 45.78 & yes \\
		\hline
		\hline
		T-(2D)\textsuperscript{2}PCA     & 17.08 & 21.45 & 25.35 & 29.43 & 33.9  & 38.77 & 44.93 & 64.81 & no \\
		\hline
		T-(2D)\textsuperscript{2}PCA-OP     & 19.73 & 24.47 & 28.54 & 32.63 & 37.04 & 41.78 & 48.03 & 67.91 & yes \\
		\hline\hline
		T-(2D)\textsuperscript{2}PCA-A     & 17.65 & 23.03 & 28.89 & 34.59 & 38.98 & 44.12 & 53.03 & 72.69 & no \\
		\hline
		T-(2D)\textsuperscript{2}PCA-A-OP     & 20.42 & 26.19 & 32.18 & 37.78 & 42.08 & 47.27 & 56.25 & 77.12 & yes \\
		\hline 
		
		\hline 
		
		\hline 
	\end{tabular}%
	\label{tab:with-local-optimization}
\end{table*}

\subsection{``Horizontal'' comparison}
The comparison of related algorithms on the same images is conducted in a ``horizontal'' experiment.  The SVHN (Street View House Number) image set is chosen for the experiments. The SVHN image set contains over sixty hundred thousand
$32\times 32\times 3$ RGB digit images \footnote{\url{http://ufldl.stanford.edu/housenumbers/}}. 
The first $100$ images from its training set are chosen for the experiments.

To unify the PCA-based approximations, one needs to organize the mean-subtracted images into a $32\times 32\times 3\times 100$ array of real numbers. This underlying order-four array can be considered a canonical tensor of four modes (i.e., rows, columns, color channels, and samples) or a g-tensor of three modes (i.e., rows, columns, and samples) with the mode of color channels  chosen to characterize t-scalars.
Also worthy of notice is that the little-endian protocal mentioned in Section \ref{section:little-endian} requires permuting the grayscale indices of the underlying arrays, transforming their sizes from $32\times 32\times 3$ to $3\times 32\times 32$.

\begin{figure*}[htbp]
\centering
	\small
	\begin{tabular}{c}
		\hspace{6em}
		Some SVHN image samples \\
		\hspace{4.6em}
		\includegraphics[width=0.8\textwidth]{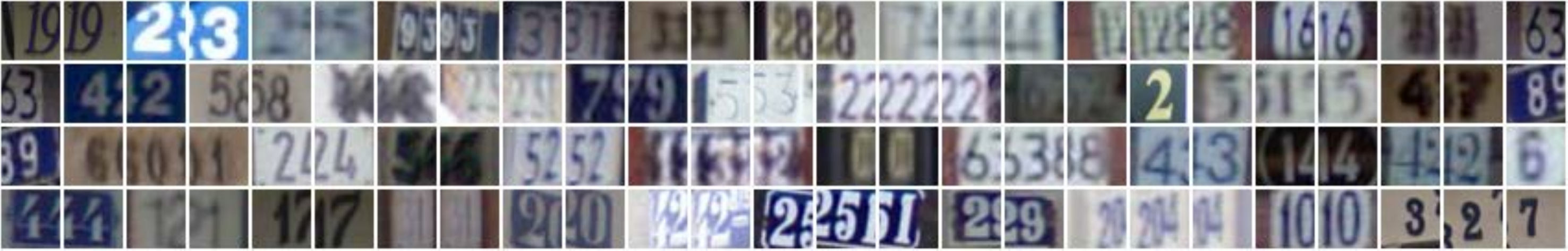} 
		\vspace{1em}\\
	\end{tabular}
	
	\begin{tabular}{cc}
		\hspace{5em}
		\liaoliangtab{~~~~~~
			PSNRs by (2D)\textsuperscript{2}PCA
		}{and PSNR gain by T-(2D)\textsuperscript{2}PCA}
		& 
		\hspace{3em}
		\liaoliangtab{~~~~~~
			PSNRs by (2D)\textsuperscript{2}PCA-OP
		}{and PSNR gain by T-(2D)\textsuperscript{2}PCA-OP}
		\\
		\hspace{3em}\includegraphics[height=0.25\textwidth]{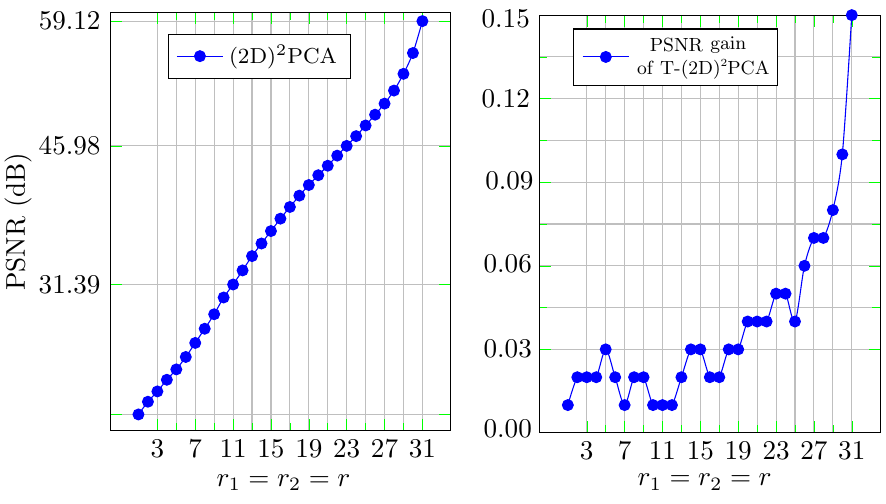} 
		& 
		\includegraphics[height=0.25\textwidth]{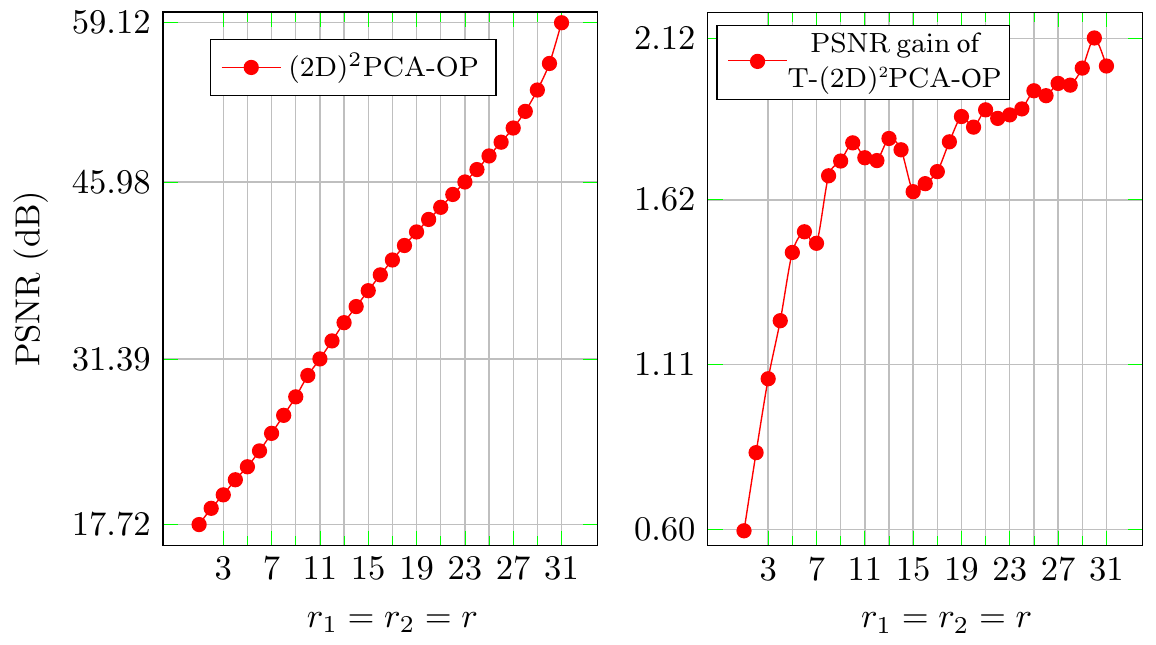}  \\
		\multicolumn{2}{c}{
			\liaoliangtab{\hspace{19em}A quantitative PSNR comparison of}{
				(2D)\textsuperscript{2}PCA and 
				T-(2D)\textsuperscript{2}PCA with some representative parameter tuples on the identical SVHN image set}
		} \\
		\multicolumn{2}{c}{
			\hspace{3em}\includegraphics[width=0.86\textwidth]{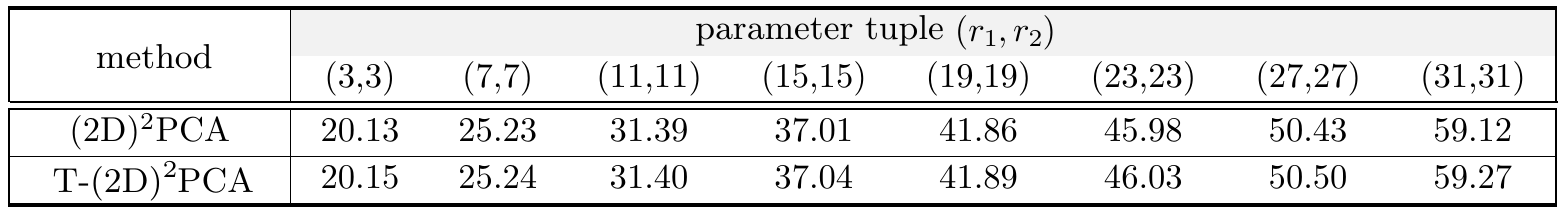}
		}  \\
		\multicolumn{2}{c}{\vspace{-0.7em}} \\
		\multicolumn{2}{c}{
			\liaoliangtab{\hspace{19em}A quantitative  PSNR comparison of}{
				optimzied algorithms with some representative parameter tuples on the identical SVHN image set}
		} \\
		\multicolumn{2}{c}{
			\hspace{3em}\includegraphics[width=0.86\textwidth]{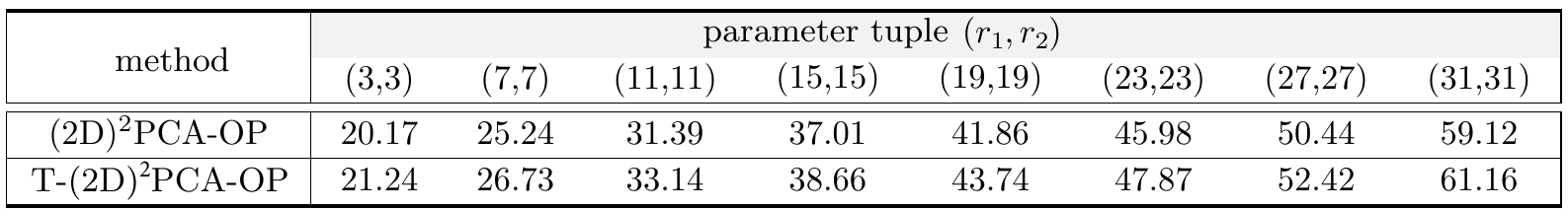}
		}
	\end{tabular}%
	\caption{A ``horizontal'' comparision of the results 
		on the same SVHN images 
		by (2D)\textsuperscript{2}PCA and 
		T-(2D)\textsuperscript{2}PCA with or without alternating optimization}
	\label{figure:SVHN-2D2PCA-reconstruction}
\end{figure*}

The top row of Figure \ref{figure:SVHN-2D2PCA-reconstruction} 
shows the chosen SVHN images. 
The first two subfigures of the second row show the PSNRs by (2D)\textsuperscript{2}PCA and
the PSNR gain of T-(2D)\textsuperscript{2}PCA on the same SVHN images. 
The last two subfigures of the second row show the PSNRs
on the same images by the optimized algorithms (2D)\textsuperscript{2}PCA-OP and 
T-(2D)\textsuperscript{2}PCA-OP.
Some quantitative PSNRs with representative parameters $r_1$, $r_2$ are also given in Figure \ref{figure:SVHN-2D2PCA-reconstruction}.

The ``horizontal'' comparison shows that the generalized algorithms 
T-(2D)\textsuperscript{2}PCA and  
T-(2D)\textsuperscript{2}PCA 
outperform their canonical counterparts 
(2D)\textsuperscript{2}PCA and 
(2D)\textsuperscript{2}PCA
on the same SVHN images.

Another ``horizontal'' experiment compares the performances of PCA and TPCA. TPCA is a ``deeper-order'' generation of PCA \cite{2017Ren,liao2020generalized}. 
Therefore, one can employ HOSVD and THOSVD to realize PCA and TPCA, respectively.

Given a $32\times 32\times 3\times 100$ array formed by the $100$ mean-subtracted SVHN image samples, PCA on these samples is equivalent to HOSVD conducted merely on the sample mode of the underlying array.
The $32\times 32\times 3 \times 100$ array can also be interpreted as a g-tensor in $C^{\,32\times 32\times 100}$ (i.e., $32$ rows, $32$ columns, $100$ samples, and each t-scalar entry containing three numbers).
Analogous to PCA, TPCA is equivalent to THOSVD performed merely with the ``sample'' mode of the g-tensor.

It is noted that to implement 
specific g-tensors using the little-endian protocol discussed in Section  \ref{section:little-endian}, one must permute the underlying $32\times 32\times 3\times 100$ array to an underlying $3\times 32\times 32\times 100$ array.

\begin{figure*}[tbh]
	\centering
	\small
	\begin{tabular}{c}
		\hspace{3em}PSNR by PCA and PSNR gain by TPCA with different $r$ on the same SVHN image set \\
		\includegraphics[width=0.7\textwidth]{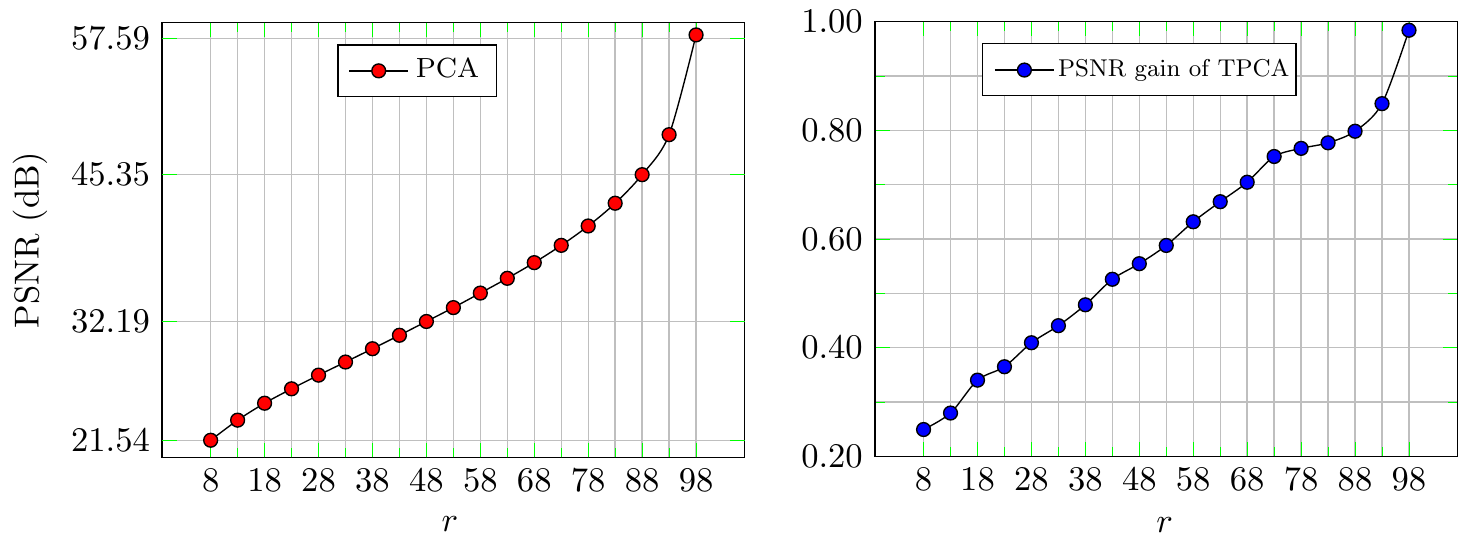} \hspace{1em} \\
		PSNRs by PCA and TPCA 
		with some representative parameter $r$
		on the identical SVHN image set \\
		\includegraphics[width=0.8\textwidth]{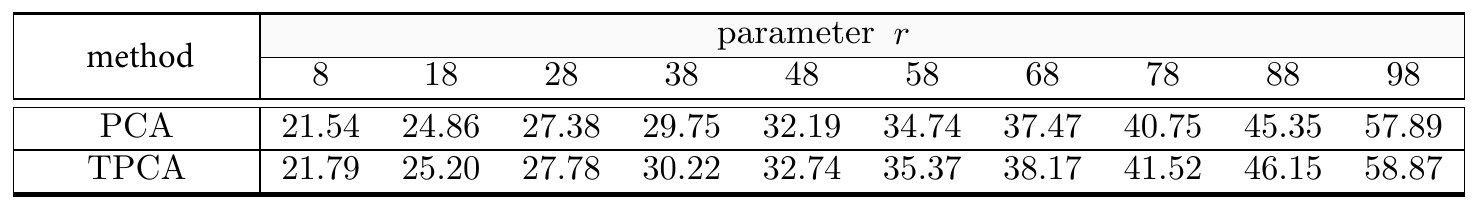} \\
	\end{tabular}
	\caption{A horizontal comparison of PSNRs by PCA and TPCA on the same SVHN image samples}
	\label{figure:PCA-TPCA}	
\end{figure*}

The PSNR curves yielded by PCA and TPCA on the same image SVHN samples are shown in Figure \ref{figure:PCA-TPCA}. 
A quantitative comparison of PCA and TPCA with some representative parameter $r \in [99]$ is also tabulated in Figure \ref{figure:PCA-TPCA}. It shows that TPCA consistently outperforms PCA with different $r$ on the same SVNH images. 
It reconfirms the conclusion that an algorithm established over t-scalars outperforms its counterpart over canonical scalars (i.e., complex numbers). 

It is noted that the alternating optimization algorithm (HOOI or THOOI) does not increase the performances of PCA and TPCA.

\section{Conclusions}
\label{section:Conclusions}

This paper discusses a generalization of the higher-order singular value decomposition (HOSVD) over a finite-dimensional commutative algebra $C$ called t-algebra. The elements of the algebra, called t-scalars, are fixed-sized multiway arrays of complex numbers 
The arrays generalize the complex numbers. Vectors, matrices, and even tensors can be built over the t-algebra and behave analogously to the canonical counterparts over complex numbers. 
The generalized vectors, matrices, and tensors over t-scalars are called t-vectors, t-matrices, and g-tensors. The modules formed by t-vectors, t-matrices, or g-tensors are modules over both complex numbers and t-scalars.

This module duality requires t-scalar-valued notions to describe its $C$-linear aspect and scalar-valued notions to describe its linear aspect.  
To describe the linear aspect of t-algebras and t-matrices over the algebra, we introduce the standard matrix representation of t-scalars and t-matrices. With the matrix representation, a t-scalar is represented by a diagonal matrix whose diagonal entries are the eigenvalues of the t-scalar as a linear operator. As a result, a t-matrix is representable by a (canonical) block-diagonal matrix. The proposed matrix representation links many t-scalar-valued notions  
to their canonical versions in matrix theory.

With the introduced matrix representation, this paper shows how to generalize HOSVD (Higher Order Singular Value Decomposition) to THOSVD (t-algebra based HOSVD). THOSVD generalizes SVD and HOSVD \cite{2000A} and unifies a wide range of PCA-based algorithms, such as PCA, 2DPCA \cite{yang2004two}, MPCA \cite{lu2008mpca}, and their generalized versions over t-scalars, i.e., TPCA \cite{2017Ren, liao2020generalized}, T-2DPCA \cite{liao2020generalized}, and TMPCA.

The approximation by THOSVD is usually not optimal. Instead, this paper proposes a generalization of the alternating optimization algorithm HOOI (Higher Order Orthogonal Iteration) over t-scalars. The experiments on public data show that the generalized alternating optimization algorithm THOOI (t-algebra based HOOI) improves the results of THOSVD.

This paper also introduces a pixel neighborhood strategy to extend a ``shallower-order'' image to its ``deeper-order'' version. If the pixel neighborhood strategy is nested, one can further increase the image's order. Experiments show that the generalized algorithms using deeper-order t-scalars compare favorably with their counterparts using shallower-order t-scalars. The canonical algorithms are the special cases with the ``shallowest order'' t-scalars (the order-zero t-scalars, i.e., complex numbers).

The theory of t-vectors, t-matrices, and g-tensors provides a consistent, straightforward framework for generalizing many (canonical) matrix and tensor algorithms. 




\bibliographystyle{IEEEtran}
\bibliography{liaoliang}

\begin{thebibliography}{10}
\providecommand{\url}[1]{#1}
\csname url@samestyle\endcsname
\providecommand{\newblock}{\relax}
\providecommand{\bibinfo}[2]{#2}
\providecommand{\BIBentrySTDinterwordspacing}{\spaceskip=0pt\relax}
\providecommand{\BIBentryALTinterwordstretchfactor}{4}
\providecommand{\BIBentryALTinterwordspacing}{\spaceskip=\fontdimen2\font plus
\BIBentryALTinterwordstretchfactor\fontdimen3\font minus
  \fontdimen4\font\relax}
\providecommand{\BIBforeignlanguage}[2]{{%
\expandafter\ifx\csname l@#1\endcsname\relax
\typeout{** WARNING: IEEEtran.bst: No hyphenation pattern has been}%
\typeout{** loaded for the language `#1'. Using the pattern for}%
\typeout{** the default language instead.}%
\else
\language=\csname l@#1\endcsname
\fi
#2}}
\providecommand{\BIBdecl}{\relax}
\BIBdecl

\bibitem{1991Algebraic}
Z.~Q. Hong, ``Algebraic feature extraction of image for recognition,''
  \emph{Pattern Recognition}, vol.~24, no.~3, pp. 211--219, 1991.

\bibitem{2005Mei}
M.~Tian, S.-W. Luo, and L.-Z. Liao, ``An investigation into using singular
  value decomposition as a method of image compression,'' in \emph{2005
  International Conference on Machine Learning and Cybernetics}, 8, Ed., 2005,
  pp. 5200--5204.

\bibitem{2019Dabass}
M.~Dabass, S.~Vashisth, and R.~Vig, ``Lossy color image compression technique
  using reduced bit plane-quaternion svd,'' in \emph{2019 9th International
  Conference on Cloud Computing, Data Science Engineering (Confluence)}, 2019,
  pp. 504--509.

\bibitem{2020RST}
R.~K. Senapati, S.~Srivastava, and P.~Mankar, ``Rst invariant blind image
  watermarking schemes based on discrete tchebichef transform and singular
  value decomposition,'' \emph{Optik - International Journal for Light and
  Electron Optics}, vol.~45, no.~4, pp. 3331--3353, 2020.

\bibitem{2021Hybrid}
Z.~Zainol, J.~S. Teh, M.~Alawida, A.~Alabdulatif \emph{et~al.}, ``Hybrid
  svd-based image watermarking schemes: A review,'' \emph{IEEE Access}, vol.~9,
  pp. 32\,931--32\,968, 2021.

\bibitem{2017Computing}
K.~Batselier, W.~Yu, L.~Daniel, and N.~Wong, ``Computing low-rank
  approximations of large-scale matrices with the tensor network randomized
  svd,'' \emph{SIAM Journal on Matrix Analysis and Applications}, 2017.

\bibitem{2015Li}
M.~Li, W.~Bi, J.~T. Kwok, and B.-L. Lu, ``Large-scale nystr{\"o}m kernel matrix
  approximation using randomized svd,'' \emph{IEEE transactions on neural
  networks and learning systems}, vol.~26, no.~1, pp. 152--164, 2014.

\bibitem{1928Multiple}
F.~L. Hitchcock, ``Multiple invariants and generalized rank of a p-way matrix
  or tensor,'' \emph{Journal of Mathematics and Physics}, vol.~7, no. 1-4, pp.
  39--79, 1928.

\bibitem{Tucker1963}
L.~R. Tucker, ``Implications of factor analysis of three-way matrices for
  measurement of change,'' \emph{In Problems in Measuring Change(C W. Harris
  ed.),University of Wisconsin Press}, pp. 122--137, 1963.

\bibitem{Tucker1964}
------, ``The extension of factor analysis to three-dimensional matrices,''
  \emph{In Contributions to Mathematical Psychology(H. Gulliksen and N.
  Frederiksen, eds.)}, pp. 109--127, 1964.

\bibitem{Tucker1966}
------, ``Some mathematical notes on three-mode factor analysis,''
  \emph{Psychometrika}, vol.~31, pp. 279--311, 1966.

\bibitem{2017Tensor}
N.~D. Sidiropoulos, L.~D. Lathauwer, X.~Fu, K.~Huang, E.~E. Papalexakis, and
  C.~Faloutsos, ``Tensor decomposition for signal processing and machine
  learning,'' \emph{IEEE Transactions on Signal Processing}, vol.~65, no.~13,
  pp. 3551--3582, 2017.

\bibitem{2021Panagakis}
Y.~Panagakis, J.~Kossaifi, G.~G. Chrysos, J.~Oldfield, M.~A. Nicolaou,
  A.~Anandkumar, and S.~Zafeiriou, ``Tensor methods in computer vision and deep
  learning,'' \emph{Proceedings of the IEEE}, vol. 109, no.~5, pp. 863--890,
  2021.

\bibitem{2018Pre}
R.~K. Renu, V.~Sowmya, and K.~P. Soman, ``Pre-processed hyperspectral image
  analysis using tensor decomposition techniques,'' \emph{Advances in Signal
  Processing and Intelligent Recognition Systems}, pp. 205--216, 2019.

\bibitem{2017Ren}
Y.~Ren, L.~Liao, S.~J. Maybank, Y.~Zhang, and X.~Liu, ``Hyperspectral image
  spectral-spatial feature extraction via tensor principal component
  analysis,'' \emph{IEEE Geoscience and Remote Sensing Letters}, vol.~14,
  no.~9, pp. 1431--1435, 2017.

\bibitem{1994Blind}
L.~De~Lathauwer, B.~De~Moor, and J.~Vandewalle, ``Blind source separation by
  higher-order singular value decomposition,'' in \emph{Proc. EUSIPCO}, vol.~1,
  1994, pp. 175--178.

\bibitem{2000A}
L.~D. Lathauwer, ``A multilinear singular value decomposition,'' \emph{{SIAM}
  Journal on Matrix Analysis and Applications}, vol.~21, no.~4, pp. 1253--1278,
  2000.

\bibitem{2011Higher}
T.~J. Gregor, ``Higher order singular value decomposition of tensors for fusion
  of registered images,'' \emph{Journal of Electronic Imaging}, vol.~20, no.~1,
  pp. 9--15, 2011.

\bibitem{2012Image}
J.~Liang, H.~Yang, L.~Ding, and X.~Zeng, ``Image fusion using higher order
  singular value decomposition,'' \emph{IEEE Trans Image Process}, vol.~21,
  no.~5, pp. 2898--2909, 2012.

\bibitem{2015Denoising}
X.~Zhang, Z.~Xu, N.~Jia, W.~Yang, Q.~Feng, W.~Chen, and Y.~Feng, ``Denoising of
  3d magnetic resonance images by using higher-order singular value
  decomposition,'' \emph{Medical Image Analysis}, vol.~19, no.~1, pp. 75--86,
  2015.

\bibitem{2013A}
X.~Geng, L.~Ji, Y.~Zhao, and F.~Wang, ``A small target detection method for the
  hyperspectral image based on higher order singular value decomposition
  (hosvd),'' \emph{IEEE Geoscience and Remote Sensing Letters}, vol.~10, no.~6,
  pp. 1305--1308, 2013.

\bibitem{shi2019quaternion}
J.~Shi, X.~Zheng, J.~Wu, B.~Gong, Q.~Zhang, and S.~Ying, ``Quaternion grassmann
  average network for learning representation of histopathological image,''
  \emph{Pattern Recognition}, vol.~89, pp. 67--76, 2019.

\bibitem{miao2020low}
J.~Miao, K.~I. Kou, and W.~Liu, ``Low-rank quaternion tensor completion for
  recovering color videos and images,'' \emph{Pattern Recognition}, vol. 107,
  2020.

\bibitem{miao2021color}
J.~Miao and K.~I. Kou, ``Color image recovery using low-rank quaternion matrix
  completion algorithm,'' \emph{IEEE Transactions on Image Processing},
  vol.~31, pp. 190--201, 2021.

\bibitem{hosny2019new}
K.~M. Hosny and M.~M. Darwish, ``New set of multi-channel orthogonal moments
  for color image representation and recognition,'' \emph{Pattern Recognition},
  vol.~88, pp. 153--173, 2019.

\bibitem{chen2019low}
Y.~Chen, X.~Xiao, and Y.~Zhou, ``Low-rank quaternion approximation for color
  image processing,'' \emph{IEEE Transactions on Image Processing}, vol.~29,
  pp. 1426--1439, 2019.

\bibitem{yu2019quaternion}
Y.~Yu, Y.~Zhang, and S.~Yuan, ``Quaternion-based weighted nuclear norm
  minimization for color image denoising,'' \emph{Neurocomputing}, vol. 332,
  pp. 283--297, 2019.

\bibitem{zou2016quaternion}
C.~Zou, K.~I. Kou, and Y.~Wang, ``Quaternion collaborative and sparse
  representation with application to color face recognition,'' \emph{IEEE
  Transactions on image processing}, vol.~25, no.~7, pp. 3287--3302, 2016.

\bibitem{zou2019grayscale}
C.~Zou, K.~I. Kou, L.~Dong, X.~Zheng, and Y.~Y. Tang, ``From grayscale to
  color: Quaternion linear regression for color face recognition,'' \emph{IEEE
  Access}, vol.~7, pp. 154\,131--154\,140, 2019.

\bibitem{miao2020quaternion}
J.~Miao and K.~I. Kou, ``Quaternion-based bilinear factor matrix norm
  minimization for color image inpainting,'' \emph{IEEE Transactions on Signal
  Processing}, vol.~68, pp. 5617--5631, 2020.

\bibitem{jia2019robust}
Z.~Jia, M.~K. Ng, and G.-J. Song, ``Robust quaternion matrix completion with
  applications to image inpainting,'' \emph{Numerical Linear Algebra with
  Applications}, vol.~26, no.~4, p. e2245, 2019.

\bibitem{kilmer2011factorization}
M.~E. Kilmer and C.~D. Martin, ``Factorization strategies for third-order
  tensors,'' \emph{Linear Algebra and its Applications}, vol. 435, no.~3, pp.
  641--658, 2011.

\bibitem{kilmer2013third}
M.~E. Kilmer, K.~Braman, N.~Hao, and R.~C. Hoover, ``Third-order tensors as
  operators on matrices: A theoretical and computational framework with
  applications in imaging,'' \emph{{SIAM} Journal on Matrix Analysis and
  Applications}, vol.~34, no.~1, pp. 148--172, 2013.

\bibitem{liao2020generalized}
L.~Liao and S.~J. Maybank, ``Generalized visual information analysis via
  tensorial algebra,'' \emph{Journal of Mathematical Imaging and Vision},
  vol.~62, pp. 560--584, 2020.

\bibitem{liao2020general}
------, ``General data analytics with applications to visual information
  analysis: A provable backward-compatible semisimple paradigm over
  t-algebra,'' \emph{arXiv preprint arXiv:2011.00307}, pp. 1--53, 2020.

\bibitem{osgood2019lectures}
B.~G. Osgood, \emph{Lectures on the Fourier Transform and its
  Applications}.\hskip 1em plus 0.5em minus 0.4em\relax American Mathematical
  Soc., 2019, vol.~33.

\bibitem{krizhevsky2012imagenet}
A.~Krizhevsky, I.~Sutskever, and G.~E. Hinton, ``Imagenet classification with
  deep convolutional neural networks,'' \emph{Advances in neural information
  processing systems}, vol.~25, pp. 1097--1105, 2012.

\bibitem{zeiler2014visualizing}
M.~D. Zeiler and R.~Fergus, ``Visualizing and understanding convolutional
  networks,'' in \emph{European conference on computer vision}.\hskip 1em plus
  0.5em minus 0.4em\relax Springer, 2014, pp. 818--833.

\bibitem{Simonyan15}
K.~Simonyan and A.~Zisserman, ``Very deep convolutional networks for
  large-scale image recognition,'' in \emph{International Conference on
  Learning Representations}, 2015.

\bibitem{szegedy2015going}
C.~Szegedy, W.~Liu, Y.~Jia, P.~Sermanet, S.~Reed, D.~Anguelov, D.~Erhan,
  V.~Vanhoucke, and A.~Rabinovich, ``Going deeper with convolutions,'' in
  \emph{Proceedings of the IEEE conference on computer vision and pattern
  recognition}, 2015, pp. 1--9.

\bibitem{he2016deep}
K.~He, X.~Zhang, S.~Ren, and J.~Sun, ``Deep residual learning for image
  recognition,'' in \emph{Proceedings of the IEEE conference on computer vision
  and pattern recognition}, 2016, pp. 770--778.

\bibitem{goodfellow2016deep}
I.~Goodfellow, Y.~Bengio, and A.~Courville, \emph{Deep learning}.\hskip 1em
  plus 0.5em minus 0.4em\relax MIT press, 2016, ch. 9.1 The convolution
  operation.

\bibitem{zhang2017exact}
Z.~Zhang and S.~Aeron, ``Exact tensor completion using t-{SVD},'' \emph{IEEE
  Transactions on Signal Processing}, vol.~65, no.~6, pp. 1511--1526, 2017.

\bibitem{hou2021robust}
J.~Hou, F.~Zhang, H.~Qiu, J.~Wang, Y.~Wang, and D.~Meng, ``Robust
  low-tubal-rank tensor recovery from binary measurements,'' \emph{IEEE
  Transactions on Pattern Analysis and Machine Intelligence}, 2021.

\bibitem{jiang2019robust}
Q.~Jiang and M.~Ng, ``Robust low-tubal-rank tensor completion via convex
  optimization,'' in \emph{Proc. IJCAI}, 2019, pp. 2649--2655.

\bibitem{dian2019hyperspectral}
R.~Dian and S.~Li, ``Hyperspectral image super-resolution via subspace-based
  low tensor multi-rank regularization,'' \emph{IEEE Transactions on Image
  Processing}, vol.~28, no.~10, pp. 5135--5146, 2019.

\bibitem{cheng2018tensor}
M.~Cheng, L.~Jing, and M.~K. Ng, ``Tensor-based low-dimensional representation
  learning for multi-view clustering,'' \emph{IEEE Transactions on Image
  Processing}, vol.~28, no.~5, pp. 2399--2414, 2018.

\bibitem{yin2018multiview}
M.~Yin, J.~Gao, S.~Xie, and Y.~Guo, ``Multiview subspace clustering via
  tensorial t-product representation,'' \emph{IEEE transactions on neural
  networks and learning systems}, vol.~30, no.~3, pp. 851--864, 2018.

\bibitem{dolgov2014alternating}
S.~V. Dolgov and D.~V. Savostyanov, ``Alternating minimal energy methods for
  linear systems in higher dimensions,'' \emph{{SIAM} Journal on Scientific
  Computing}, vol.~36, no.~5, pp. A2248--A2271, 2014.

\bibitem{liu2021tensors}
Y.~Liu, \emph{Tensors for Data Processing: Theory, Methods, and
  Applications}.\hskip 1em plus 0.5em minus 0.4em\relax Elsevier, 2021, ch.
  {1.2.3.}

\bibitem{cohen1981holy}
D.~Cohen, ``On holy wars and a plea for peace,'' \emph{Computer}, vol.~14,
  no.~10, pp. 48--54, 1981.

\bibitem{de2000best}
L.~De~Lathauwer, B.~De~Moor, and J.~Vandewalle, ``On the best rank-1 and
  rank-($r_1,r_2,\cdots ,r_n$) approximation of higher-order tensors,''
  \emph{{SIAM} journal on Matrix Analysis and Applications}, vol.~21, no.~4,
  pp. 1324--1342, 2000.

\bibitem{1980Principal}
P.~M. Kroonenberg and J.~D. Leeuw, ``Principal component analysis of three-mode
  data by means of alternating least squares algorithms,''
  \emph{Psychometrika}, vol.~45, no.~1, pp. 69--97, 1980.

\bibitem{sheehan2007higher}
B.~N. Sheehan and Y.~Saad, ``Higher order orthogonal iteration of tensors
  ({HOOI}) and its relation to {PCA} and {GLRAM},'' in \emph{Proceedings of the
  2007 SIAM International Conference on Data Mining}.\hskip 1em plus 0.5em
  minus 0.4em\relax SIAM, 2007, pp. 355--365.

\bibitem{zhang20052d}
D.~Zhang and Z.-H. Zhou, ``{(2D)}\textsuperscript{2}{PCA}: {T}wo-directional
  two-dimensional {PCA} for efficient face representation and recognition,''
  \emph{Neurocomputing}, vol.~69, no. 1-3, pp. 224--231, 2005.

\bibitem{lu2008mpca}
H.~Lu, K.~N. Plataniotis, and A.~N. Venetsanopoulos, ``{MPCA}: {M}ultilinear
  principal component analysis of tensor objects,'' \emph{IEEE transactions on
  Neural Networks}, vol.~19, no.~1, pp. 18--39, 2008.

\bibitem{liao2021TPCA}
L.~Liao, X.~Zhang, X.~Wang, S.~Lin, and X.~Liu, ``Generalized image
  reconstruction over t-algebra,'' in \emph{Proceedings of the 2021 3rd
  International Conference on Advances in Computer Technology}.\hskip 1em plus
  0.5em minus 0.4em\relax IEEE, 2021, pp. 387--392.

\bibitem{yang2004two}
J.~Yang, D.~Zhang, A.~F. Frangi, and J.-y. Yang, ``Two-dimensional {PCA}: a new
  approach to appearance-based face representation and recognition,''
  \emph{IEEE transactions on pattern analysis and machine intelligence},
  vol.~26, no.~1, pp. 131--137, 2004.

\end{thebibliography}

\section*{Appendix: Eigen solutions, Idempotency, and decomposition}

\begin{center}
Liang Liao \\
liaoliang@ieee.org 	
\end{center}

\subsection{Eigen solutions}
As mentioned above, each t-scalar is representable by a diagonal matrix whose diagonal entries are the eigenvalues of the t-scalar. Specifically, let $X_{T}$ be a t-scalar in $C$. An eigenvalue $\lambda$ is a complex number such that the t-scalar 
$ (X_{T} - \lambda \cdot E_{T})
$ is multiplicatively non-invertible
or equivalently, the matrix  
$M(X_{T}) - \lambda \cdot I_{K}  
$ is rank deficient, where $I_{K}$ denotes the $K\times K$ identity matrix and $K$ the dimension of $C$. 

The Eigen solution of a t-scalar $X_{T}\in C$, 
corresponding to an 
eigenvalue $\lambda \in \mathbbm{C}$, is a rank-one and norm-one t-scalar $Y_{T}$ such that the following conditions hold
\begin{equation} 
X_{T} \circ Y_{T} = \lambda \cdot Y_{T} \;, 	
\label{equation:eigen}
\end{equation}
where $\operatorname{rank} Y_{T} = 1$ and $\|Y_{T}\|_{F} = 1$.

Using the matrix representations to equations (\ref{equation:eigen}), we have the following equation
\begin{equation}
M(X_{T}) \cdot M(Y_{T}) = \lambda \cdot M(Y_{T}) 
\end{equation}
where the matrix $M(Y_{T})$ is rank-one and norm-one, i.e., 
$\operatorname{rank} M(Y_{T}) = \|M(Y_{T})\|_{F} = 1$.

Because the eigenvalue $\lambda$ is a diagonal entry of 
$M(X_{T}) \doteq 
\operatorname{diag} [F_1(X_T),\cdots, F_K(X_T)] $, without losing generality, let $\lambda \doteq F_{k}(X_{T}) \in \mathbbm{C}$ and the Eigen t-scalar of equation (\ref{equation:eigen}) be $D_{{T}, \,k}$. Then, 
its matrix representation $M(D_{{T}, \,k})$ is uniquely given by 
\begin{equation}
(M(D_{T,\,k}) \,)_{i, j} = \delta_{i, k} \cdot \delta_{j, k} \in \{0, 1\}
\end{equation}
where $(M(D_{T,\,k}) \,)_{i, j}$ denotes the $(i,j)$-th entry of the matrix $M(D_{T,\,k}) $ and 
$\delta: (i, j) \mapsto \delta_{i, j} \in \{0, 1\}$ denotes the Kronecker delta function.

Namely, the following equation holds for all $k \in [K]$, 
\begin{equation}
\begin{aligned}	
 X_{T} \circ D_{T, \,k} = F_k(X_{T}) \cdot D_{T, \,k}  \;,\vspace{0.5em}\\	
 \operatorname{rank} D_{T, \,k} = 1,\;\; \|D_{T, \,k}\|_F = 1 \;. 
\end{aligned}
\label{equation:eigen-equation}
\end{equation}

\subsection{Idempotency}

The rank-one and norm-one t-scalars $D_{T,k} \in C, \forall k \in [K]$ are idempotent such that the following condition holds, 
\begin{equation}
D_{T,k} \circ D_{T,k} = D_{T,k}\;, \forall k \in [K] \,.
\end{equation}

There are only $K$ normalized rank-one idempotent t-scalars. 
These rank-one idempotent t-scalars are critical to decomposing the t-algebra $C$ and modules over $C$ into a finite number of simpler factors.  

It shows that the Eigen solutions of equation (\ref{equation:eigen}) are always the rank-one idempotent t-scalars for any t-scalar $X_{T} \in C$.

These rank-one idempotent t-scalar $D_{T,\,1},\cdots,D_{T,\,K}$ are ``eigenvectors'' because they are all elements of the underlying vector space of $C$.  Furthermore, the following condition holds
\begin{equation}
	D_{T,\,1} + \cdots + D_{T,\,K} = E_T \;.
\label{equation:SVD-of-E}
\end{equation}

It shows that one can decompose the t-scalar $E_{T}$ into a finite number of rank-one and normalized components. By definition, one can call equation (\ref{equation:SVD-of-E}) the (canonical) singular value decomposition of $E_{T}$, all singular values equal to $1$, 
i.e., the coefficients of the above linear combination. 

By equation (\ref{equation:SVD-of-E}), the following condition holds for all t-scalars 
$X_T \in C$, such that 
\begin{equation}
\begin{aligned}	
X_{T} &\equiv X_T \circ (D_{T,\,1} + \cdots + D_{T,\,K}) \\
      &= F_1(X_T) \cdot D_{T,\,1} +\cdots + F_K(X_T) \cdot  D_{T,\,K} \\
      &= |F_1(X_T)| \cdot \hat{D}_{T,\,1}  +\cdots + 
         |F_K(X_T)| \cdot \hat{D}_{T,\,K} 
\end{aligned}
\label{equation:SVD-of-a-tscalar}
\end{equation}
where $\hat{D}_{T,\,k} \doteq F_k(X_T) \cdot |F_k(X_T)|^{-1} \cdot D_{T,\,k}$ if  
$F_k(X_T) \neq 0$, otherwise, $\hat{D}_{T,\,k} \doteq {D}_{T,\,k}$, and  
the condition $\operatorname{rank} \hat{D}_k = \|\hat{D}_{T,\,k}\|_F = 1$ holds 
for all $k \in [K]$. 

Equation (\ref{equation:SVD-of-a-tscalar}) shows that any t-scalar can be decomposed to a finite number of rank-one normalized components. By definition, equation (\ref{equation:SVD-of-a-tscalar}) is called the (canonical) singular value decomposition of a t-scalar, the singular values being the moduli of the eigenvalues of the t-scalar.


One might still feel uncertain on the interpretation of equation (\ref{equation:eigen-equation}) since,  until now, one can only represent each t-scalar by a square matrix rather than a Euclidean vector.

\subsection{Vector representation}
To the above concern, besides the diagonal matrix representation, we introduce the  vector representation of a t-scalar in $C$. 
The vector representation of any t-scalar $X_{T}$ is a $K$-tuple $V(X_{T})$ given by
\begin{equation}
V(X_{T}) \doteq \big(
F_{1}(X_T),\cdots,F_{K}(X_T) 
\,\big)  \in \mathbbm{C}^{K} \;.
\end{equation}

One might like to organize the $K$-tuple as a column vector by customizing a 
suitable-sized left matrix multiplication to a column vector. 
Then, the following equivalences hold for all $X_{T}, Y_{T} \in C$ and $\alpha, \beta \in \mathbbm{C}$, 
\begin{equation}
\begin{matrix}
\|X_{T}\|_{F} \doteq \|M(X_{T}) \|_{F} \equiv 
 \| V(X_{T}) \|_F \,,\vspace{0.5em}\\
\alpha \cdot X_{T} + \beta \cdot  Y_{T} \,\overset{V}{\sim}\, 
\alpha \cdot  V(X_{T}) + \beta \cdot   V(Y_{T}) ,\vspace{0.5em}\\
X_{T} \circ Y_{T} \,\overset{V}{\sim}\, V(X_{T} \circ Y_{T}) \equiv  M(X_{T}) \cdot V(Y_{T})\,.
\end{matrix}
\end{equation}
where $\overset{V}{\sim}$ denotes the 
the equivalence relationship under the vector representation $V: X_{T} \mapsto V(X_{T}) \in \mathbbm{C}^{K}$. 

Then, applying the vector representation to equation (\ref{equation:eigen-equation}) leads to the following equivalent equation, 
consistent with the canonical eigenvalue-eigenvector formulation 
\begin{equation} 
M(X_{T}) \cdot V(D_{T,k}) = F_k(X_T) \cdot V(D_{T,k}) \in \mathbbm{C}^{K}  
\end{equation}
where $V(D_{T,k})$, an eigenvector of the matrix $M(X_{T})$, is norm-one and, apparently, rank-one since it is a non-zero column vector.

One can extend the vector representation of a t-scalar to a t-vector. 
Let $X_\mathit{TV}$ be a t-vector in $C^{N}$, whose $i$-th t-scalar entry is denoted by $X_{T, i} \doteq [X_{T}]_{i} \in C$ for all $i \in [N]$. Then, the vector representation of $X_\mathit{TV}$ is given by following the $(KN)$-tuple 
organized as a column of complex numbers
\begin{equation}
V(X_\mathit{TV}) \doteq \big(
V(X_{T, 1}),\cdots, 
V(X_{T, N})
\,\big) \in \mathbbm{C}^{KN} \;.
\end{equation} 

Then, the following conditions hold for all $\alpha, \beta \in \mathbbm{C}$ and 
$X_\mathit{TM}, X_\mathit{TV}, Y_\mathit{TV}$ of appropriate sizes,  
\begin{equation}
\begin{matrix}
\|X_\mathit{TV} \|_F \doteq \|M(X_\mathit{TV}) \|_F \equiv 
 \|V(X_\mathit{TV})\|_F  \;, \vspace{0.5em}\\
\alpha \cdot X_\mathit{TV} + 
\beta \cdot  Y_\mathit{TV} \,\overset{V}{\sim}\, 
\alpha \cdot V(X_\mathit{TV}) + 
\beta \cdot  V(Y_\mathit{TV}) \vspace{0.5em}\\ 
X_\mathit{TM} \circ Y_{TV} 
\,\overset{V}{\sim}\, V(X_\mathit{TM} \circ Y_{TV}) \equiv
M(X_\mathit{TM}) \cdot  V(Y_\mathit{TV}) \;.   
\end{matrix}
\end{equation}

Notice that, except t-scalars and t-vectors, one can never represent a nontrivial t-matrix as a canonical vector. 
Again notice that, given any t-scalar or t-vector, its (canonical) rank is defined by the rank of its matrix representation, which is usually not equal to the rank of its vector representation whose value is always $1$ or $0$.

\subsection{SVD versus TSVD}

The (canonical) singular value decomposition (SVD) of any
t-matrix $X_\mathit{TM} \in C^{D_1\times D_2}$ reformulates $X_\mathit{TM}$ as the linear combination of a finite number of rank-one normalized t-matrices.
Specifically, the SVD of $X_\mathit{TM}$ gives the following linear combination,  
\begin{equation}
X_\mathit{TM} =  
\lambda_1 \cdot \hat{X}_{\mathit{TM},\,1}  + \cdots + 
\lambda_\mathit{KD}  \cdot \hat{X}_{\mathit{TM},\,
	\mathit{KD}}
\label{equation:SVD-a-matrix001}
\end{equation}
where $D \doteq \min(D_1, D_2)$, 
$\hat{X}_{\mathit{TM},\,i} \in C^{D_1\times D_2}$, 
$\operatorname{rank} \hat{X}_{\mathit{TM},\,i} = \|\hat{X}_{\mathit{TM},\,i}\|_F = 1$ 
and $\lambda_i \geqslant 0$ 
for all $i \in [KD]$.

The SVD of a t-matrix $X_\mathit{TM}$ is equivalent to the SVD of the matrix $M(X_\mathit{TM})$. Let the SVD of $M(X_\mathit{TM}) \in \mathbbm{C}^{KD_1\times KD_2} $ be 
$M(X_\mathit{TM}) = U \cdot S \cdot V^{*}$ such that 
$U^{*} U = V^{*} V = I_\mathit{KD}$ and $S = \operatorname{diag}(\sigma_1,\cdots, \sigma_{KD})$. Then, the matrix $M(X_\mathit{TM})$ can be written by the following linear combination, 
\begin{equation}
M(X_\mathit{TM}) = 
\sigma_1 \cdot (u_1  v_1^{*}) 
+\cdots+
\sigma_\mathit{KD} \cdot (u_\mathit{KD}  v_\mathit{KD}^{*})
\label{equation:SVD-of-tmatrix-matrix-representation}
\end{equation}
where $u_i $ denotes the 
$i$-th column of the matrix $U$, $v_i$ the 
$i$-th column of the matrix $V$ 
and $\sigma_i \geqslant 0$ the $i$-th singular value of the matrix $M(X_\mathit{TM})$
 for all $i \in [KD]$.

If the matrix $M(X_\mathit{TM})$ is full rank, 
it is not difficult to prove that there exists a unique 
norm-one, 
rank-one t-matrix $\hat{X}_{\mathit{TM}, i}$ 
such that $M(\hat{X}_{\mathit{TM}, i}) = u_i  v_i^{*}$, or equivalently, 
the following result holds for all $i\in [KD]$, 
\begin{equation}
u_i  v_i^{*} \in 
M(C_{1}^{D_1\times D_2} \cup \cdots \cup 
C_{K}^{D_1\times D_2} \setminus \{Z_\mathit{TM}\} ).
\label{equation:set-for-t-matrix}
\end{equation}
where $C_{k}^{D_1\times D_2} \doteq D_{T,\,k} \circ C^{D_1\times D_2} $ for all $k\in [K]$ and $Z_\mathit{TM}$ denotes the zero t-matrix in $C^{D_1\times D_2}$.

If the matrix $M(X_\mathit{TM})$ is rank deficient, the sequence of components $(u_1v_1^{*}),\cdots,(u_\mathit{KD}v_\mathit{KD}^{*})$ in 
equation (\ref{equation:SVD-of-tmatrix-matrix-representation}) is not unqiue. For example, let $\sigma_{k} = \cdots =  \sigma_\mathit{KD} = 0$ and 
\begin{equation}
\begin{aligned}
&[u_{k},\cdots,u_\mathit{KD}] \leftarrow 
\,[u_{k},\cdots,u_\mathit{KD}] 
\cdot A \\
&[v_{k},\cdots,v_\mathit{KD}\,] 
\leftarrow \,[v_{k},\cdots,v_\mathit{KD} \,] 
\cdot B \\
\end{aligned}
\end{equation}
where $A$ and $B$ denote any two $(KD-k +1)\times (KD-k +1)$ unitary matrices. Then, equation (\ref{equation:SVD-of-tmatrix-matrix-representation}) still holds
using any variants transfomed above.

However, among all variants of equation (\ref{equation:SVD-of-tmatrix-matrix-representation}), there must exist at least one variant such that  
equation (\ref{equation:set-for-t-matrix})
for all $i \in [KD]$.  
One can find the solution for computing such a variant in Section \ref{section:From-SVD-to-TSVD}.

Applying the mapping $M^{-1}$ to both sides of 
such an eligible variant of 
equation (\ref{equation:SVD-of-tmatrix-matrix-representation}) leads to the following linear combination, 
\begin{equation}
X_\mathit{TM} = \sigma_1 \cdot M^{-1}(u_1  v_1^{*}) 
+\cdots+
\sigma_\mathit{KD} \cdot M^{-1}(u_\mathit{KD}  v_\mathit{KD}^{*}) \;.
\label{equation:SVD-a-matrix2}
\end{equation}

Then, up to a permutation of combination terms, equation (\ref{equation:SVD-a-matrix2}) is identical to equation (\ref{equation:SVD-a-matrix001}). 

The TSVD (Tensorial Singular Value Decomposition) \cite{liao2020generalized,liao2021TPCA} is to decompose a t-matrix $X_\mathit{TM} \in C^{D_1\times D_2}$ as the following
$C$-linear combination
\begin{equation}
	X_\mathit{TM} = 
	S_{T,\,1} \circ Y_{\mathit{TM},\,1} + 
	\cdots + 
	 S_{T,\,D} \circ Y_{\mathit{TM},\,D}
\label{equation:TSVD}
\end{equation}
where $S_{T,\,i} \geq Z_{T}$, 
$\operatorname{rank}_t Y_{\mathit{TM},\,i} = E_{T} $ and 
$\| Y_{\mathit{TM},\,i} \|_{t, F}  = E_{T}$ 
hold for all $i \in [D]$.  

Note that the condition 
$\operatorname{rank} Y_{\mathit{TM},\,i} \doteq \operatorname{rank} M(Y_{\mathit{TM},\,i})  = K$ is  necessary but not sufficient for 
$\operatorname{rank}_{t} Y_{\mathit{TM},\,i} = E_{T}$. 
Analogously, 
$\| Y_{\mathit{TM},\,i} \|_F \doteq 
\| M(Y_{\mathit{TM},\,i}) \|_{F} = K $  is 
necessary but not sufficient for 
$\| Y_{\mathit{TM},,\,i} \|_{t, F} = E_T$. 
Thus, one can not use the condition  
$\operatorname{rank} Y_{\mathit{TM},\,i} = \| Y_{\mathit{TM},\,i} \|_F = K$
to characterize a $C$-linear model as equation (\ref{equation:TSVD}).

However, 
analogous to equation (\ref{equation:SVD-of-tmatrix-matrix-representation}), 
each t-matrix $Y_{\mathit{TM}, i}$ 
in equation (\ref{equation:TSVD})
can be written as the outer product of two t-vectors  
\begin{equation}
Y_{\mathit{TM}, i} = 
U_{\mathit{TV}, i} 
\circ  
V^{*}_{\mathit{TV}, i} \,\,,\,\forall i \in [D]
\label{equation:outproduct-t-vector}
\end{equation}
where $U_{\mathit{TV}, i} \in C^{D_1}, V_{\mathit{TV}, i} \in C^{D_2}$, 
\begin{equation}
U_{\mathit{TV}, i}^{*} \circ U_{\mathit{TV}, j} = 
V_{\mathit{TV}, i}^{*} \circ V_{\mathit{TV}, j} =
\delta_{i,j} \cdot E_{T} 
\,,\, \forall i,j \in [D], 
\label{equation:U-V-orthogonality}
\end{equation}
and $\delta_{i,j} \in  \{1, 0\}$ denotes the Kroneck delta.

The t-vector $U_{\mathit{TV},i}$ is called the $i$-th left singular t-vector, $V_{\mathit{TV},i} $ the  
$i$-th right singular t-vector and 
$S_{T, \,i}$ the $i$-th singular t-scalar of the t-matrix $X_\mathit{TM}$.

\subsection{From SVD to TSVD}
\label{section:From-SVD-to-TSVD}

One can exploit the introduced matrix/vector representation to obtain the TSVD of a t-matrix $X_\mathit{TM} \in C^{D_1\times D_2}$ via the SVD of the matrix $M(X_\mathit{TM}) \in \mathbbm{C}^{KD_1\times KD_2}$.

Let’s first rephrase the definition of the t-scalar-valued rank of a t-matrix. Interested readers are referred to \cite{liao2020generalized} for the original definition.

The t-scalar-valued rank is given as follows in terms of the matrix representation. Let the Moore-Penrose inverse of a t-matrix $X_\mathit{TM} \in C^{D_1\times D_2}$ be ${X}^{\dagger}_\mathit{TM} \in C^{D_2\times D_1}$. Then, the matrix representation of  ${X}^{\dagger}_\mathit{TM}$ is given by the pseudo-inverse of the matrix $M(X_\mathit{TM})$, namely,  
\begin{equation}   
M({X}^{\dagger}_\mathit{TM}) 
=
\big(\, M(X_\mathit{TM}) \,\big)^{\dagger}
\end{equation}

Let the TSVD of $X_\mathit{TM}$ be 
$X_\mathit{TM} = U_\mathit{TM} \circ S_\mathit{TM} \circ V^{*}_\mathit{TM}$ as in equation (\ref{equation:55}). 
It is easy to follow that the t-matrix $(S_\mathit{TM} \circ {S}^{\dagger}_\mathit{TM})$ is idempotent, namely,
\begin{equation}
(S_\mathit{TM} \circ {S}^{\dagger}_\mathit{TM}) \circ 
(S_\mathit{TM} \circ {S}^{\dagger}_\mathit{TM}) = (S_\mathit{TM} \circ {S}^{\dagger}_\mathit{TM}) \;. 
\end{equation}

The t-scalar-valued rank of $X_\mathit{TM}$ is given by the t-scalar-valued trace of the t-matrix $P_\mathit{TM} \doteq S_\mathit{TM} \circ {S}^{\dagger}_\mathit{TM}$ as follow,  
\begin{equation}
\operatorname{rank}_t(X_\mathit{TM}) = 
\operatorname{trace}_t P_\mathit{TM} \doteq  
\sum\nolimits_{i} \,
[P_\mathit{TM}]_{i,i}  
\,\geq\, 
Z_{T} \;.
\end{equation}

Let the matrix $M(X_\mathit{TM})\in \mathbbm{C}^{KD_1\times KD_2}$ be full rank and the ``economic-size''  SVD of $M(X_\mathit{TM})$ be the following matrix form 
\begin{equation}
M(X_\mathit{TM}) = U \cdot S  \cdot V^{*}
\end{equation}
where $U \in \mathbbm{C}^{KD_1 \times KD} $, 
$V \in \mathbbm{C}^{KD_2 \times KD} $,  
$S \doteq \operatorname{diag}(\sigma_1,\cdots,\sigma_\mathit{KD})$ such that 
$\sigma_i > 0, \,\forall\, i \in [KD]$, and
$D \doteq \min(D_1, D_2)$.

Let the $i$-th column of $U$ be $u_i$ for all 
$i \in [KD]$.  
It follows that 
the following results hold for all $i \in [KD]$, 
\begin{equation}
u_i \in 
V(D_{T,\,1} \circ C^{D_1} \cup \cdots \cup D_{T,\,k} \circ C^{D_1} \setminus \{Z_\mathit{TV}\})\;.
\label{equation:setminus-for-vector}
\end{equation}
where $Z_\mathit{TV}$ denotes the zero t-vector in $C^{D_1}$. 

Note that equation (\ref{equation:setminus-for-vector}) implies that $V^{-1}(u_i)$ is rank-one, namely, 
\begin{equation}
\operatorname{rank} V^{-1}(u_i) = 1 \;,i \in [KD] \;.
\label{equation:rank-V-ui}
\end{equation}

Since $u_i$ is a singular vector, it means that $V^{-1}(u_i)$ is norm-one, namely,   
\begin{equation}
\|V^{-1}(u_i)
\|_F  \equiv \|u_i\|_F  = 1 \;, \forall i \in [KD]\;.
\label{equation:norm-V-ui}
\end{equation}  

Notice that 
equation (\ref{equation:rank-V-ui})
also implies the idempotency of the t-scalar-valued rank of $V^{-1}(u_i)$, specifically, the folowing result holds for all $i \in [KD]$,       
\begin{equation}
\operatorname{rank}_t  
V^{-1}(u_i) 
\in \Big\{
D_{T,\,1},\cdots,D_{T,\,K}
\Big\}
\;.
\label{equation:78}
\end{equation}

Thus, one can partition all t-vectors 
$V^{-1}(u_i)$
into $K$ 
equicardinal 
classes $G_1,\cdots,G_K$ 
according to their t-scalar-valued ranks  
such that 
\begin{equation}
\begin{aligned}
\operatorname{card} G_k = D \doteq \min(D_1, D_2) \;.
\end{aligned}
\label{equation:cardinality}
\end{equation}  
holds for all $k \in [K]$.

All singular value $\sigma_i > 0$, 
associated with $V^{-1}(u_i) $,  
are converted to   
rank-one t-scalars   
\begin{equation}
\sigma_i \cdot 
\operatorname{rank}_t  
V^{-1}(u_i)
 \in  \Big\{
(\sigma_i \cdot D_{T,\,1}),\cdots, (\sigma_i \cdot D_{T,\,K})
 \Big\}   
\label{equation:singular-tvalue}
\end{equation} 
and partitioned into $K$ equicardinal classes $G^{S}_1,\cdots,G^{S}_K$ 
according to $\operatorname{rank}_t V^{-1}(u_i)$ 
such that the following holds for all $k \in [K]$,
\begin{equation}
\operatorname{card}(G^{S}_k) = 
\operatorname{card}(G_k) = D
\doteq \min(D_1, D_2) 
\;.
\label{equation:cardinality-equility}
\end{equation}  

Note that each class $G^{S}_k$ is totally ordered, namely,  either  
$(X_T - Y_T) \geq Z_T $ or $(Y_T - X_T) \geq Z_T $ holds  for all $X_T, Y_T \in G^{S}_k$. 
Let the t-scalars in $G^{S}_k$ be sorted descendingly and the t-vectors in 
$G_k$ are sorted accordingly for all $k \in [K]$. 
Then, the left singular t-vectors $U_{\mathit{TV}, 1},\cdots,U_{\mathit{TV}, D} \in C^{D_1}$ in equation (\ref{equation:outproduct-t-vector}) 
are given as follows, 
\begin{equation}
U_{\mathit{TV}, i} \doteq \{G_1\}_i + \cdots + \{G_K\}_i, \forall i \in [D]\;.
\label{equation:UTV}
\end{equation}
where $\{G_k\}_i \in D_{T,\,k} \circ C^{D_1} \setminus \{Z_\mathit{TV}\}$ denotes the $i$-th t-vector in $G_k$.

All these left singular t-vectors $U_{\mathit{TV},1},\cdots,U_{\mathit{TV},D}$, as columns, form the t-matrix 
$U_\mathit{TM} \in C^{D_1\times D}$ 
such that 
$U_\mathit{TM}^{*} \circ U_\mathit{TM} = I_\mathit{TM}$.

Analogously, one can compute $V_\mathit{TM} \in C^{D_2\times D}$, contaning the right singular t-vectors, such that $V_\mathit{TM}^{*} \circ V_\mathit{TM} = I_\mathit{TM}$. 

The $i$-th singular t-scalar in equation (\ref{equation:TSVD}) is given as follows,  
\begin{equation}
S_{T, \,i} \doteq  \{G^{S}_1 \}_i +\cdots + \{G^{S}_K\}_i \,\gneq Z_T \;,\; \forall i \in [D]
\label{equation:81}
\end{equation}
where  $\{G^{S}_k \}_i \in D_{T,\,k} \circ S^\mathit{nonneg} \setminus \{Z_T\} $ denotes the $i$-th t-scalar in $G^{S}_k $ for all
$k \in [K]$ and $i \in [D]$ such that 
$S_{T, \,1}  \geq \cdots \geq S_{T, \,D} \gneq  Z_T $. 

All the singular t-scalars form a diagonal t-matrix 
$S_\mathit{TM} \in C^{D \times D}$ such that the $(i,j)$-th t-scalar entry of $S_\mathit{TM}$ is given by 
\begin{equation}
[S_\mathit{TM}]_{i, j} \doteq \delta_{i,j} \cdot S_{T, i}\;, \;\forall i, j  \in [D] 
\end{equation} 
where $\delta_{i,j} \in \{0, 1\}$ denotes the Kronecker delta. 

\vskip 0.5em

When the t-matrix $X_\mathit{TM} $ is rank deficient, i.e., $\operatorname{rank} X_\mathit{TM} < KD $,  
the situation becomes a little tricky. 
Specifically, 
equation (\ref{equation:setminus-for-vector})
does not necessarily hold unless $\sigma_i \neq 0$.

If $\sigma_i \neq 0$, the corresponding t-vector $V^{-1}(u_i) $ 
is classified to the class $G_k$ if and only if 
\begin{equation}
\operatorname{rank}_t  V^{-1}(u_i) = D_{T,\,k}.
\label{equation:rank-one-t-rank}
\end{equation} 
It leads to the following result  
\begin{equation}
\begin{aligned}
\operatorname{card} G_k \leqslant D\,\,,\,\forall k \in [K]\,, \\ 
\operatorname{card} G_1 + \cdots +
\operatorname{card} G_K =\operatorname{rank} X_\mathit{TM}  
< K  D
\,.
\end{aligned}
\label{equation:cardsum}
\end{equation}

It follows from equation (\ref{equation:cardsum}) that there exists at least a ${k}' \in [K]$ such that the following inequality holds 
\begin{equation}
\operatorname{card} G_{k'} < D \;.
\end{equation}

On the other hand, one can convert 
a non-zero singular value $\sigma_i$ to its rank-one t-scalar counterpart 
$\sigma_i \cdot \operatorname{rank}_t  V^{-1}(u_i) $
via equation (\ref{equation:singular-tvalue}). Each 
of these rank-one
t-scalars 
is classified 
to the class $G^{S}_k$ if and only if 
equation (\ref{equation:rank-one-t-rank}) holds. 
It follows that the cardinalities of $G^{S}_k$ and 
$G_k$ are equal for 
all $k \in [K]$, 
\begin{equation}
\operatorname{card} G^{S}_k = \operatorname{card} G_k \;.
\end{equation}

The t-vectors in $G_k$, as columns, form a t-matrix 
$U_{\mathit{TM}, \,k}\in C^{D_1\times (\operatorname{card} G_k) }$
such that 
\begin{equation}
U_{\mathit{TM}, \,k}^{*} \circ U_{\mathit{TM}, \,k} = I_\mathit{TM}
\in C^{(\operatorname{card} G_k) \times (\operatorname{card} G_k)}
\,,
\forall k \in [K].
\end{equation}
The column module of $U_{\mathit{TM},k}$ 
is the $C$-linear span of the columns of $U_{\mathit{TM},k}$, and can be characterized by the idempotent t-matrix
$P_{\mathit{TM}, k} \doteq U_{\mathit{TM},k} \circ U^{*}_{\mathit{TM},k} \in C^{D_1\times D_1}$ 
such that 
the following result holds for all $k \in [K]$, 
\begin{equation}
\begin{aligned}
\operatorname{rank}_t P_{\mathit{TM},\,k} &\equiv 
\operatorname{rank}_t U_{\mathit{TM},k} \circ U^{*}_{\mathit{TM},k} \\ 
&= 
(\operatorname{card} G_k) \cdot D_{T,\,k} \leq D \cdot D_{T,\,k} \,.
\end{aligned}
\end{equation}

The lower-dimensional orthogonal complement of the column module 
of $U_{\mathit{TM},k}$
can be characterized by the following idempotent t-matrix $\tilde{P}_{\mathit{TM},\,k} \in C^{D_1\times D_1}$,  
\begin{equation}
\tilde{P}_{\mathit{TM},\,k} = D_{T,\,k} \circ I_{TM}  - U_{\mathit{TM},k} \circ U^{*}_{\mathit{TM},k}  \,,
\forall k \in [K].
\end{equation}

The t-scalar-valued rank of $\tilde{P}_{\mathit{TM}, \,k} $ is given as follows. 
\begin{equation} 
\operatorname{rank}_t \tilde{P}_{\mathit{TM},\,k}  = (D - \operatorname{card} G_k) \cdot D_{T,\,k} 
\,.
\end{equation}

The (canonical) rank of $\tilde{P}_{\mathit{TM}, \,k} $ is given by 
\begin{equation}
\operatorname{rank} \tilde{P}_{\mathit{TM},\,k} = (D - \operatorname{card} G_k) \cdot 
\operatorname{trace} D_{T,\,k} = D - \operatorname{card} G_k \;.
\end{equation}

Let 
the $i$-th left singular vector 
of the matrix $M(\tilde{P}_{\mathit{TM},\,k}) \in \mathbbm{C}^{KD_1\times KD_2} $ be 
$w_i \in \mathbbm{C}^{KD_1} $ for all $i \in [D - \operatorname{card} G_k]$. 
It is not difficult to verify the following result hold for all 
$i \in [D - \operatorname{card} G_k]$, 
\begin{equation}
\begin{aligned}
w_i \in V(D_{T,\,k} \circ C^{D_1} \setminus \{Z_\mathit{TV}\} )\,, \\
\operatorname{rank}_t V^{-1}(w_i) = D_{T,\,k} \,.
\end{aligned}
\end{equation}

Denote 
\begin{equation}
R_k \doteq D - \operatorname{card} G_k \;.
\label{equation:Rk}
\end{equation}
Then, adding the   
$R_k$ left singular t-vectors 
$V^{-1} (w_i) \in C^{D_1} $
to $G_k$ and $R_k$  
zero t-scalars $Z_{T}$
to $G^{S}_k$,
one has the equicardinal classes 
$G_1,\cdots,G_K$ and $G^{S}_1,\cdots,G^{S}_K$  
such that equation (\ref{equation:cardinality-equility}) holds.

Each class $G^{S}_{k}$ of nonnegative t-scalars is totally ordered such that either $(X_{T} - Y_{T})$ or $(Y_T - X_T)$ is nonnegative for all $X_T, Y_T \in G^{S}_{k}$. 
Then, as in the situation already discussed, after sorting the nonnegative t-scalars of each class of $G^{S}_1, \cdots, G^{S}_K$ descendingly, and the rank-one t-vectors of each class of $G_1, \cdots, G_K$ accordingly, one can use equations (\ref{equation:UTV}) and (\ref{equation:81}) to yield the $i$-th left singular t-vector 
$U_{\mathit{TV}, i}$ and singular t-scalar $S_{T, \,i}$ for all $i \in [D]$. 
Analogous to the left singular t-vectors, the right singular t-vectors can be computed via a series of associated SVDs.

\vskip 0.5em

It is noted that equation (\ref{equation:55}) is the ``economic-size'' TSVD of $X_\mathit{TM} \in C^{D_1\times D_2}$ such that $S_\mathit{TM}$ is a t-matrix in $C^{D\times D}$.  

If one needs the non-economic-size TSVD of $X_\mathit{TM} \in C^{D_1\times D_2}$ such that 
$U_\mathit{TM} \in C^{D_1\times D_1}$, 
$V_\mathit{TM} \in C^{D_2\times D_2}$ and 
$S_\mathit{TM} \in C^{D_1\times D_2}$, 
what one needs for having $U_\mathit{TM} \in C^{D_1\times D_1}$ 
is to rewrite equation (\ref{equation:Rk}) to 
$R_k \doteq D_1 - \operatorname{card} G_k$ such that the condition 
$\operatorname{card} G_k = D_1 $
holds for all $k \in [K]$, 
and equation (\ref{equation:UTV}) holds for all $i \in [D_1]$. Analogously, one can the t-matrix have $V_\mathit{TM} \in C^{D_2\times D_2}$. 

Further, with the singular t-scalars given by equation (\ref{equation:81}), one has $S_{TM} \in C^{D_1\times D_2}$ whose $(i,j)$-th t-scalar entry of $S_{TM}$ is given by 
\begin{equation}
[S_\mathit{TM}]_{i,j} \doteq \delta_{i,j} \cdot S_{T, \,\min(i, j)}  \geq Z_T 
\end{equation} for all $(i, j) \in [D_1]\times [D_2]$ where $\delta_{i,j} \in \{1, 0\}$ denotes the Kroneck delta. 

{\mycolor{blue}
\subsection{From TSVD to SVD}

From equations (\ref{equation:TSVD}) 
and (\ref{equation:outproduct-t-vector}), 
it is not difficult to get  the SVD of 
$X_\mathit{TM}$ as equation (\ref{equation:SVD-a-matrix001}).

First, one can always rewrite the t-matrix $X_\mathit{TM} \in C^{D_1\times D_2}$ as follows, 
\begin{equation}
\begin{aligned}	
X_{TM} &\equiv (D_{T, 1} +\cdots + D_{T, K}) \circ X_\mathit{TM}  \\
& = \sum\limits_{k=1}^{K}\sum\limits_{i=1}^{D} 
\Big(
D_{T, k} \circ S_{T, i} \circ Y_{{TM},i} 
\Big)
\end{aligned}
\end{equation}
where the following condition holds for all $(k, i) \in [K]\times [D]$,  
\begin{equation}
\begin{aligned}
&~~~D_{T, k} \circ S_{T, i} \circ Y_{{TM},i} \\ 
&= 
(D_{T, k} \circ S_{T, i}) \circ 
(D_{T, k} \circ Y_{{TM},i} )  \\
& = \big(F_k(S_{T, i}) \cdot D_{T, k} \big) \circ (D_{T, k} \circ Y_{{TM},i} ) \\
& = F_k(S_{T, i}) \cdot (D_{T, k} \circ Y_{{TM},i} ) \; \\
& = |{F}_k(S_{T, i})| \cdot  \hat{Y}_{{TM},k, i}     
\end{aligned}
\end{equation}
where $|{F}_k(S_{T, i})| \geqslant 0$ is the modulus of  $F_k(S_{T, i}) \in \mathbbm{C}$ and the t-matrix $\hat{Y}_{{TM},k, i} \in C^{D_1\times D_2}$, for all $(k, i) \in [K]\times [D]$, is given by the following equation,  
\begin{equation}
\hat{Y}_{\mathit{TM}, k, i} = 
\left\{
\begin{aligned}
& \frac{{F}_k(S_{T, i}) \cdot   D_{T,k}}{ |{F}_k(S_{T, i})|  }  \circ Y_{\mathit{TM}, i}   &\text{if\;} {F}_k(S_{T, i}) \neq 0 \\
&D_{T,k} \circ Y_{\mathit{TM}, i} &\text{otherwise\;}
\end{aligned}
\right.
\end{equation}

Since $\operatorname{rank}_t Y_{\mathit{TM},i} = \|
Y_{\mathit{TM},i} 
\|_{t, F} = E_{T}$, 
it follows that the following 
holds for all $(k, i) \in [K]\times [D]$,   
\begin{equation}
\operatorname{rank}_t \hat{Y}_{\mathit{TM},k, i} = \|
\hat{Y}_{\mathit{TM},k, i} 
\|_{t, F} = D_{T,k}  \;.
\label{equation:generalized-norm-one-rank-one}
\end{equation} 

Note that equation (\ref{equation:generalized-norm-one-rank-one}) is sufficient to stating the following condition of rank-one and norm-one holds for all $(k, i) \in [K]\times [D]$, 
\begin{equation}
\operatorname{rank} \hat{Y}_{\mathit{TM},k, i} = \|
\hat{Y}_{\mathit{TM},k, i} 
\|_{F} = 1  \;.
\label{equation:canonical-norm-one-rank-one}
\end{equation} 

Thus, one can decompose the t-matrix $X_\mathit{TM}$ as the following linear sum of a finite number of  rank-one, norm-one components, 
\begin{equation}
X_\mathit{TM} = 
\sum\limits_{k=1}^{K}
\sum\limits_{i=1}^{D} |F_{k}(S_{T, i})| \cdot \hat{Y}_{\mathit{TM},k, i}
\;,
\nonumber
\end{equation}
which, 
equivalent to equation (\ref{equation:SVD-a-matrix001}), is the SVD of the t-matrix $X_\mathit{TM}$.  
}

\section*{Acknowledgments}

Liang Liao and Stephen John Maybank contribute equally to the theory of t-algebra, t-scalars, t-vectors, t-matrices, g-tensors, and the matrix representation. 
Liang Liao designs the experiments of this paper. Other authors help significantly with the experiments presented in this paper. Liang Liao would like to thank the Birkbeck Institute of Data Analytics for free use of the high-performance computing facilities during Liao's visit to Birkbeck College, University of London.

\section*{Code repository}

\begin{center}
\url{https://github.com/liaoliang2020/talgebra}
\end{center}
	
\balance

\end{document}